\documentclass[table]{ai2style/ai2}
\usepackage{amssymb}
\usepackage{multirow}
\usepackage{bigdelim}
\usepackage{todonotes}
\usepackage{longtable}
\usepackage{tabularray}
\usepackage{wrapfig}
\usepackage[most]{tcolorbox} 
\usepackage{url}
\usepackage{xspace}
\usepackage{svg}
\usepackage[absolute]{textpos} %
\usepackage{fdsymbol}   %
\usepackage{csvsimple}
\usepackage{nicematrix}
\newcolumntype{L}[1]{>{\raggedright\let\newline\\\arraybackslash\hspace{0pt}}m{#1}}
\newcolumntype{C}[1]{>{\centering\let\newline\\\arraybackslash\hspace{0pt}}m{#1}}
\newcolumntype{R}[1]{>{\raggedleft\let\newline\\\arraybackslash\hspace{0pt}}m{#1}}
\newcolumntype{P}[1]{>{\centering\let\newline\\\arraybackslash\columncolor{ai2lightpink}}m{#1}}
\usepackage{booktabs}              
\usepackage{adjustbox}             
\usepackage[dvipsnames,table]{xcolor}  
\usepackage[utf8]{inputenc} %
\usepackage[T1]{fontenc}    %
\usepackage{url}            %
\usepackage{booktabs}       %
\usepackage{amsfonts}       %
\usepackage{nicefrac}       %
\usepackage{microtype}      %
\usepackage[table]{xcolor}         %
\usepackage{amsmath}
\usepackage[most]{tcolorbox}
\usepackage{csquotes}

\usepackage{siunitx}
\usepackage{graphicx}
\usepackage{arydshln}
\usepackage{wrapfig}
\usepackage{enumitem}
\usepackage{soul} %

\usepackage{booktabs}
\usepackage{multirow}
\usepackage{threeparttable}
\usepackage{multirow}
\usepackage{xspace}
\usepackage{adjustbox}
\usepackage{pifont}
\usepackage{caption}
\usepackage{makecell}
\usepackage{subcaption}
\usepackage{bold-extra}
\usepackage{url}

\usepackage{booktabs,rotating,xcolor}

\usepackage{hyperref}
\definecolor{linkcolor}{RGB}{0, 0, 128}
\hypersetup{
     colorlinks   = true,
     citecolor    = linkcolor,
     linkcolor    = linkcolor,
     urlcolor     = linkcolor,
}
\usepackage{pifont}%
\newcommand{\cmark}{\ding{51}}%
\newcommand{\xmark}{\ding{55}}%
\usepackage{listings}
\usepackage{booktabs}
\usepackage{multirow}
\usepackage{rotating}
\usepackage{makecell}
\usepackage{array}
\setlist[itemize]{leftmargin=*,itemsep=0em,parsep=0.3em,topsep=0.3em}

\usepackage{array}  

\usepackage{tikz}

\definecolor{maroon}{HTML}{F26035}
\definecolor{yellow}{HTML}{FDBC42}
\definecolor{darkred}{RGB}{156, 39, 33}
\definecolor{darkblue}{RGB}{31, 90, 153}
\definecolor{forestgreen}{rgb}{0.13, 0.55, 0.13}

\newcommand{\molmoact}{\textsc{MolmoAct}\xspace}
\newcommand{\molmoactt}{\textsc{MolmoAct2}\xspace}

\newcommand{\molmoactfast}{\textsc{MolmoAct2-FAST Tokenizer}\xspace}

\newcommand{\molmoactdata}{\textsc{MolmoAct Dataset}\xspace}
\newcommand{\molmoer}{\textsc{Molmo2-ER}\xspace}
\newcommand{\molmot}{\textsc{Molmo2}\xspace}
\newcommand{\molmoacttpre}{\textsc{MolmoAct2-Pretrain}\xspace}
\newcommand{\molmoacttpos}{\textsc{MolmoAct2-Post}\xspace}
\newcommand{\molmoacttlib}{\textsc{MolmoAct2-LIBERO}\xspace}

\newcommand{\molmothink}{\textsc{MolmoAct2-Think}\xspace}
\newcommand{\molmoacttdroid}{\textsc{MolmoAct2-DROID}\xspace}
\newcommand{\molmoacttso}{\textsc{MolmoAct2-SO100/101}\xspace}
\newcommand{\molmoacttyam}{\textsc{MolmoAct2-BimanualYAM}\xspace}
\newcommand{\molmoacttdroiddata}{\textsc{MolmoAct2-DROID Dataset}\xspace}
\newcommand{\molmoacttsodata}{\textsc{MolmoAct2-SO100/101 Dataset}\xspace}
\newcommand{\molmoacttyamdata}{\textsc{MolmoAct2-BimanualYAM Dataset}\xspace}

\newcommand{\molmothinklibero}{\textsc{MolmoAct2-Think-LIBERO}\xspace}

\newcommand{\libero}{LIBERO\xspace}

\newcommand{\huggingface}{\raisebox{-1.5pt}{\includegraphics[height=1.05em]{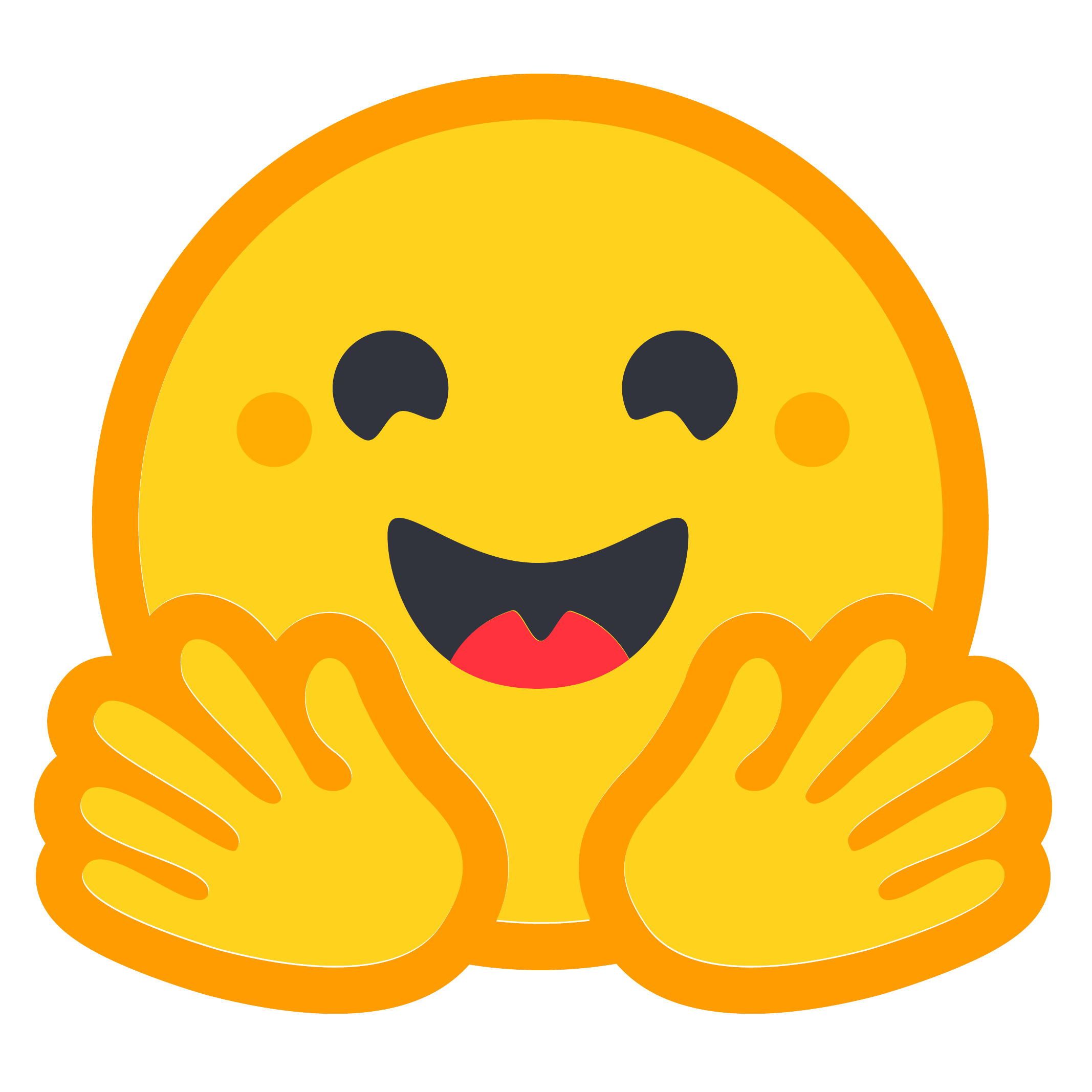}}\xspace}

\newcommand{\hfdataset}{\raisebox{-1.5pt}{\includegraphics[height=1.05em]{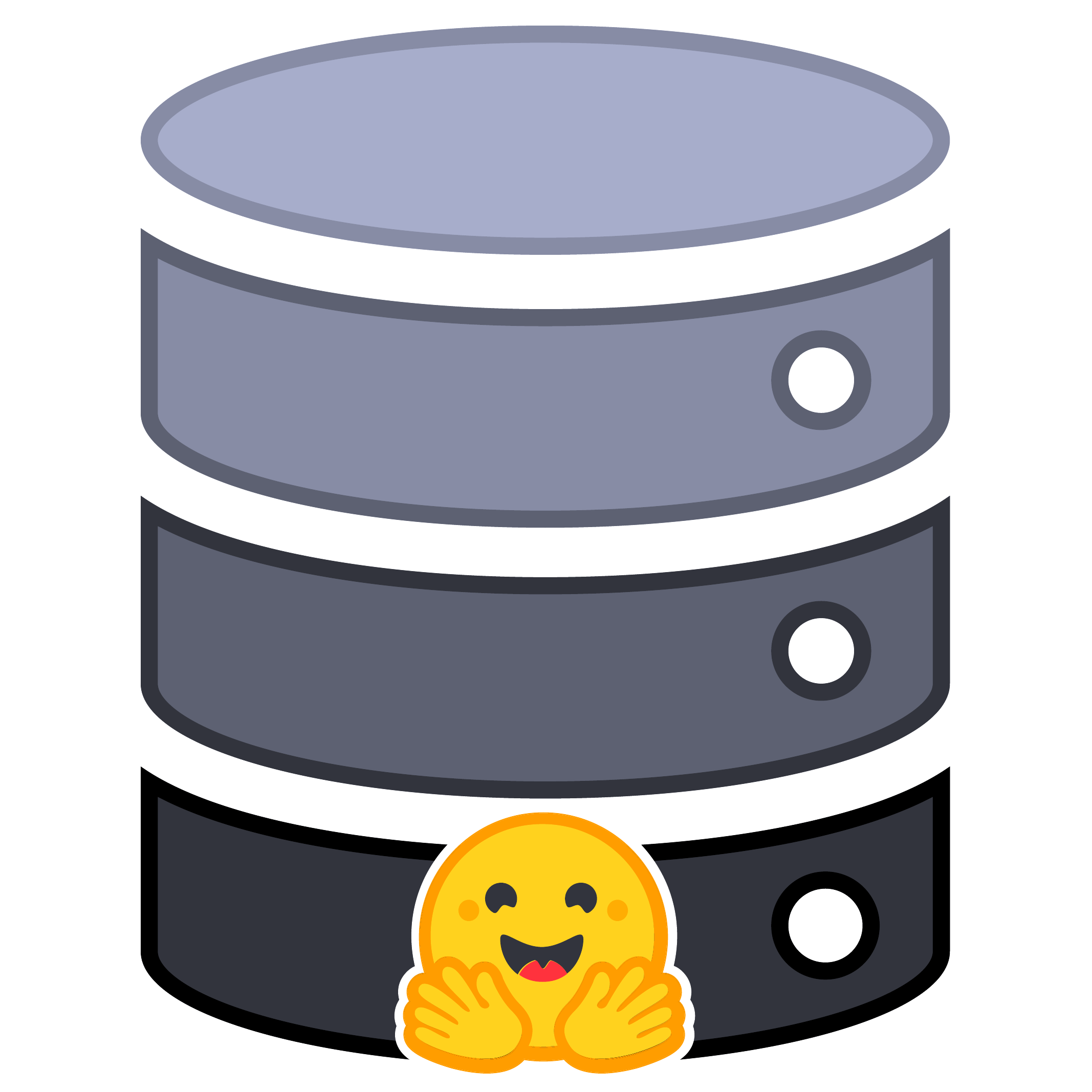}}\xspace}
\newcommand{\github}{\raisebox{-1.5pt}{\includegraphics[height=1.05em]{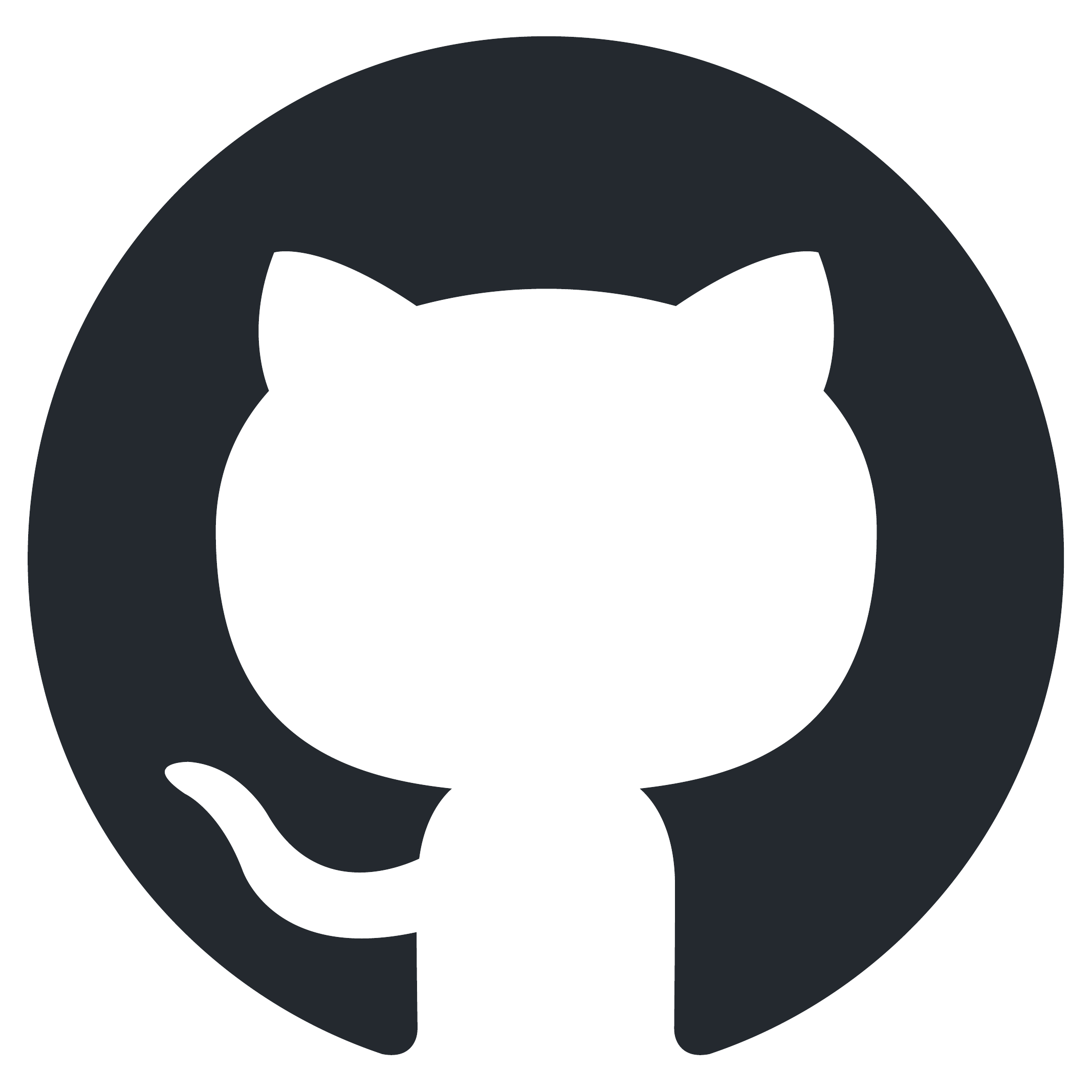}}\xspace}
\newcommand{\aitoo}{\raisebox{-1.5pt}{\includegraphics[height=1.05em]{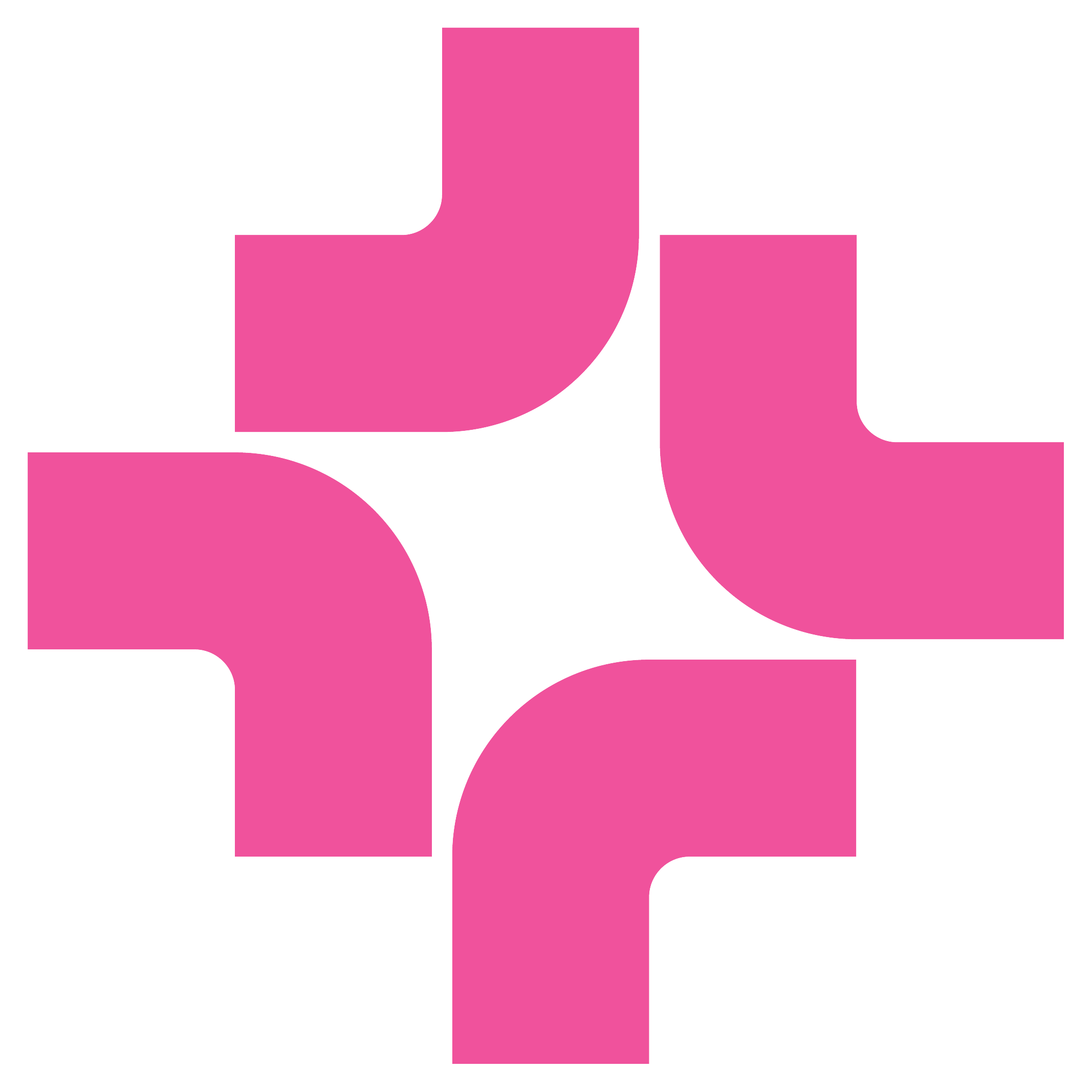}}\xspace}

\usepackage{setspace}
\usepackage{quoting}      

\usepackage{nicematrix}
\newcolumntype{L}[1]{>{\raggedright\let\newline\\\arraybackslash\hspace{0pt}}m{#1}}
\newcolumntype{C}[1]{>{\centering\let\newline\\\arraybackslash\hspace{0pt}}m{#1}}
\newcolumntype{R}[1]{>{\raggedleft\let\newline\\\arraybackslash\hspace{0pt}}m{#1}}
\newcolumntype{P}[1]{>{\centering\let\newline\\\arraybackslash\columncolor{ai2lightpink}}m{#1}}
\newcolumntype{W}[1]{>{\columncolor{white}}c}  %
\title{MolmoAct2\\{\fontsize{18pt}{12pt}\selectfont Action Reasoning Models for Real-World Deployment}}

\newcommand{\core}{\textsuperscript{\textcolor{ai2pink}{\ding{170}}}}
\authorOne[1,2*]{Haoquan Fang\core}
\authorOne[1,2,3*]{Jiafei Duan\core}

\authorTwo[1,2]{Donovan Clay\core}
\authorTwo[1,4]{Sam Wang\core}
\authorTwo[2]{Shuo Liu\core}
\authorTwo[1,2]{Weikai Huang\core}
\authorTwo[1,2]{Xiang Fan\core}
\authorTwo[2]{Wei-Chuan Tsai\core}
\authorTwo[1,2]{Shirui Chen\core}
\authorTwo[1,2]{Yi Ru Wang\core}
\authorTwo[2]{Shanli Xing\core}

\authorThree[1,2,5]{Jaemin Cho}
\authorThree[1]{Jae Sung Park}
\authorThree[1,2]{Ainaz Eftekhar}
\authorThree[1]{Peter Sushko}
\authorThree[1]{Karen Farley}
\authorThree[2]{Angad Wadhwa}
\authorThree[6]{Cole Harrison}
\authorThree[1]{Winson Han}
\authorThree[2]{Ying-Chun Lee}
\authorThree[1]{Eli VanderBilt}
\authorThree[1]{Rose Hendrix}
\authorThree[7]{Suveen Ellawela}
\authorThree[7]{Lucas Ngoo}

\authorFour[8]{Joyce Chai}
\authorFour[1,2,9]{Zhongzheng Ren\core}
\authorFour[1,2]{Ali Farhadi\core}
\authorFour[1,2]{Dieter Fox\core}
\authorFour[1,2]{Ranjay Krishna\core}

\affiliation[1]{Allen Institute for AI}
\affiliation[2]{University of Washington}
\affiliation[3]{National University of Singapore}
\affiliation[4]{University of Pennsylvania}
\affiliation[5]{Johns Hopkins University}
\affiliation[6]{Amazon}
\affiliation[7]{Cortex AI}
\affiliation[8]{University of Michigan}
\affiliation[9]{University of North Carolina at Chapel Hill}

\contribution[]{
* denotes equal contribution in no particular order. $\textcolor{ai2pink}{\varheartsuit}$ marks core contributors. See full author contributions \hyperref[sec:contrib]{here}.
}

\abstract{

Vision-Language-Action (VLA) models aim to provide a single generalist controller for robots, but today's systems fall short for real-world deployment. Frontier models are closed; open-weight alternatives are tied to expensive hardware; reasoning-augmented policies pay prohibitive latency for their grounding; and fine-tuned success rates remain below the threshold for dependable use. 
We present \molmoactt, a fully open action reasoning model built for practical deployment, advancing its predecessor, \molmoact along five axes. 
(1) \molmoactt is built on top of our new \molmoer, a VLM backbone specialized for spatial and embodied reasoning, trained on a 3.3M-sample corpus with a specialize-then-rehearse recipe. 
(2) We release three new robot datasets spanning low-to-medium cost platforms: \molmoacttyamdata, 720 hours of teleoperated bimanual trajectories that constitute the largest open bimanual dataset to date; \molmoacttdroiddata, a quality-filtered Franka subset of DROID; and \molmoacttsodata, a quality-filtered SO-100/101 subset. 
(3) We train and release \molmoactfast, an open-weight, open-data action tokenizer trained on millions of trajectories across five embodiments.
(4) We design a new VLA architecture to graft the discrete-token VLM into the flow-matching continuous-action expert via per-layer key-value (KV) conditioning.
(5) we propose \molmothink, an adaptive-depth reasoning variant that re-predicts depth tokens only for scene regions that change between timesteps, retaining geometric grounding at a fraction of prior latency. 
In the most extensive empirical study of any open VLA to date, spanning 7 simulation and real-world benchmarks, \molmoactt outperforms strong baselines including $\pi_{0.5}$, while \molmoer surpasses GPT-5 and Gemini Robotics ER-1.5 across 13 embodied-reasoning benchmarks. We release model weights, training code, and complete training data.

}

\metadata[\quad\huggingface MolmoAct2:]{
\href{https://huggingface.co/collections/allenai/molmoact2-finetuned-models}{\texttt{MolmoAct2-Finetune}} \quad \href{https://huggingface.co/allenai/MolmoAct2}{\texttt{MolmoAct2}} \quad \href{https://huggingface.co/allenai/MolmoAct2-Think}{\texttt{MolmoAct2-Think}} \quad \href{https://huggingface.co/allenai/MolmoAct2-Pretrain}{\texttt{MolmoAct2-Pretrain}} \quad \href{https://huggingface.co/allenai/Molmo2-ER-4B}{\texttt{Molmo2-ER}}
}

\metadata[\vspace{.5em}\quad\hfdataset MolmoAct2 Data:]{
\href{https://huggingface.co/collections/allenai/molmoact2-bimanualyam-dataset}{\texttt{MolmoAct2-BimanualYAM}} \quad
\href{https://huggingface.co/datasets/allenai/MolmoAct2-SO100_101-Dataset}{\texttt{MolmoAct2-SO100/101}} \quad \href{https://huggingface.co/datasets/allenai/MolmoAct2-DROID-Dataset}{\texttt{MolmoAct2-DROID}} \quad \href{https://huggingface.co/collections/allenai/molmo2-er-datasets}{\texttt{Molmo2-ER}} 
}

\metadata[\vspace{.5em}\quad\github Code:]{
    \href{https://github.com/allenai/molmoact2}{\texttt{allenai/molmoact2}}
}

\metadata[\vspace{.5em}\quad\aitoo Blog:]{\href{https://allenai.org/blog/molmoact2}{\texttt{allenai.org/blog/molmoact2}}}

\begin{document}

\maketitle

\setcounter{tocdepth}{2}%


\section{Introduction}
\label{sec:intro}


Physical intelligence is fundamentally organized around perception and action~\citep{tversky2025yourbody}. Rather than reasoning over abstract internal computation, human think by constructing spatial representations, simulating actions, and interacting with the world through their bodies~\citep{tversky2019mind,tversky2013visualizing}.
Although we conflate today's robot foundation models as displaying such intelligence; from a cognitive science perspective, they remain incomplete models of intelligence. They often lack structured spatial representations~\cite{qu2025spatialvla}, rely on heavyweight internal reasoning processes that impede real-time interaction~\citep{lee2025molmoactactionreasoningmodels,kim2026cosmos,zhu2025unified}, and are difficult to adapt or extend due to limited openness~\citep{black2410pi0,intelligence2025pi05visionlanguageactionmodelopenworld}.

Recent work has shown promise that reasoning processes improve performance but the improvements come  at the cost of high inference latency.
Recent systems including MolmoAct~\citep{lee2025molmoactactionreasoningmodels} and others~\citep{zhao2025cot,sun2024emma,zheng2024tracevla} have shown that grounded spatial reasoning, predicted goal images, point trajectories, or full world-model rollouts improve both action quality and interpretability. In current implementations, however, this reasoning dominates inference latency: hundreds of tokens or entire predicted frames must be generated before a single action is emitted; emerging world models~\citep{kim2026cosmos,zhu2025unified} compound the problem with heavyweight per-step rollouts. The very mechanism intended to make policies more reliable thus renders them too slow for closed-loop control.

Reasoning, by itself, is only as good as the underlying foundation model that uses the reasoning process. Most frontier robot policies remain closed off; open-source alternatives are embodiment-specific and hard to adapt to new tasks or embodiments.
Frontier vision-language-action (VLA) models~\citep{team2024gemini,team2025gemini,rhoda2026dva,generalist2026gen1} are effectively closed systems: their training data, recipes, and model weights are proprietary. The few exceptions release weights alone, withholding the data and training procedures needed to reproduce or extend them. This opacity both impedes scientific progress and prevents practitioners from adapting these models to their own robots or fine-tuning on in-house demonstrations.
The few open-weights VLAs ~\citep{black2410pi0,intelligence2025pi05visionlanguageactionmodelopenworld} that can be run out-of-the-box are tied to expensive or specialized robot platforms beyond the reach of most academic labs and independent researchers. This constrains not only who can use these models, but also the diversity of settings in which they can be evaluated and improved.
Zero-shot performance remains brittle, and even after task-specific fine-tuning, success rates on realistic tasks fall well below the threshold required for dependable deployment.

\begin{figure}[t]
\centering
\includegraphics[width=\linewidth]{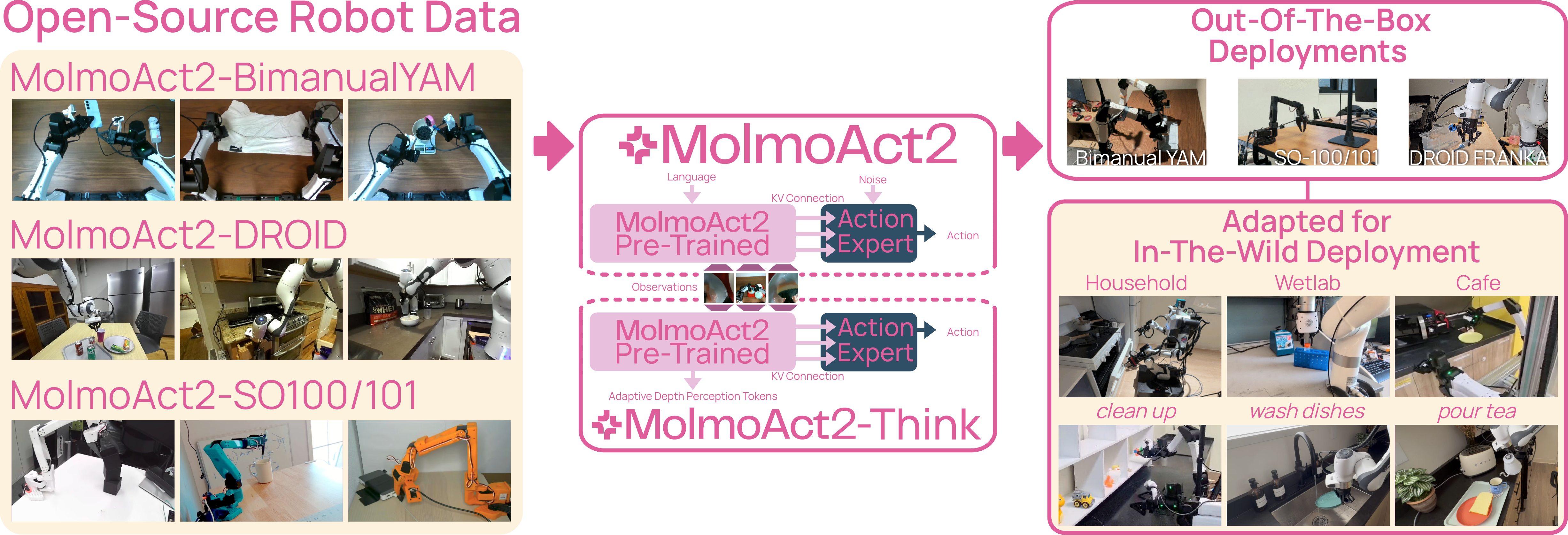}
\caption{\textbf{Overview of \molmoactt.} \molmoactt is a fully open action reasoning model for real-world deployment. From a suite of high-quality robot datasets that we collect, filter, and curate at scale across three platforms spanning the low-to-medium cost range (\textbf{left}), we train \molmoactt and its adaptive-depth reasoning variant \molmothink, coupled to the VLM backbone through per-layer KV conditioning (\textbf{center}). The resulting models deploy out-of-the-box on bimanual YAM, SO-100/101, and DROID Franka, and adapt to in-the-wild tasks such as cleaning up, washing dishes, wetlab automation, and pouring tea (\textbf{right}).}
\label{fig:teaser}
\end{figure}

We present \textbf{\molmoactt}, an action reasoning model built for real-world deployment: fully open, deployable out-of-the-box on multiple embodiments, performant, and capable of fast, interpretable reasoning (Figure~\ref{fig:teaser}). \molmoactt improves over its predecessor MolmoAct~\citep{lee2025molmoactactionreasoningmodels} along five axes vital for a strong action reasoning model: (1) a stronger open embodied-reasoning VLM backbone, (2) new open-source training datasets, (3) \molmoactfast, an open-source multi-embodiment action tokenizer, (4) a new VLA architecture design, and (5) a new adaptive reasoning paradigm for efficient inference.

First, we train \textbf{\molmoer} (Sec.~\ref{sec:er}), a VLM specialized for spatial and embodied reasoning.
General-purpose VLMs rarely train on or test the embodied skills a robot policy needs: they need to understand metric distances, free space, cross-view object tracking, and scene geometry. 
We address this by training \molmoer on a 3.3M-sample spatial–embodied corpus using a specialize-then-rehearse recipe. 

Second, existing open robot datasets remain fragmented across embodiments, uneven in quality, and too small or noisy to support reliable multi-embodiment action learning.
To support training action models on top of our VLM backbone, we release three action datasets (Sec.~\ref{sec:data}) targeting three platforms across a low-to-medium cost range: \textbf{\molmoacttyamdata}, 720 hours of teleoperated YAM trajectories spanning tabletop and household tasks—the largest open bimanual dataset to date; \textbf{\molmoacttsodata}, a filtered subset of internet SO-100/101 data with mislabeled and low-quality trajectories removed; and \textbf{\molmoacttdroiddata}, a quality-filtered Franka subset of DROID. For the two filtered datasets, we also re-annotate the language instructions for improved diversity and accuracy. 

Third, existing action tokenizers are either closed, tied to specific action spaces, or insufficiently documented, making it difficult to train reproducible discrete-action VLAs across embodiments.
To make the trajectories in our data usable under a discrete autoregressive objective, we introduce \molmoactfast, an open-weight, open-data implementation following FAST~\citep{pertsch2025fast}. We train and release its weights along with the millions of trajectories across five embodiments used to train it. \molmoactfast compresses one second of 32-D continuous actions into a compact discrete sequence.

Fourth, discrete-token VLMs provide strong reasoning grounding, but their native output space is poorly matched to the continuous, high-frequency trajectories required by robot control. A new architecture is therefore needed to connect the discrete reasoning capacity with smooth continuous actions.
We design a new architecture that conditions each layer of the continuous action expert on the keys and values from the corresponding VLM layer.
This design is in contrast to existing VLAs with action experts in two ways. First, we use a DiT-style transformer trained with flow matching objective, giving the continuous controller a modern denoising-transformer architecture that has proven more effective than many alternatives for diffusion and flow-based generative modeling. Second, instead of conditioning the expert on VLM hidden states, we condition each expert layer on the corresponding VLM keys and values. This preserves a similar compute profile while exposing the attention state used by the VLM itself, which has been shown to be more effective in our ablation studies.

Fifth, prior reasoning-based VLAs improve action quality by generating dense intermediate representations at every step~\cite{lee2025molmoactactionreasoningmodels}, but this repeats nearly identical computation across largely static scenes and makes inference too slow for real-time control.
We introduce \textbf{\molmothink}, the reasoning variant of \molmoactt, which performs \textit{adaptive} depth reasoning by autoregressively predicting only the tokens for scene regions that change between timesteps. This exploits trajectory-level temporal redundancy to reduce latency proportional to the static scene fraction while retaining the geometric grounding that significantly improves model performance.

To understand the capabilities of \molmoactt and assess its readiness for real-world deployment, we conduct the most extensive empirical study of any open VLA to date, spanning 7 environment benchmarks across both simulation and the real world. Across all of them, \molmoactt, with \molmothink, outperforms every strong baseline. We show that \molmoactt's fine-tuned checkpoints, \molmoacttdroid and \molmoacttso, can be deployed out of the box on their respective embodiments without any additional fine-tuning, significantly surpassing $\pi_{0.5}$~\citep{intelligence2025pi05visionlanguageactionmodelopenworld} (Sec.\ref{sec:exp-outbox}). We further demonstrate that \molmoactt is built on one of the strongest embodied-reasoning VLM backbones available: \molmoer surpasses models such as GPT-5~\citep{singh2025openai} and Gemini Robotics Embodied Reasoning (ER)-1.5~\citep{team2025gemini15} on 13 standard embodied-reasoning benchmarks (Sec.\ref{sec:exp-er}). 

For rapid real-world deployment via efficient fine-tuning to new embodiments, \molmoactt opens large performance gaps over strong general-purpose VLAs such as $\pi_{0.5}$~\citep{intelligence2025pi05visionlanguageactionmodelopenworld} not only on the two mainstream simulation benchmarks LIBERO~\citep{liu2023libero} and RoboEval~\citep{wang2025roboeval}, but also on a comprehensive evaluation suite of 8 real-world tasks on the bimanual YAM setup (Sec.\ref{sec:exp-ft}). Our ablations demonstrate that \molmothink models yield further gains over \molmoactt while providing additional interpretability that aids both diagnosis and performance (Sec.\ref{depth}).

\molmoactt is fully open in every respect, and beyond that, is capable of supporting real-world deployment for practical tasks: we release the model weights, training code, and the complete training dataset. We aim for \molmoactt to be more than an academic robotics foundation model; we want it to be a model that can be deployed in real-world workflows and deliver meaningful social impact.

\section{\molmoer}
\label{sec:er}

\begin{table}[t]
\centering
\footnotesize
\setlength{\tabcolsep}{4pt}
\renewcommand{\arraystretch}{1.1}
\caption{\textbf{\molmoactt multimodal web data training corpus (combining data from \molmot, \molmoer, and Tulu-3).} Sizes denote the number of samples used in our mixture (we subsample several datasets to balance the mixture). Sampling weights are the marginal proportion of each group within the non-robotics portion of the pretraining mixture.
}
\label{tab:molmoact2-vqa-data}
{
\begin{tabular}{lcc!{\vrule}lcc}
 & \textbf{\#Samples} & \textbf{Weight} &  & \textbf{\#Samples} & \textbf{Weight} \\
\toprule
\multicolumn{2}{l}{\textbf{\molmot Dataset}} & & \multicolumn{2}{l}{\textbf{\molmoer Dataset}} \\
Image QA  & 2.4M  & 0.115 & Image Embodied QA  & 1.3M  & 0.11 \\
Video QA  & 2.4M  & 0.092 & Image Pointing  & 780K  & 0.11 \\
Image Pointing  & 1.1M  & 0.046 & Image Detection  & 100K  & 0.01 \\
Video Pointing  & 370K  & 0.069 & Video Embodied QA  & 703k  & 0.10 \\
Video Tracking  & 800K  & 0.069 & Multi-image/ Ego-Exo  & 700K  & 0.09 \\
Captions/Long QA  & 1.2M  & 0.069 & Abstract Reasoning  & 150K  & 0.04 \\
\textbf{Subtotal} & \textbf{8.3M} & \textbf{0.46} & \textbf{Subtotal} & \textbf{3.3M} & \textbf{0.46} \\
\toprule
\textbf{Tulu-3 SFT Dataset} & \textbf{980k}  & \textbf{ 0.08} & \textbf{Total} & \textbf{12.5M} & \textbf{1.0} \\
\end{tabular}%
}
\end{table}

Existing VLM backbones are optimized for semantic image understanding rather than the metric, geometric, and temporally grounded reasoning required for robot control.
We develop \molmoer as a strong VLM backbone for embodied reasoning (ER). 
We finetune \molmot~\citep{clark2026molmo2openweightsdata} for specialized embodied perception skills that downstream action reasoning depends on, including scene understanding, pixel-accurate pointing, multi-image and egocentric reasoning, exocentric correspondence, and video temporal reasoning. 
\molmoer outperforms every open-weight baseline as well as the strongest closed-source models, including Gemini Robot-ER 1.5 Thinking and GPT-5, on $9$ of $13$ established embodied reasoning benchmarks (\autoref{tab:er_results}), reaching an overall average of $63.8\%$ and improving over its \molmot starting point by $17$ points. The remainder of this section describes the new training data we introduce for \molmoer (summarized in \autoref{tab:molmoact2-vqa-data}) and the two-stage training recipe used to inject these skills.

\subsection{Training data}
\label{sec:molmoer-data}

On top of \molmot's original multimodal pre- and mid-training data~\citep{clark2026molmo2openweightsdata}, we curate a new embodied reasoning corpus of approximately $3.3$M samples spanning six complementary capability pillars: single-image embodied QA, image pointing, image detection, video embodied QA, multi-image and ego--exo reasoning, and abstract embodied reasoning. Each pillar is covered by two or three datasets with diverse supervision sources (simulator ground truth, 3D-annotated real scans, template-generated QA, and a small amount of LLM-generated chain-of-thought), so that the model is exposed to a wide distribution of spatial reasoning phenomena rather than over-fitting to a single template style. The composition of this corpus is summarized in \autoref{tab:molmoact2-vqa-data}; we briefly describe each constituent below.

\paragraph{Image embodied QA.}
We assemble a mixture of image QA sources chosen to cover complementary axes
of spatial competence rather than to maximize any single one.
SAT~\citep{ray2025satdynamicspatialaptitude} supplies simulator-grounded
supervision for \emph{dynamic} reasoning (egocentric motion, perspective
taking, action consequences) that is hard to harvest from static web data.
RoboPoint-QA~\citep{yuan2024robopoint} contributes general VQA breadth to
guard against forgetting base perception during spatial fine-tuning.
RefSpatial~\citep{zhou2025roborefer} bridges web
images, real indoor scans, and procedural simulation, and is our main source
of chain-of-thought referring---the bridge from spatial language to
the pointing actions used downstream; we draw 250K multiple choice questions, 250K Chian-of-thought, and 80K
pointing examples. VST-P~\citep{yang2025visualspatialtuning} normalizes
inputs onto a uniform virtual camera, giving metric-consistent depth,
distance, direction, and size supervision (200K single-image + 200K
cross-view). VSI-590K~\citep{yang2025cambriansspatialsupersensingvideo}
extends coverage to in-the-wild robotics and tour footage via 3D-grounded
label propagation (200K images and 300K videos). Together these span static/dynamic, synthetic/real, and single-/multi-view regimes.

\paragraph{Video embodied QA.}
To extend reasoning across time we pair two complementary sources.
SIMS-VSI~\citep{brown2025simsvsimulatedinstructiontuningspatial} gives clean
simulator labels for distance, direction, count, and temporal-order
questions over agent trajectories (203K). RoboVQA~\citep{sermanet2023robovqamultimodallonghorizonreasoning}
covers the other end of the distribution: human-annotated long-horizon
embodied video targeting planning, affordance, and future prediction---i.e.
the question types that map most directly onto policy
behavior; we use a 200K subset.

\paragraph{Pointing and object detection.}
Pointing is our primary action interface, so we deliberately oversample
pixel-accurate localization. Beyond RefSpatial's pointing split, we use the
full RoboPoint procedural pointing corpus (700K normalized $(x,y)$ targets
for object reference and free-space selection) together with its 100K
LVIS-sourced detection split~\citep{yuan2024robopoint}, which ground spatial
language directly in the coordinate format the policy
consumes.

\paragraph{Multi-image and ego\textendash exo correspondence.}
Embodied policies routinely reconcile multiple camera views and switch
between first- and third-person frames, an ability under-served by
single-image corpora. We therefore include SenseNova-SI~\citep{cai2026scalingspatialintelligencemultimodal}
(500K subset), whose distinguishing emphasis is multi-image and
egocentric--exocentric correspondence~\citep{grauman2024egoexo4dunderstandingskilledhuman},
together with the 200K cross-view split of VST-P.

\paragraph{Abstract embodied reasoning.}
Finally, we add two synthetic diagnostics to harden compositional reasoning
in a low-bias setting. CLEVR~\citep{johnson2016clevrdiagnosticdatasetcompositional}
(50K) targets compositional attribute--relation reasoning. GRiD-3D~\citep{lee_grid3d_2022}
(100K) specifically isolates \emph{object-intrinsic} relative direction
(front/left of a referent's own frame, not the camera's)---a
frame-of-reference distinction that matters for instruction following but is
rarely labeled in natural data.

\subsection{Specialize-then-rehearse training recipe}
\label{sec:molmoer-training}

To avoid the redundant compute of re-running \molmot's full multimodal training with our new corpus integrated, we build on the released \molmot checkpoint with a two-stage \emph{specialize-then-rehearse} recipe.

\paragraph{Stage 1: Embodied specialization.}
Starting from the \molmot--4B mid-training checkpoint~\citep{deitke2024molmo,qwen3}, we fine-tune for $20$K steps on the \molmoer corpus augmented with $8\%$ Tulu-3~\citep{lambert2025tulu3pushingfrontiers} text-only data to preserve language competence, using sequence length $4{,}200$ and a global batch size of $64$ (device batch size $4$ across $2$ nodes $\times$ $8$ H100 GPUs). This stage rapidly moves the model onto the embodied data manifold: pointing accuracy, video embodied QA, and multi-image reasoning all improve sharply.

\paragraph{Stage 2: Joint refinement.}
We continue training the Stage~1 checkpoint for $1.5$K additional steps on a mixture that interleaves our embodied corpus with \molmot's original multimodal mid-training data (general VQA, captioning, academic benchmarks, tracking, and \molmot pointing). Holding the NLP rate at $8\%$, the remaining $92\%$ budget is split as $p \cdot 0.92$ embodied and $(1-p) \cdot 0.92$ general, with each side's internal proportions preserved. Sweeping $p \in \{0.30, 0.50, 0.70, 0.90\}$, we find $p = 0.5$ yields the best Pareto trade-off between the embodied-reasoning benchmarks in \autoref{tab:er_results} and \molmot's general benchmarks. To accommodate the long multi-image and long-video examples in the general mixture, Stage~2 uses a longer sequence length ($16{,}384$ vs.~$4{,}200$ in Stage~1) with per-device batch size reduced to $1$; all other hyperparameters follow \molmot.

\section{Data}
\label{sec:data}

Training a generalist vision-language-action model requires data that is
simultaneously large in scale, diverse across embodiments, tasks, and scenes,
and high in quality. The robotics community has released several substantial
public corpora toward this goal---most notably the Open X-Embodiment (OXE)
mixture~\citep{o2024open}, DROID~\citep{khazatsky2024droid}, and a rapidly
growing collection of LeRobot community datasets contributed by
SO-10x users~\citep{shukor2025smolvlavisionlanguageactionmodelaffordable}. While each is valuable, none of these
resources individually, nor their union, is sufficient for training a model
intended for real-world deployment. OXE offers breadth across embodiments
but its constituent datasets vary widely in quality, control conventions,
and language-annotation fidelity; DROID provides scale on a single Franka
platform, but a substantial fraction of its episodes contain idle segments,
failed attempts, or repetitive task instructions; and crowd-sourced LeRobot
data, although rich in embodiment and scene diversity, frequently contains
placeholder annotations such as ``lerobot\_test'' alongside genuine
demonstrations~\citep{shukor2025smolvlavisionlanguageactionmodelaffordable}.
Critically, no public corpus contains high-quality bimanual manipulation
data on the platform we target for deployment, and the tasks that are
covered are often collected with limited variation in scene configuration,
object instances, or spatial variation that mimic the realistic of real-world tasks for deployment.

\begin{figure}[t]
    \centering
    \includegraphics[width=\linewidth]{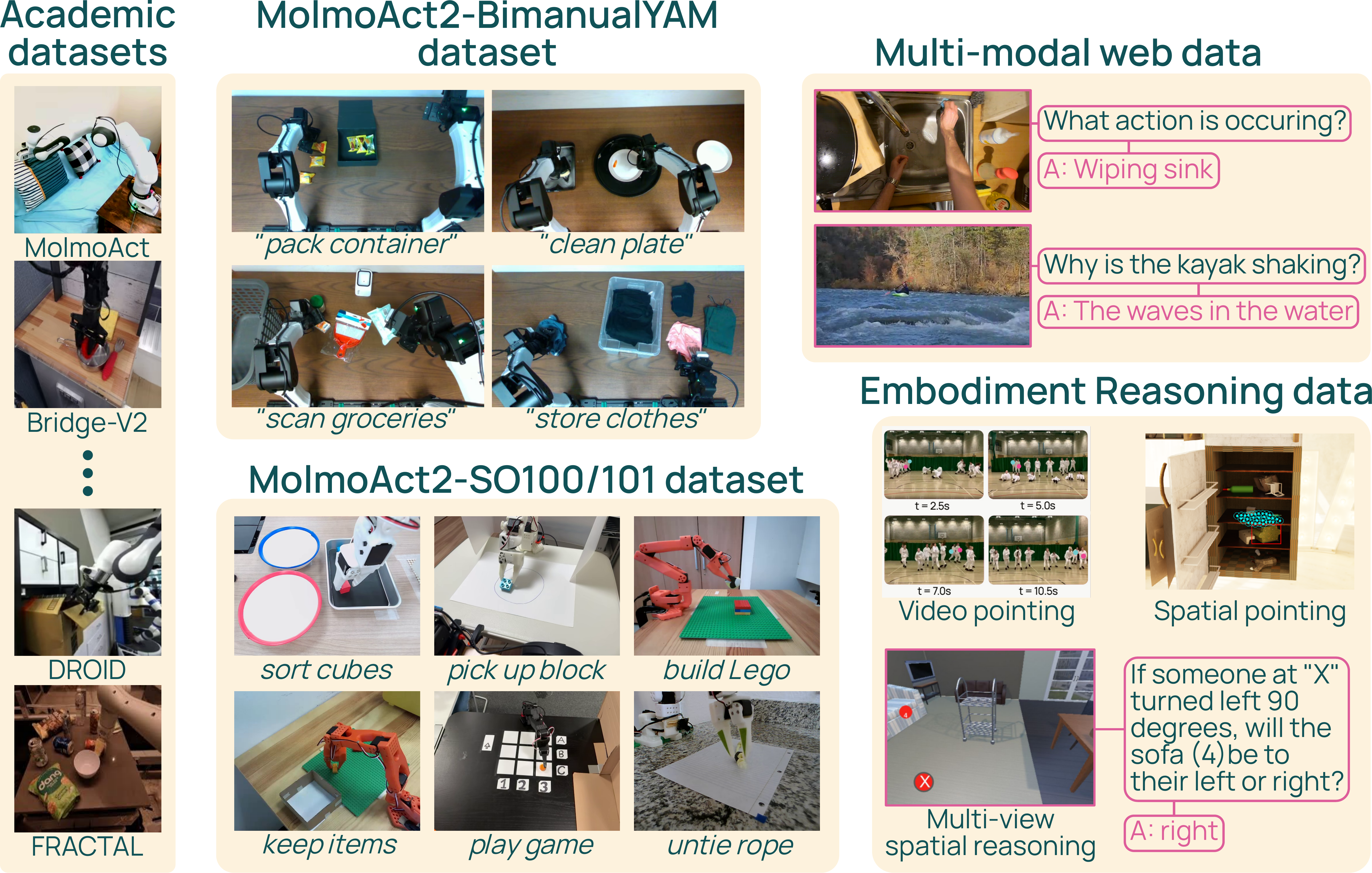}
    \caption{\textbf{Overview of the \molmoactt training data.} Our data mixture combines public academic robot datasets, \molmoacttyamdata, \molmoacttdroiddata, \molmoacttsodata, multimodal web data, and embodied-reasoning data.}
    \label{yamsetup}
\end{figure}

To close these gaps, we assemble \molmoactt's training mixture from three
complementary sources, summarized in Fig.~\ref{yamsetup}. First, we collect
\molmoacttyamdata, a new bimanual manipulation dataset emphasizing task
repeatability and task, object, and scene diversity across a wide range of
useful behaviors. Second, we curate and filter two large public corpora,
\molmoacttsodata (drawn from community LeRobot data) and \molmoacttdroiddata
(drawn from DROID), using structural, licensing, and quality-based
filtering pipelines, and re-annotate their language instructions to improve
both accuracy and diversity. Third, we co-train with a targeted subset of
academic robotics data for additional embodiment breadth, together with
multimodal and embodied-reasoning data to preserve the broad visual and
linguistic competence of the underlying VLM. The remainder of this section
describes each component of the mixture in turn, together with the language
re-annotation pipeline shared across our robot datasets.

\molmoactt is trained on a diverse set of datasets spanning robot data, multimodal reasoning, and embodied reasoning, summarized in Fig.~\ref{yamsetup}. Additionally, we collected \molmoacttyamdata, curated \molmoacttsodata, and filtered \molmoacttdroiddata for use in pre-training and post-training. Below, we describe each dataset and detail its respective collection, curation, or filtering process.

\subsection{\molmoacttyamdata}
\begin{figure}[t]
    \centering
    \includegraphics[width=0.6\linewidth]{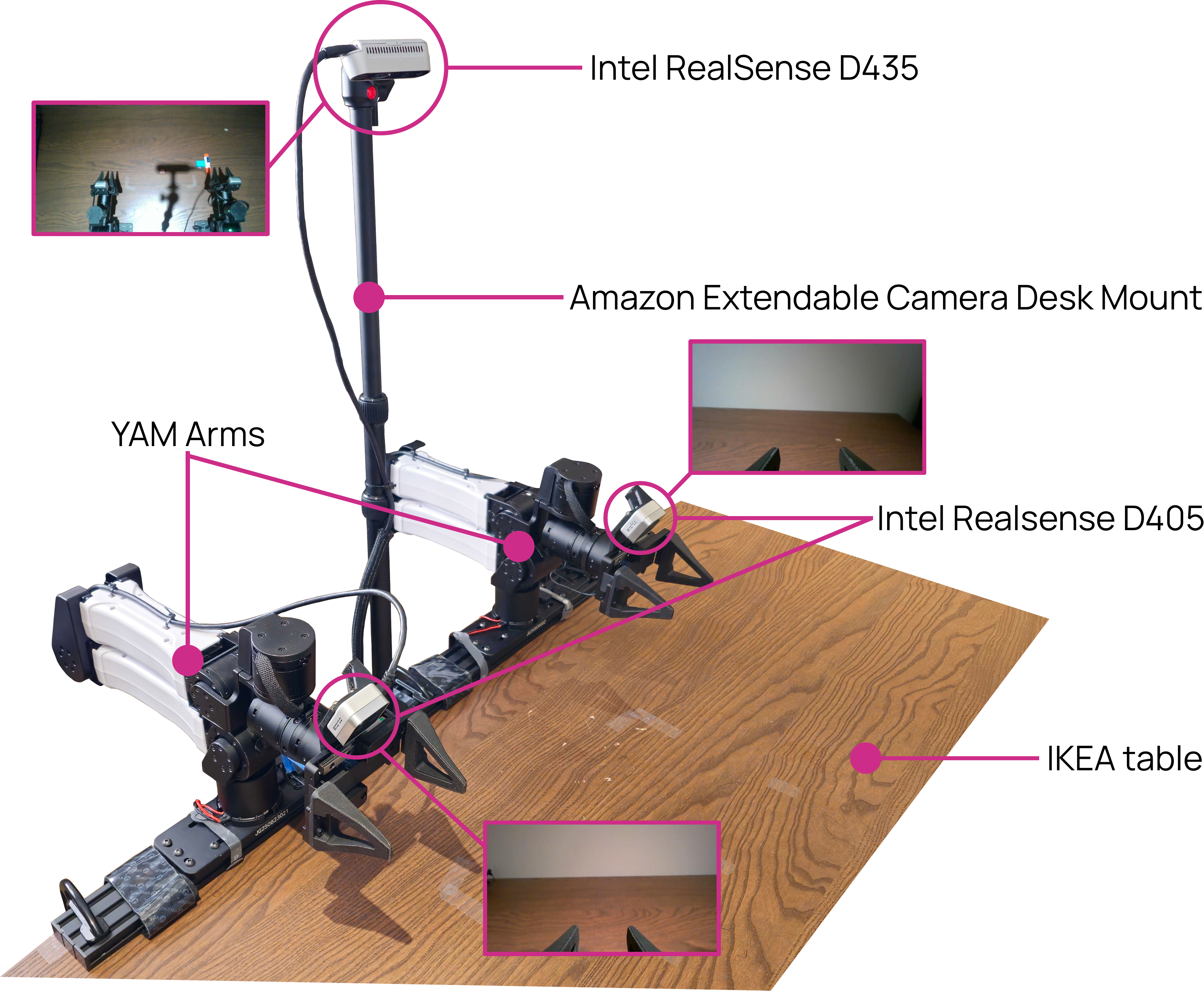}
    \caption{\textbf{\molmoacttyamdata collection setup.}
Our standardized \molmoacttyamdata collection setup. Every component is readily
available for purchase, and the total cost of the entire setup is under \$6{,}000 USD.}
    \label{fig:dataset-setup}
\end{figure}

To support real-world deployment, we introduce \molmoacttyamdata, an open-source robot manipulation dataset that emphasizes task repeatability, task diversity, and object diversity across a broad range of useful behaviors spanning household, factory, and coffee-shop settings. All data is collected on our custom bimanual YAM (Yet Another Manipulator) setup, shown in Figure~\ref{fig:dataset-setup}. \molmoacttyamdata is designed to deliver both scale and quality. It contains over 28 unique real-world tasks from folding clothes and untangling cables to bussing tables, scanning groceries, and packing medication, each captured with substantial variation in scene configuration, object instances, and object placement. In total, the dataset comprises 34.5k robot demonstrations totaling over 720 hours of robot data, collected over a two-month period. Data collection was supported by Cortex AI, with strict protocols governing the number of permitted failure retries and the maximum duration of no-op segments to ensure consistently high data quality. Further details on \molmoacttyamdata are provided in the appendix.

\subsection{\molmoacttsodata}
\label{sec:so100-dataset}
SO-100/101 is a low-cost robotic platform from Hugging Face that is accessible to a broad community of users. As a result, we utilize the diverse open-source SO-10x data collected by the community to improve our model's capability for deployment in the wild \citep{chen2025depi}. Specifically, we curate \molmoacttsodata from 1{,}222 public LeRobot datasets contributed by 377 users. This corpus contains 38{,}059 robot demonstration episodes, 19.8M frames, and approximately 184 hours of interaction data.

To prioritize quality while preserving diversity, we apply a four-stage filtering pipeline: (i) structural validity checks (required schema fields, valid action/state tensors, no NaN/corrupt samples), (ii) removal of eval-style datasets, (iii) license/codebase eligibility checks, and (iv) a final TOPReward quality gate ~\citep{chen2026topreward}. In stage (iv), we keep datasets whose mean TOPReward over the last 3 sampled episodes is above a threshold obtained by averaging the TOPReward over a collection of human-audited high-quality datasets.

The filtered data spans both SO-100 and SO-101 embodiments, multiple camera configurations, varied object manipulation tasks, and diverse real-world collection environments. Compared with centrally collected robot datasets, this community-sourced corpus provides broader coverage of user setups, backgrounds, objects, and task annotations, making it a useful source of embodiment and environment diversity for improving real-world robustness.

\subsection{\molmoacttdroiddata}
DROID (Distributed Robot Interaction Dataset)~\citep{khazatsky2024droid} is a large-scale in-the-wild robot manipulation dataset, collected across a wide range of real-world deployment scenarios with a unified Franka robot setup. To ensure the quality of our training data, we leverage the supplementary annotations released in the accompanying HuggingFace repository~\citep{pertsch2024droidannotations} to filter the original DROID release. Specifically, we use (i) the extended language annotations, which provide three natural-language instructions for 95\% of the 75k successful episodes, and (ii) the provided idle-frame filter, which retains only contiguous non-idle action segments of at least one second. Furthermore, we did language re-annotation for the filtered DROID dataset before using for training. The resulting subset, which we refer to as \molmoacttdroiddata, contains 74,604 valid episodes comprising a total of 17{,}758{,}044 frames. Each episode in this subset is marked as successful, contains at least one valid language instruction, and is free of significant pauses.

\subsection{Language annotation pipeline}
\label{sec:lang_annot}
Following standard practice in imitation learning, each demonstration in our datasets is paired with a language instruction that describes the task completed by the teleoperator. However, we find that these annotations often are imperfect along two key axes. First, datasets where a specific set of tasks is collected repeatedly at scale often have repetitive language instructions. For example, the dataset collected in \cite{jang2022bc} contains 104 unique instructions for 39350 episodes, representing $0.26\%$ unique instructions in the dataset. Secondly, as found in prior work \citep{shukor2025smolvlavisionlanguageactionmodelaffordable}, crowd-sourced datasets like the \molmoacttsodata often have inaccurate or meaningless task annotations, like "lerobot\_test" and "Test run".

To improve both the diversity and accuracy of the language instructions in our robotics datasets, we use an open-source VLM (Qwen3.5-27B) to re-annotate the datasets. We prompt the VLM with a sample of the frames and the original instruction. The VLM is asked to generate a instruction that describes the demonstration. To increase the diversity of prompts, we randomly sample a number and ask the VLM to make this instruction roughly that many words. Relabeling the robotics datasets through this pipeline doubles the unique labels in the overall from 71121 (22\%) to 146485 (46\%) in dataset. More details and the prompt can be found in Appendix \ref{supp:language-annotation-details}.

\subsection{Academic robotics datasets}
To broaden the range of embodiments, camera layouts, task families, and control conventions seen during pre-training, we include additional public academic robotics data. This pool consists of a targeted subset of the Open X-Embodiment mixture~\citep{o2024open}, including BC-Z~\citep{jang2022bc}, BridgeData V2~\citep{walke2023bridgedata},
and RT-1~\citep{brohan2022rt}, together with \molmoactdata from \molmoact~\citep{lee2025molmoactactionreasoningmodels}. These datasets are used as complementary sources rather than as the primary deployment embodiments, since YAM, SO-100/101, and DROID supply the majority of robot training examples.

\subsection{Multimodal datasets}
Past research on robotics foundation models has found benefits in incorporating multimodal data into the training mix along with robotics datasets \citep{lee2025molmoactactionreasoningmodels, intelligence2025pi05visionlanguageactionmodelopenworld}. We co-train the VLA with a multimodal data mixture where 46\% consists of the \molmoer dataset mixture described in Sec.~\ref{sec:molmoer-data}, 46\% consists of the \molmot dataset mixture~\citep{clark2026molmo2openweightsdata}, and 8\% consists of the text-only data from Tulu-3 ~\citep{lambert2025tulu3pushingfrontiers}. The high level composition of this data can be seen in Table \ref{tab:molmoact2-vqa-data}.


\section{\molmoactt}
\label{sec:molmoact2}

\begin{figure}[t]
\centering
\includegraphics[width=\linewidth]{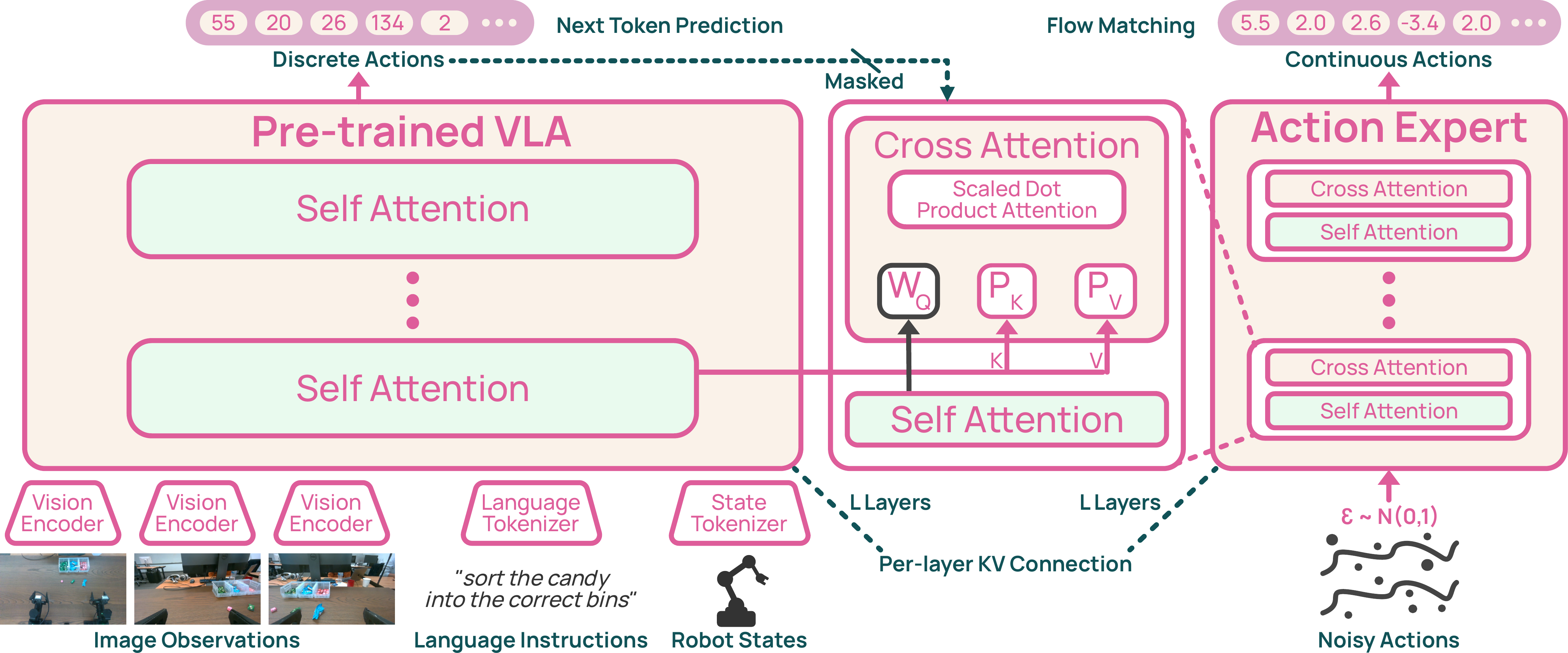}
\caption{\textbf{Overview of \molmoactt.} Image observations, language
instructions, and robot states are tokenized and processed by a pre-trained
VLA backbone with self-attention layers. Post-training attaches a DiT-style
action expert with the same number of layers as the backbone. At each layer,
the backbone key and value tensors are projected and reused as the key and
value inputs to the action expert's cross-attention, enabling per-layer KV
conditioning from visual-language context to continuous control. The expert is
trained with flow matching to denoise a noisy action trajectory into a
continuous robot trajectory. During training, the backbone is also supervised
with next-token prediction over discrete action tokens, while the target
action-token span is masked from the expert so continuous action prediction
cannot condition on the ground-truth discrete actions.}
\label{fig:overview}
\end{figure}

\molmoactt follows a three-stage training pipeline, with the post-training architecture summarized in Fig.~\ref{fig:overview}. Pre-training~(Sec.~\ref{sec:pretraining}) adapts the Molmo2-ER vision-language backbone into a discrete autoregressive robot policy, using the \molmoactfast~(Sec.~\ref{sec:openfast}) to map continuous trajectories into a compact action vocabulary under a unified next-token recipe~(Sec.~\ref{sec:pretraining-recipe}). Post-training~(Sec.~\ref{sec:posttraining}) attaches a flow-matching action expert with per-layer KV conditioning on the autoregressive backbone~(Sec.~\ref{sec:posttraining-action-expert}), and co-trains discrete and continuous action supervision~(Sec.~\ref{sec:posttraining-recipe}) to produce continuous control. Deployment~(Sec.~\ref{sec:deployment}) then covers embodiment-specific fine-tuning~(Sec.~\ref{sec:finetuning}) and inference optimization~(Sec.~\ref{sec:inference}). Shared implementation details for training infrastructure, packing, prompt formatting, and stage-level hyperparameters are provided in \autoref{supp:training}.

\molmoactt is designed around a practical tension: we want to preserve the scaling properties and general visual-language competence of a pretrained VLM, while producing precise continuous robot actions across heterogeneous embodiments. Directly training a VLM and a continuous action expert together from the beginning makes optimization unnecessarily difficult: the model must simultaneously learn a robot-aware token interface, align new state/action embeddings, and fit a flow-matching controller. We therefore use a three-stage pipeline. Pre-training first turns \molmoer into an action-aware VLA using only autoregressive supervision, which gives the backbone aligned robot, state, and action-token embeddings under the same next-token objective used by the base VLM. Post-training then attaches the continuous action expert to this already action-aware backbone, allowing the flow-matching controller and its VLM conditioning interface to align quickly on a broad robot-data mixture. We keep post-training separate from embodiment-specific fine-tuning for efficiency: the post-training mixture is intentionally diverse across datasets, robots, control rates, camera layouts, and task families, so extending it until the model is directly deployment-ready for every target embodiment would require a long and expensive training stage. Instead, post-training produces a general continuous-control base model, and fine-tuning adapts that model quickly to a specific embodiment, environment, and use case while keeping the same architecture and output interfaces.

\subsection{Pre-training}
\label{sec:pretraining}

\molmoacttpre adapts the \molmoer vision-language backbone into a discrete autoregressive robot policy while keeping the Molmo2 token interface intact. Images and video frames are still encoded by a ViT, pooled and projected by the vision-language connector, and passed to the LLM together with text. Robot examples extend this sequence with two additional token streams: state tokens that describe the current robot configuration, and action tokens that describe the future one-second motion.

This subsection focuses on the two ingredients needed to make that formulation practical. First, we describe \molmoactfast, which converts continuous, embodiment-specific robot trajectories into compact discrete action tokens. Second, we describe the pre-training recipe that lets these robot sequences be mixed with standard multimodal data: single-crop visual inputs, a small number of sampled video frames, setup/control and action-output markers, state tokenization, and packed training sequences. Together, these choices preserve a single next-token prediction objective across text, vision-language, state, and action targets, without introducing a separate continuous action head at this stage.

\subsubsection{\molmoactfast}
\label{sec:openfast}

Robot actions are continuous, embodiment-specific, and often produced at different control rates, so they cannot be inserted into a language-model pre-training stream directly. We therefore train \molmoactfast, an open-weight and open-data action tokenizer following FAST~\citep{pertsch2025fast}. The goal is not to introduce a different tokenization principle, but to make this component fully inspectable and reproducible: existing FAST weights are useful, but their training data mixture is not fully specified. \molmoactfast fills this transparency gap by releasing both the tokenizer weights and the action mixture used to train it. It maps a one-second action trajectory into a compact sequence of discrete tokens by representing the trajectory with a frequency-domain transform, quantizing the resulting coefficients, and applying byte-pair encoding to produce tokens from a 2048-token action vocabulary.

Unlike the prior FAST tokenizer, whose released weights are not paired with a fully specified training distribution, \molmoactfast is trained from a transparent action mixture. We subsample one million action sequences across five embodiments, balancing the main \molmoactt deployment platforms with smaller sources that broaden the control vocabulary (\autoref{tab:fast-data}). This gives \molmoactfast coverage over bimanual YAM, SO-100/101, DROID Franka, BC-Z \citep{jang2022bc}, BridgeData V2 \citep{walke2023bridgedata}, RT-1 \citep{brohan2022rt}, and \molmoactdata trajectories, including both absolute joint control and delta end-effector control.

Before fitting the tokenizer, we put all raw robot telemetry into a common action format. Each training sequence corresponds to one second of robot motion, so the number of actions in a chunk is set by the control frequency of the source dataset. Each action in the chunk is padded to 32 dimensions, so embodiments with different action dimensionalities share the same tokenizer input space. Continuous dimensions are normalized with 1--99 percentile statistics, which limits the effect of outliers while preserving the useful dynamic range of each control dimension. Gripper commands are handled separately from this continuous normalization, since they are typically binary or narrow-range open/close signals. This standardized representation allows the same tokenizer to cover both joint-space and end-effector control across multiple robot platforms.

\begin{table}[t]
\centering
\small
\setlength{\tabcolsep}{5pt}
\renewcommand{\arraystretch}{1.05}
\caption{\textbf{Embodiment mixture for \molmoactfast training.} The tokenizer is trained on one million subsampled action sequences spanning the main robot embodiments used by \molmoactt, plus smaller sources that broaden control-mode coverage.}
\label{tab:fast-data}
\begin{tabular}{lcll}
\textbf{Dataset} & \textbf{Mix} & \textbf{Robot} & \textbf{Action representation} \\
\midrule
\molmoacttyamdata & 30\% & YAM & Absolute joint \\
\molmoacttsodata & 30\% & SO-100/101 & Absolute joint \\
\molmoacttdroiddata & 30\% & Franka & Absolute joint \\
Fractal~\citep{brohan2022rt} & 3.33\% & Google Robot & Delta end-effector \\
BC-Z~\citep{jang2022bc} & 3.33\% & Google Robot & Delta end-effector \\
Bridge~\citep{walke2023bridgedata} & 3.33\% & WidowX & Delta end-effector \\
\end{tabular}
\end{table}

\subsubsection{Pre-training recipe}
\label{sec:pretraining-recipe}

\paragraph{Model.}
We initialize from the \molmoer checkpoint and keep the Molmo2 vision-language architecture~\citep{clark2026molmo2openweightsdata}. Visual observations are encoded by the SigLIP2 ViT~\citep{tschannen2025siglip}, converted into language-model tokens by the connector, and passed to the LLM together with the language instruction and robot-specific text. The connector uses the same Molmo2 design: features from the third-to-last and ninth-from-last ViT layers are pooled into compact image or video-frame tokens and projected into the LLM embedding space. Full backbone and added-token details are given in \autoref{supp:model-backbone}, with architecture hyperparameters in \autoref{tab:hyperparams_model}.

\paragraph{Data.}
The training mixture combines multimodal examples with robot trajectories so that the model retains vision-language capability while learning to predict actions; we describe the data sources and curation in Sec.~\ref{sec:data}. We allocate 10\% of sampling to multimodal data and 90\% to robot trajectories. Within the robot portion, YAM, SO-100/101, and DROID each receive 30\% of the robot sampling weight; the remaining 10\% is split across smaller BC-Z, BridgeData V2, RT-1, and \molmoactdata sources. For all visual inputs in this stage, we use a single resized crop rather than high-resolution tiled crops. For videos, we sample at most 8 frames at up to 2 FPS, and we skip examples that exceed the sequence budget, which primarily removes unusually long text examples and videos with long captions. We also apply the image augmentation described in \autoref{supp:training} during pre-training, combining light geometric perturbations with color jitter and occasional blur. For multi-camera robot episodes, we randomize the input camera order at the episode level to prevent the model from relying on a fixed camera slot. Each robot example also includes setup and control strings using the same special-token wrappers used during training, e.g., \texttt{\detokenize{<setup_start>bimanual yam robotic arms in molmoact2<setup_end>}}, and \texttt{\detokenize{<control_start>absolute joint pose<control_end>}} or \texttt{\detokenize{<control_start>delta end-effector pose<control_end>}}; the corresponding chat template and output trigger formatting are detailed in \autoref{supp:training}.

\paragraph{Action and state representation.}
This stage is deliberately limited to discrete autoregressive supervision: the model predicts action tokens, but does not yet train the continuous action expert used in post-training. Each robot target is represented as a one-second action chunk, so the number of raw actions in the chunk follows the control frequency of the source dataset. Before tokenization, continuous action and state dimensions are normalized with 1--99 percentile statistics; gripper commands are treated separately from this percentile scaling when they are represented as binary or narrow-range open/close signals. Action vectors are padded to a 32-dimensional space and then encoded with the 2048-token vocabulary of \molmoactfast. Proprioceptive state is represented separately: after normalization, each state value is uniformly discretized into one of 256 state tokens and appended to the prompt before the action target.

\paragraph{Training.}
We train for 200K steps with a maximum sequence length of 4200 tokens. Examples vary substantially in length: a text-only example may use only a few hundred tokens, while multi-camera robot examples must carry images, states, and action targets. We therefore use on-the-fly packing to combine multiple short examples into one 4200-token sequence. Packed examples share the same forward pass, but their text, visual tokens, state tokens, and action targets remain separated by the training attention mask, so the model does not condition one example on another. Appendix~\ref{supp:training} gives the implementation details for this packing procedure and the shared training stack. We train the vision encoder, connector, language model, and newly added tokens' embeddings with a global batch size of 128 across 64 H100 GPUs for around 5,760 GPU hours. The vision encoder and connector use a learning rate of \(5\times10^{-6}\), and the language model uses \(1\times10^{-5}\). This produces a discrete VLA checkpoint that can later be adapted to continuous control.

\subsection{Post-training}
\label{sec:posttraining}

\molmoacttpre learns a robot policy through the same autoregressive interface used by the VLM: given language, visual observations, setup/control descriptors, and state tokens, it predicts discrete action tokens produced by \molmoactfast. This makes large-scale robot pre-training simple and stable, but the deployed policy should output continuous action trajectories directly. Post-training therefore adds a flow-matching action expert to the pre-trained VLM, producing the final \molmoactt model.

This section describes the two parts of that adaptation. First, we introduce the action expert and its per-layer KV conditioning on the VLM, which lets the continuous controller condition on the full visual-language attention state without replacing the backbone. Second, we describe the post-training recipe: how we keep the pre-training data construction, co-train discrete and continuous action supervision, mask padded or target-leaking tokens, and adjust packing and sequence lengths for the added action-expert compute. Additional model hyperparameters and implementation details are given in \autoref{supp:model-action-expert}, with shared training-system details in \autoref{supp:training}.

\subsubsection{Action expert and per-layer KV conditioning}
\label{sec:posttraining-action-expert}

Post-training adds a continuous action expert on top of the pre-trained autoregressive VLM. We use a DiT-style transformer expert because flow matching and diffusion models have shown that denoising transformers scale well for continuous generation, and because action trajectories are naturally represented as short continuous sequences rather than as text-like token streams. This choice separates responsibilities: the VLM provides visual-language grounding and robot-state context, while the expert models the continuous velocity field needed to transform noisy action chunks into executable trajectories.

The key architectural question is how this expert should receive VLM context. A shallow projection of the final hidden state compresses the backbone into a single residual-stream representation. Instead, \molmoactt conditions the expert on the VLM at every layer. Each action-expert block cross-attends to the corresponding VLM layer's keys and values, after lightweight learned projections map the VLM attention state into the expert's cross-attention width. This gives the continuous controller access to the same attention state used by the VLM itself, while preserving a modular interface between the backbone computation and the newly trained action expert.

Given a normalized target action chunk \(a\), Gaussian noise \(\epsilon\), and a sampled time \(t \in [0,1]\), we interpolate between noise and data,
\begin{equation}
    x_t = (1-t)\epsilon + t a,
    \qquad
    u^\star = a - \epsilon.
\end{equation}
The expert \(f_\theta\) predicts the target velocity \(u^\star\) from the noisy action chunk, the time embedding, and the VLM context \(c\), which contains the task, visual observations, setup/control descriptors, and discrete state tokens:
\begin{equation}
    \mathcal{L}_{\mathrm{flow}}
    =
    \mathbb{E}_{a,\epsilon,t}
    \left[
    \left\|
    m \odot \left(f_\theta(x_t,t,c) - u^\star\right)
    \right\|_2^2
    \right],
\end{equation}
where \(m\) masks padded action steps and padded action dimensions. At inference, we start from Gaussian noise and integrate the predicted velocity field to produce a continuous action trajectory.

Architecturally, the expert has the same depth as the VLM, where both of them use \(L=36\) layers. The expert first embeds the current noisy action sequence. Each block then applies action self-attention, cross-attention to the VLM, and an MLP, with the time embedding producing DiT-style shift, scale, and gate parameters for all three residual branches. Schematically, block \(\ell\) computes
\begin{align}
    h'_\ell
        &= h_\ell + g^{\mathrm{sa}}_\ell
        \operatorname{SA}\!\left(\operatorname{AdaRMS}^{\mathrm{sa}}_\ell(h_\ell,t)\right), \\
    \bar{h}_\ell
        &= h'_\ell + g^{\mathrm{ca}}_\ell
        \operatorname{CA}\!\left(
            \operatorname{AdaRMS}^{\mathrm{ca}}_\ell(h'_\ell,t),
            \tilde{K}_\ell,
            \tilde{V}_\ell
        \right), \\
    h_{\ell+1}
        &= \bar{h}_\ell + g^{\mathrm{ff}}_\ell
        \operatorname{MLP}\!\left(\operatorname{AdaRMS}^{\mathrm{ff}}_\ell(\bar{h}_\ell,t)\right).
\end{align}

The action expert uses per-layer KV conditioning rather than hidden-state conditioning. For each VLM layer \(\ell\), we collect the keys and values \((K^{\mathrm{vlm}}_\ell,V^{\mathrm{vlm}}_\ell)\) produced by that layer's self-attention. We then project them into the action-expert width with learned adapter projections \(P_K\) and \(P_V\):
\begin{equation}
    \tilde{K}_\ell =
    \operatorname{reshape}\!\left(P_K K^{\mathrm{vlm}}_\ell\right),
    \qquad
    \tilde{V}_\ell =
    \operatorname{reshape}\!\left(P_V V^{\mathrm{vlm}}_\ell\right).
\end{equation}
Here, \(P_K\) and \(P_V\) are linear VLM-to-expert adapter layers that align the VLM KV dimensionality with the expert's cross-attention width; they are separate from the VLM self-attention projections that produced the keys and values. After projection, the conditioning context is organized into the expert's attention heads. The cross-attention in expert block \(\ell\) attends to the projected keys and values from the corresponding VLM layer:
\begin{equation}
    \operatorname{CA}(Q_\ell,\tilde{K}_\ell,\tilde{V}_\ell)
    =
    \operatorname{softmax}\!\left(
        \frac{Q_\ell \tilde{K}_\ell^\top}{\sqrt{d_h}}
    \right)\tilde{V}_\ell,
\end{equation}
where \(d_h\) is the expert head dimension. This per-layer KV conditioning gives every action-expert block direct access to the visual-language attention state at the same depth in the backbone. During post-training we detach this conditioning path from the VLM, so the flow loss trains the expert and its adapter projections without sending gradients back through the VLM keys and values.

\subsubsection{Post-training recipe}
\label{sec:posttraining-recipe}

\paragraph{Overview.}
Post-training starts from the 200K-step \molmoacttpre checkpoint and turns it into the final \molmoactt model. We keep the VLM architecture and pre-training data construction from Sec.~\ref{sec:pretraining-recipe}, including the image augmentation described in \autoref{supp:training}. The main change is that robot examples now supervise both output interfaces: the LLM continues to predict discrete action tokens, while the action expert described in Sec.~\ref{sec:posttraining-action-expert} learns to generate continuous action chunks.

\paragraph{Multiple flow samples.}
For each robot action chunk, we evaluate the flow objective at multiple noise levels. Let \(a\) be a normalized continuous action chunk and \(c\) be the VLM context for the corresponding robot prompt. For each training example we draw \(K\) independent pairs \(\{(\epsilon_i,t_i)\}_{i=1}^K\), form \(x_{t_i}=(1-t_i)\epsilon_i+t_i a\), and train the expert to predict \(a-\epsilon_i\):
\begin{equation}
    \mathcal{L}_{\mathrm{flow}}(a,c)
    =
    \frac{1}{K}
    \sum_{i=1}^{K}
    \left\|
    m \odot
    \left(
    f_\theta(x_{t_i},t_i,c) - (a-\epsilon_i)
    \right)
    \right\|_2^2 .
\end{equation}
We use \(K=4\), so each robot sample contributes four points on the same flow trajectory while reusing the same visual-language context. We didn't use \(K=8\) as in the fine-tuning stage (Sec.~\ref{sec:finetuning}) mainly due to GPU memory constraints.

\paragraph{Padding and packing.}
Post-training retains a common action tensor shape across embodiments. Continuous actions are right-padded to a maximum horizon of 30 steps and a maximum width of 32 dimensions. The horizon mask removes padded time steps from the flow loss, and the dimension mask zeroes padded action dimensions in the noisy inputs, targets, and predictions before averaging the loss over valid dimensions. This lets datasets with different control rates and action widths share the same expert without training on artificial padded coordinates.

Packing is also extended to the continuous expert. A packed language sequence can contain multiple robot examples; each action chunk is paired with the sub-example that produced it, and the expert attends only to that example's VLM context. To keep memory use stable when packed batches contain different numbers of robot chunks, we pad the packed action-chunk axis up to a small fixed cap of five chunks per packed sequence when possible. Padded chunk rows are marked invalid and excluded from the flow loss; sequences with more chunks are still used, but processed without this fixed-cap padding.

\paragraph{Training objectives.}
The post-training objective combines the autoregressive loss from pre-training with the flow-matching loss:
\begin{equation}
    \mathcal{L}_{\mathrm{post}}
    =
    \mathcal{L}_{\mathrm{LM}}
    +
    \mathcal{L}_{\mathrm{flow}} .
\end{equation}
The language-model loss is applied to the same next-token targets used in pre-training, including discrete action tokens for robot examples and text tokens for VLM examples. The flow loss is applied only to continuous robot action chunks. Since robot examples contain both the discrete action target and the continuous action target, we mask the discrete action-token span out of the expert's VLM conditioning path. The expert therefore conditions on the task, observations, setup/control descriptors, and state tokens, but not on the target action tokens it is supposed to predict.

We also apply knowledge insulation \citep{driess2025knowledgeinsulation}, where the VLM is isolated from the continuous-action loss. The expert conditions on the VLM keys and values, but these tensors are detached before they enter the expert, so \(\mathcal{L}_{\mathrm{flow}}\) updates the action expert and its adapter projections without back-propagating through the VLM. The VLM is still updated by \(\mathcal{L}_{\mathrm{LM}}\).

\paragraph{Training.}
Robot batches use a sequence length of 2100 tokens, while the non-robot VLM batches keep the 4200-token context length from pre-training. We use this split because robot batches additionally run the action expert and four flow samples per action chunk, whereas VLM batches only train the autoregressive model. We train for 100K updates with a global batch size of 128 and a device batch size of 2 across 64 H100 GPUs for around 2,304 GPU hours. The VLM learning rates match pre-training: \(5\times10^{-6}\) for the vision encoder and connector, and \(1\times10^{-5}\) for the language model. The action expert is trained with a larger learning rate of \(5\times10^{-5}\).

\subsection{Deployment}
\label{sec:deployment}

\subsubsection{Embodiment-specific fine-tuning}
\label{sec:finetuning}

\paragraph{Shared recipe.}
Embodiment-specific fine-tuning starts from the post-trained \molmoacttpos checkpoint and keeps the same VLM--action-expert architecture described in Sec.~\ref{sec:posttraining}. We continue to co-train the discrete autoregressive action output and the continuous flow-matching action expert, use the same 32-dimensional padded action width, and keep the same learning rates as post-training: \(5\times10^{-6}\) for the vision encoder and connector, \(1\times10^{-5}\) for the language model, and \(5\times10^{-5}\) for the action expert. We also keep single resized crops, packed sequences, setup/control descriptors, discrete state tokens, and the same action-token vocabulary, using the shared prompt, image-augmentation, and packing implementation described in \autoref{supp:training}.

The fine-tuning stage differs from post-training in four ways. First, each run is robot-only: we do not include the multimodal VLM mixture or a separate VLM dataloader. Second, we increase the number of sampled flow times from 4 to 8 per action chunk, which gives the action expert denser supervision along the same flow trajectory. Third, we do not use knowledge insulation during fine-tuning: gradients from the flow loss are allowed to update the VLM through the action-expert conditioning path, since we did not observe a consistent performance gain from detaching this path at this stage. Finally, for the main embodiment checkpoints below, we do not tune the added-token input embeddings. As in standard VLM fine-tuning, we assume that the token embeddings learned during pre-training and post-training are already well placed, and tune the language-model output head and final normalization instead.

\paragraph{Bimanual YAM.}
The \molmoacttyam checkpoint is fine-tuned on the bimanual YAM mixture. The camera order is fixed as top, left, and right, matching the data collection setup. We use annotated language instructions, absolute joint-pose actions, and a 30-step action chunk, corresponding to the 30 Hz control rate of the dataset. We train with a sequence length of 2100, global batch size 128, and 100K updates on 64 H100 GPUs, for roughly 2,304 GPU hours.

\paragraph{DROID.}
The \molmoacttdroid checkpoint is fine-tuned on the filtered DROID mixture. DROID provides two exterior cameras and one wrist camera. For DROID-style two-view evaluation, we use a fixed exterior-then-wrist ordering; when training variants expose both exterior choices as alternatives, the loader samples one exterior camera and pairs it with the wrist camera during training. We use absolute joint-pose actions with a 15-step action chunk, matching DROID's 15 Hz control rate. We do not use the additional language annotations in this fine-tune, to keep the comparison with prior DROID-trained models fair. We train with a sequence length of 2100, global batch size 64, and 100K updates on 32 H100 GPUs, for roughly 1,152 GPU hours.

\paragraph{SO-100/101.}
The \molmoacttso checkpoint is fine-tuned on the SO-100/101 mixture. Because this data is aggregated from internet demonstrations with diverse and inconsistent camera layouts, we randomize the input camera order at the episode level rather than imposing a fixed view naming convention. We use annotated language instructions, absolute joint-pose actions, and a 30-step action chunk for the 30 Hz control rate. We train with a sequence length of 2100, global batch size 64, and 100K updates on 32 H100 GPUs, for roughly 1,152 GPU hours.

\paragraph{LIBERO.}
The \molmoacttlib checkpoint is fine-tuned on the full LIBERO training mixture, combining the Spatial, Object, Goal, and Long suites rather than training a separate model for each suite. The camera order is fixed as front view followed by wrist view. We do not use annotated language instructions. LIBERO uses relative end-effector control at 10 Hz, so we train with a 10-step action chunk. We use a sequence length of 2100, global batch size 64, and train for 50K updates on 32 H100 GPUs, for roughly 1,152 GPU hours; for evaluation, we select the best-performing checkpoint, which occurs at 40K updates.

\paragraph{Other evaluation fine-tunes.}
For smaller task- or benchmark-specific fine-tunes used in downstream evaluation, we follow the same recipe unless otherwise noted. These runs use fixed metadata camera order rather than camera-order randomization, no language annotations, 2100-token sequences, packing, 8 sampled flow times, and robot-only data. The action horizon is set to the dataset control frequency, and the control representation follows the dataset, either joint pose or end-effector pose. Real-world evaluation runs use 8 H100 GPUs, global batch size 16, and 50K updates, and we choose the best-performing checkpoint for evaluation.

\subsubsection{Inference optimization}
\label{sec:inference}
For the \molmoactt model, inference is dominated by the continuous action expert, which integrates a fixed-step flow-matching trajectory conditioned on the VLM context. Within one action chunk, this context is invariant across flow steps, while only the noisy action state and flow time evolve. We therefore cache reusable action-expert intermediates, including context-dependent cross-attention states and fixed position-dependent terms, and reuse them throughout the flow loop. We further capture the fixed-shape flow loop with CUDA Graphs, reducing Python and kernel-launch overhead during repeated action generation.

\section{\molmothink}
\label{sec:reasoning}

\begin{figure}[t]
    \centering
    \includegraphics[width=0.6\linewidth]{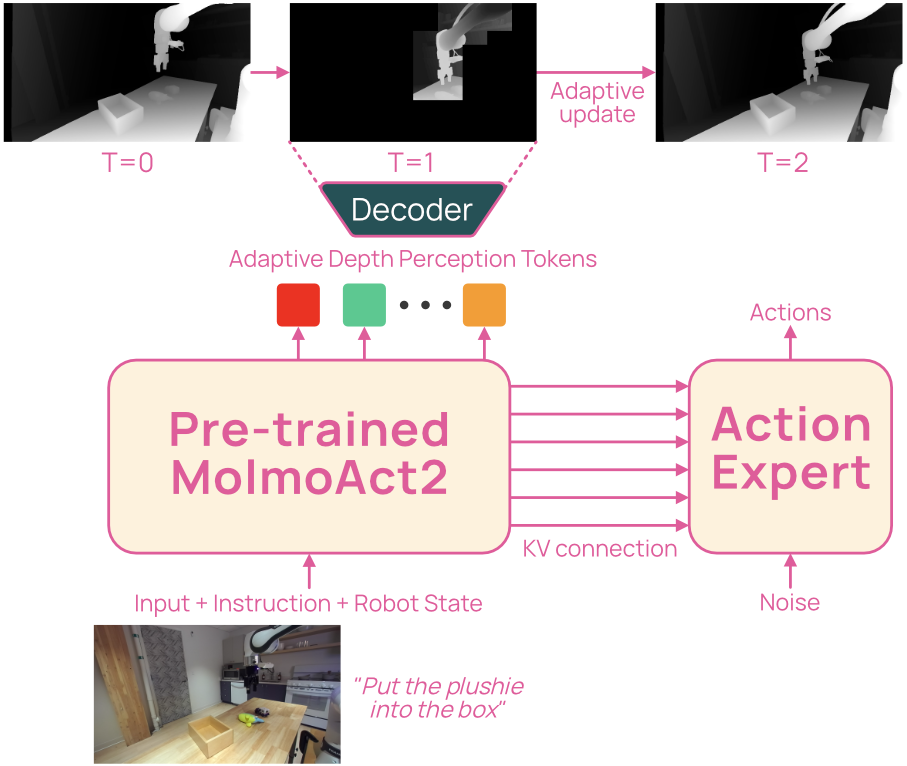}
    \caption{\textbf{Overview of \molmothink.} 
\molmothink augments the \molmoactt action-generation pipeline with adaptive 
depth-token reasoning, reusing cached depth codes for static regions and regenerating 
depth codes for changed regions before conditioning the action expert.}
    \label{fig:think-adaptive-depth}
\end{figure}

Robotic manipulation depends on spatial information that is only indirectly
supervised by action imitation: object distance, free space, occlusion, and
surface layout all affect the action, but standard behavior-cloning objectives
do not ask the model to make this structure explicit before acting. \molmoact
\citep{lee2025molmoactactionreasoningmodels} addressed this problem by adding
depth-token prediction as an intermediate reasoning step. \molmothink extends
the same idea to \molmoactt: before producing an action, the model predicts a
compact discrete depth representation that conditions the action expert through
per-layer KV conditioning.

The depth representation is intentionally lightweight. Each observation depth
map is quantized into a \(10 \times 10\) grid, giving 100 spatial code
positions, and each position takes one of 128 learned depth-code values. These
codes are represented as ordinary autoregressive tokens, which makes depth reasoning compatible with the same next-token interface used throughout \molmoactt while exposing an interpretable intermediate prediction.

The main distinction from \molmoact is that \molmothink makes depth prediction
adaptive across time. Robot trajectories contain substantial temporal
redundancy: many cells in a scene-level depth grid remain unchanged from one
control step to the next. Instead of re-predicting every depth code at every
step, \molmothink reuses cached codes for static regions and autoregressively
predicts only the cells whose RGB evidence changes. The result is a
depth-aware policy whose geometric reasoning cost scales with scene change
rather than with the full 100-token grid.

\subsection{Adaptive depth perception data}
\label{sec:think-data}

For every robotics dataset used by \molmothink in the mixtures described in
\autoref{sec:data}, we attach depth annotations to the policy observation
stream. We use the same camera stream that is presented to the policy for depth
reasoning: single-view datasets use their canonical observation camera,
and multi-view datasets use the first policy view with available depth annotations. For heterogeneous LeRobot datasets, when a camera is not specified
explicitly, the video feature is selected from the dataset metadata.

For each RGB frame, we estimate a dense monocular depth map with Depth Anything
V2~\citep{depth_anything_v2}. We then quantize that depth map with the trained
depth VQ-VAE used by \molmoact~\citep{lee2025molmoactactionreasoningmodels},
following the tokenization scheme of \citet{ning2023tokensunifyingoutputspace}.
The VQ-VAE operates on a \(320 \times 320\) depth image with a downsampling
factor of 32, producing a \(10 \times 10\) grid of codebook indices in
\(\{0,\ldots,127\}\).

The same preprocessing pass also constructs the adaptive-depth side channels
used during training and inference. Let \(d_t \in \{0,\ldots,127\}^{100}\) be
the full VQ depth codes for frame \(t\), flattened in raster order. We maintain
a buffer \(b_t\) and update mask \(m_t \in \{0,1\}^{100}\). For the first frame
of an episode, all positions are updated and \(b_1=d_1\). For later frames, the
RGB image is resized to \(320 \times 320\), divided into the same \(10 \times
10\) grid of \(32 \times 32\) patches, and each patch is compared to the
corresponding patch in the previous frame by cosine similarity. A cell is
marked updated when the similarity is below 0.996:
\begin{equation}
    m_{t,i}
    =
    \mathbf{1}
    \left[
        \cos(x_{t,i}, x_{t-1,i}) < 0.996
    \right],
    \qquad
    b_{t,i}
    =
    \begin{cases}
    d_{t,i}, & m_{t,i}=1, \\
    b_{t-1,i}, & m_{t,i}=0.
    \end{cases}
\end{equation}
Thus each frame stores the full depth codes \(d_t\), the carried-forward depth
buffer \(b_t\), and the binary update mask \(m_t\). The model is supervised on
the buffer codes, matching the representation it will maintain during adaptive
inference.

\subsection{Training}
\label{sec:think-training}

\paragraph{Post-training.}
\molmothink starts from the same 200K-step \molmoacttpre checkpoint as the
standard \molmoactt post-training stage. We keep the post-training recipe in
\autoref{sec:posttraining}: the same data construction, action expert, per-layer
KV conditioning, sequence lengths, optimizer settings, 100K training
updates, global batch size of 128, and 64 H100 GPUs. On top of \molmoactt, we added depth-related special tokens for training \molmothink, where the full depth-token interface is in \autoref{supp:model-depth}.

Within robot data, post-training uniformly samples three tasks: action
prediction, depth prediction, and depth-and-action prediction. The action task
trains the usual discrete and continuous action objectives. The depth task
trains autoregressive prediction of the 100 depth-buffer tokens. In the
depth-and-action task, the model first predicts depth tokens, then predicts
discrete action tokens autoregressively while the action expert conditions on
the input context and predicted depth state to generate continuous actions. The
prompt templates and assistant-side triggers for these output styles are given
in \autoref{supp:training}. As in \autoref{sec:posttraining}, the target
action-token span is masked from the expert conditioning path, so the expert
can use the task, observations, state, and depth tokens, but not the discrete
action target it is trained to predict.

\paragraph{Fine-tuning.}
The \molmothinklibero checkpoint is fine-tuned from the depth-aware
post-trained checkpoint using the same full-\libero recipe and compute budget as
\autoref{sec:finetuning}: robot-only training on the combined Spatial, Object,
Goal, and Long suites, with front-view then wrist-view camera order,
2100-token sequences, global batch size 64, 32 H100 GPUs, and 50K updates. We select the
best checkpoint at 30K updates for evaluation.

Fine-tuning differs from depth post-training in three targeted ways. First, we
sample only action and depth-and-action examples, again uniformly, and remove
the pure depth-prediction style. Second, because inference conditions on depth
tokens predicted by the model rather than oracle depth tokens, we inject noise
into the teacher-forced depth prefix during training: 10\% of depth-code input
tokens are replaced by uniformly sampled depth codes, while the prediction
targets remain unchanged. Third, we add a learned per-layer gate on the depth
portion of the action expert's per-layer KV conditioning. For action-expert layer
\(\ell\), let \(M_t=1\) denote positions belonging to the depth-output trigger,
depth delimiters, or depth-code tokens, and let \(A_t\) denote valid context
positions. The gate is computed from the non-depth context of the corresponding
VLM layer,
\begin{equation}
    c_\ell
    =
    \frac{
        \sum_t A_t(1-M_t)V^{\mathrm{vlm}}_{\ell,t}
    }{
        \sum_t A_t(1-M_t)
    },
    \qquad
    g_\ell
    =
    \sigma(w_\ell^\top c_\ell + b_\ell),
\end{equation}
and then applied only to depth-token keys and values:
\begin{equation}
    \bar{K}^{\mathrm{vlm}}_{\ell,t}
    =
    \left(1-M_t+M_t g_\ell\right)K^{\mathrm{vlm}}_{\ell,t},
    \qquad
    \bar{V}^{\mathrm{vlm}}_{\ell,t}
    =
    \left(1-M_t+M_t g_\ell\right)V^{\mathrm{vlm}}_{\ell,t}.
\end{equation}
The gated \((\bar{K}^{\mathrm{vlm}}_{\ell},\bar{V}^{\mathrm{vlm}}_{\ell})\)
are then projected into the action expert as in \autoref{sec:posttraining}.
The gate is initialized with bias \(-4\), so fine-tuning begins close to the
standard action-conditioning path and learns how strongly each expert layer
should use the depth prefix. Additional model-level details for the depth
extension are summarized in \autoref{supp:model-depth}.

\subsection{Adaptive depth inference}
\label{sec:adaptive-depth-inference}

\paragraph{Inference pipeline.}
At inference, \molmothink uses the depth-and-action output style. Given the task,
current observation, and proprioceptive state, the prompt requests
\texttt{<depth\_output><action\_output>}. The model first performs a prefill
over the prompt and images. If no depth cache is available, as at the beginning
of a rollout or after a reset, it autoregressively predicts the full depth
sequence: \texttt{<depth\_start>}, 100 depth-code tokens, and
\texttt{<depth\_end>}.

When a depth cache is available, the model compares the current first
observation image to the cached previous image using the same \(10 \times 10\)
RGB-patch cosine threshold described in \autoref{sec:think-data}. Updated cells
are generated by argmax decoding from the depth-token logits. Unchanged cells
are replayed from the previous predicted depth buffer and consumed by the model
as known depth-token inputs. Consecutive unchanged spans are replayed together,
whereas changed spans are decoded token by token. After all 100 cells are
filled, the model emits \texttt{<depth\_end>} and stores the current image and
the newly filled 100-code depth buffer as the cache for the next control step.

The action is then generated from the depth-conditioned state. For continuous
control, the action expert receives the VLM keys and values corresponding to
the prompt plus the filled depth prefix and integrates the flow-matching
velocity field to produce the action chunk. Adaptive depth changes only how the intermediate depth prefix is produced. The
action interface remains the same as the depth-and-action training objective.

\paragraph{Inference optimization.}
Adaptive depth inference introduces a different systems challenge from the continuous action expert: each frame can contain a different mixture of regenerated depth cell and replayed cached cells, so the overall decode schedule is data-dependent. Capturing the full adaptive loop would require a separate graph for each update pattern. We therefore keep the adaptive scheduler eager, including span-level replay for unchanged cells, while using a preallocated static KV cache to make the decode state stable across steps. For regenerated depth tokens, we capture the fixed-shape transformer work from post-attention through the next layer's pre-attention as CUDA Graph stages, and leave attention itself eager because its effective KV length changes throughout decoding. This preserves adaptive depth reuse while reducing the launch bubbles in the repeated one-token decode path.

\begin{table*}[t]
    \caption{\textbf{Embodied reasoning results.} We evaluate on benchmarks across spatial reasoning, pointing, and embodied QA tasks. The best-performing open-weight model on each benchmark is in \textbf{bold}, and the second best is \underline{underlined}.}
    \renewcommand{\arraystretch}{1.15}
    \setlength{\tabcolsep}{3pt}
    \centering
    \resizebox{\textwidth}{!}{%
    \begin{tabular}{@{}lcccccccccccccc@{}}
        \textbf{\large Models/Benchmarks} &
        \rotatebox{90}{\shortstack{\textbf{\large Point-Bench} \\ \citep{cheng2025pointarena}}} &
        \rotatebox{90}{\shortstack{\textbf{\large RefSpatial} \\ \citep{zhou2025roborefer}}} &
        \rotatebox{90}{\shortstack{\textbf{\large RoboSpatial-Poi} \\ \citep{song2025robospatial}}} &
        \rotatebox{90}{\shortstack{\textbf{\large Where2Place} \\ \citep{yuan2024robopoint}}} &
        \rotatebox{90}{\shortstack{\textbf{\large BLINK} \\ \citep{fu2024blink}}} &
        \rotatebox{90}{\shortstack{\textbf{\large CV-Bench} \\ \citep{tong2024cambrian}}} &
        \rotatebox{90}{\shortstack{\textbf{\large ERQA} \\ \citep{team2025gemini}}} &
        \rotatebox{90}{\shortstack{\textbf{\large EmbSpatial} \\ \citep{du2024embspatial}}} &
        \rotatebox{90}{\shortstack{\textbf{\large MindCube} \\ \citep{luopyspatial}}} &
        \rotatebox{90}{\shortstack{\textbf{\large RoboSpatial-VQ} \\ \citep{song2025robospatial}}} &
        \rotatebox{90}{\shortstack{\textbf{\large SAT} \\ \citep{ray2025satdynamicspatialaptitude}}} &
        \rotatebox{90}{\shortstack{\textbf{\large OpenEQA} \\ \citep{majumdar2024openeqa}}} &
        \rotatebox{90}{\shortstack{\textbf{\large VSI-Bench} \\ \citep{yang2025thinking}}} &
        \rotatebox{90}{\textbf{\large Overall Avg.}} \\
        \midrule
        \multicolumn{15}{@{}l}{\textbf{\textit{\large API call only}}} \\
        {\large GR-ER 1.5 Thinking~\citep{team2025gemini}} & {\large 71.6} & {\large 48.5} & {\large 31.1} & {\large 59.0} & {\large 57.8} & {\large 84.3} & {\large 54.8} & {\large 78.4} & {\large 54.7} & {\large 79.3} & {\large 76.7} & {\large 55.0} & {\large 45.8} & {\large 61.3} \\
        {\large GR-ER 1.5~\citep{team2025gemini}}          & {\large 73.3} & {\large 41.8} & {\large 25.3} & {\large 48.0} & {\large 65.2} & {\large 83.6} & {\large 47.0} & {\large 73.4} & {\large 47.7} & {\large 57.0} & {\large 62.0} & {\large 50.5} & {\large 39.9} & {\large 55.0} \\
        {\large Gemini 2.5 Pro~\citep{comanici2025gemini}} & {\large 62.7} & {\large 33.6} & {\large 8.3} & {\large 37.0} & {\large 69.2} & {\large 85.9} & {\large 56.0} & {\large 78.0} & {\large 59.2} & {\large 71.3} & {\large 74.7} & {\large 55.0} & {\large 51.1} & {\large 57.1} \\
        {\large GPT-5~\citep{singh2025openai}}             & {\large 43.6} & {\large 23.5} & {\large 19.0} & {\large 37.0} & {\large 71.3} & {\large 86.1} & {\large 59.0} & {\large 81.5} & {\large 58.0} & {\large 69.3} & {\large 86.7} & {\large 64.4} & {\large 52.9} & {\large 57.9} \\
        {\large GPT-5-mini ~\citep{singh2025openai}}       & {\large 39.5} & {\large 23.0} & {\large 12.5} & {\large 33.5} & {\large 56.4} & {\large 85.9} & {\large 57.3} & {\large 78.8} & {\large 55.6} & {\large 70.7} & {\large 81.3} & {\large 59.2} & {\large 46.2} & {\large 53.8} \\
        \midrule
        \multicolumn{15}{@{}l}{\textbf{\textit{\large Open weights only}}} \\
        {\large Qwen3-VL-4B~\citep{qwen3}}                 & {\large 63.8} & {\large \underline{49.5}} & \textbf{\large 62.3} & \textbf{\large 63.0} & {\large 65.2} & \underline{\large 85.8} & {\large 39.5} & {\large 78.1} & {\large 27.4} & {\large 62.7} & {\large 60.7} & \underline{\large 47.6} & {\large 61.4} & {\large 59.0} \\
        {\large Qwen3-VL-8B~\citep{qwen3}}                 & {\large 64.2} & {\large 47.5} & \underline{\large 56.6} & \textbf{\large 63.0} & \underline{\large 66.6} & {\large 85.6} & {\large 41.5} & \underline{\large 78.2} & {\large 33.4} & \textbf{\large 73.7} & \underline{\large 70.0} & \textbf{\large 49.3} & \underline{\large 63.3} & \underline{\large 61.0} \\
        {\large LLaVA-OV-7B~\citep{li2024llava}}           & {\large 9.4} & {\large 3.0} & {\large 0.0} & {\large 19.0} & {\large 44.2} & {\large 70.3} & {\large 39.8} & {\large 69.7} & \underline{\large 45.6} & {\large 68.9} & {\large 58.0} & {\large 36.2} & {\large 31.2} & {\large 38.1} \\
        {\large InternVL3.5-4B~\citep{wang2025internvl3}}  & {\large 43.0} & {\large 21.5} & {\large 23.0} & {\large 53.0} & {\large 58.7} & {\large 80.5} & {\large 37.0} & {\large 72.0} & {\large 42.6} & {\large 61.0} & {\large 49.3} & {\large 44.6} & {\large 56.5} & {\large 49.4} \\
        {\large InternVL3.5-8B~\citep{wang2025internvl3}}  & {\large 45.5} & {\large 35.0} & {\large 27.0} & \underline{\large 57.0} & {\large 59.1} & {\large 80.9} & {\large 41.0} & {\large 74.7} & {\large 39.7} & {\large 62.7} & {\large 58.7} & {\large 43.5} & {\large 56.4} & {\large 52.4} \\
        \midrule
        \multicolumn{15}{@{}l}{\textbf{\textit{\large Molmo family: Open weights, Open data, Open code}}} \\
        {\large Molmo2 ~\citep{clark2026molmo2openweightsdata}}                                 & \underline{\large 76.9} & \textbf{\large 52.5} & {\large 13.9} & {\large 18.0} & {\large 50.8} & {\large 64.4} & \underline{\large 46.3} & {\large 67.0} & {\large 37.6} & {\large 57.9} & {\large 50.0} & {\large 46.9} & {\large 26.1} & {\large 46.8} \\
        {\color{ai2pink} \large \molmoer}                              & \textbf{\large 77.3} & \textbf{\large 52.5} & {\large 32.0} & {\large 54.0} & \textbf{\large 72.5} & \textbf{\large 87.8} & \textbf{\large 46.8} & \textbf{\large 78.8} & \textbf{\large 57.0} & \underline{\large 73.4} & \textbf{\large 78.0} & {\large 44.7} & \textbf{\large 74.5} & \textbf{\large 63.8} \\
    \end{tabular}%
    }
    \label{tab:er_results}
\end{table*}

\section{Experiments}
\label{sec:exp}

\providecommand{\ci}[1]{\textnormal{\textcolor{gray}{\scriptsize$\pm$#1}}}

\begin{table}[t]
\centering
\caption{\textbf{Results across MolmoSpace tasks.} We evaluate all models on four manipulation skill categories---Pick, Pick \& Place, Open, and Close---and report mean success rate (\%) with standard error across three independent evaluation runs. \textbf{Bold} denotes the best and \underline{underline} the second-best per task. MolmoAct2-DROID achieves the highest overall average (37.7), outperforming the strongest prior baseline $\pi_{0.5}$-DROID (34.5) by +3.2 points, with particularly large gains on Pick (+7.3) and Pick \& Place (+13.1), where contact-rich, multi-stage reasoning is required. On Close, MolmoAct2-DROID also sets a new best at 70.8, while on Open it trails the leading methods, suggesting that articulated-object interaction remains a direction for further improvement. Standard errors remain comparable across methods ($\leq 3.2$), indicating that the observed gains are stable across runs rather than artifacts of evaluation variance.}
\label{tab:ms_results}
\begin{tabular}{lccccc}
\textbf{Models/Tasks} & \textbf{Pick} & \textbf{Pick \& Place} & \textbf{Open} & \textbf{Close} & \textbf{Average} \\
\midrule
StereoVLA ~\citep{deng2025stereovla}           & 7.0\ci{2.3} & --   & -- & -- & 7.0 \\
LAP-VLA ~\citep{zha2026lap}            & 24.9\ci{2.7} & 6.6\ci{1.6}   & \underline{11.4}\ci{1.5} & 45.9\ci{2.9} & 22.2 \\
$\pi_{0}$-DROID ~\citep{black2410pi0}    & 16.2\ci{2.6} & 12.5\ci{2.0} & 11.0\ci{2.0} & 53.1\ci{3.2} & 23.2 \\
$\pi_{0.5}$-DROID ~\citep{intelligence2025pi05visionlanguageactionmodelopenworld}  & \underline{36.4}\ci{2.9} & \underline{13.6}\ci{2.2} & \textbf{22.7}\ci{2.6} & \underline{65.1}\ci{3.1} & \underline{34.5} \\
{\color{ai2pink}\molmoacttdroid}         & \textbf{43.7}\ci{3.1} & \textbf{26.7}\ci{2.8} & 9.5\ci{2.0} & \textbf{70.8}\ci{3.0} & \textbf{37.7} \\

\end{tabular}
\end{table}

\begin{table}[t]
\centering
\caption{\textbf{Evaluation on simulation held-out environments.} Simulation success rates are evaluated over 1000 episodes per task. For pick-and-place tasks, we report both oracle success (first number, which is the success conditions being fulfilled at any timestep) and success at end (second number, the success conditions being fulfilled at the final timestep). The delta between captures both unstable/unsuitable placement and the inability of policies to determine when a specified task is already completed, e.g. by repeatedly picking up an object which has already been placed correctly. We additionally report the half-width of the 95\% confidence interval bounds for each result. \textbf{Bold} denotes the best and \underline{underline} the second-best per task.}
\label{tab:sim_results}
\resizebox{\textwidth}{!}{
\begin{tabular}{@{}l@{\hspace{0.5em}}lc|ccc|ccc|c}
&\textbf{Models/Tasks} & \textbf{Pick MSProc} & \textbf{Pick Classic} & \textbf{Pick} & \textbf{Pick Rand.-Cam.} & \textbf{Pick\&Place} & \textbf{PnP Next-To} & \textbf{PnP Color} & \textbf{Avg.} \\
\toprule
&StereoVLA~\citep{deng2025stereovla} & 6.6\ci{2.6} & 4.3\ci{1.5} & 1.1\ci{1.0} & - & - & - & - & - \\
&X-VLA~\citep{zheng2025x} & 3.3\ci{1.0} & 0.5\ci{0.5} & 0.7\ci{0.5} & 0.8\ci{0.5} & 0.1\ci{0.2}/0.1\ci{0.2} & 1.9\ci{0.9}/1.0\ci{0.7} & 0.9\ci{0.6}/0.5\ci{0.5} & 1.2 \\
&LAP-VLA~\citep{zha2026lap} & \underline{19.4}\ci{2.4} & 2.4\ci{1.0} & 3.1\ci{1.1} & 2.7\ci{1.0} & 3.81\ci{1.5}/1.59\ci{1.0} & 6.48\ci{2.8}/3.41\ci{2.1} & 3.1\ci{1.1}/1.5\ci{0.8} & 4.8 \\
&$\pi_{0.5}$-DROID~\citep{intelligence2025pi05visionlanguageactionmodelopenworld}& 18.1\ci{2.4} & \underline{6.4}\ci{1.5} & \underline{7.0}\ci{1.6} & \underline{8.0}\ci{1.9} & \underline{11.7}\ci{2.1}/\underline{7.6}\ci{1.7} & \underline{8.2}\ci{2.2}/\underline{6.2}\ci{1.9} & \underline{10.4}\ci{1.9}/\underline{6.7}\ci{1.6} & \underline{10.0} \\
&{\color{ai2pink}\molmoacttdroid} & $\mathbf{35.6}$\ci{3.0} & $\mathbf{18.9}$\ci{2.6} & $\mathbf{20.5}$\ci{2.6} & $\mathbf{15.4}$\ci{2.4} & $\mathbf{15.8}$\ci{2.4}/$\mathbf{7.8}$\ci{1.8} & $\mathbf{20.9}$\ci{2.6}/$\mathbf{14.4}$\ci{2.3} & $\mathbf{17.2}$\ci{2.5}/$\mathbf{8.8}$\ci{2.0} & $\mathbf{20.6}$ \\
\end{tabular}
}
\end{table}

\begin{table}[t]
\centering
\caption{\textbf{Success rates (\%) across manipulation tasks.} Each cell reports the percentage of successful trajectories out of 15 trials. \textbf{Bold} denotes the best and \underline{underline} the second-best per task.}
\label{tab:task_success}
\resizebox{\textwidth}{!}{%
\begin{tabular}{lcccccc}
\textbf{Models/Tasks} & \textbf{Apple on plate} & \textbf{Pipette in tray} & \textbf{Red cube in tape roll} & \textbf{Knife in box} & \textbf{Objects in bowl} & \textbf{Average} \\
\midrule
$\pi_{0.5}$-DROID~\citep{intelligence2025pi05visionlanguageactionmodelopenworld}& 66.7                    & 33.3                    & \underline{53.3}        & 26.7                    & \underline{46.2} & 45.2 \\
MolmoBot~\citep{deshpande2026molmob0t}          & \underline{86.7}        & \underline{53.3}        & 33.3                    & \underline{40.0}        & 28.6             & \underline{48.4} \\
{\color{ai2pink}\molmoacttdroid}         & \textbf{100.0}          & \textbf{86.7}           & \textbf{93.3}           & \textbf{93.3}           & \textbf{62.0}    & \textbf{87.1} \\
\end{tabular}%
}
\end{table}

\begin{table}[h]
\centering
\caption{\textbf{Success rates (\%) across manipulation tasks on the SO-100 platform.} Each cell reports the mean score over 15 trials, where partial credit is awarded for sub-task completion (0.25 for reaching, 0.5 for pickup, 1.0 for successful placement). \textbf{Bold} denotes the best and \underline{underline} the second-best per task.}
\label{tab:so100_task_success}
\resizebox{\textwidth}{!}{%
\begin{tabular}{lcccccc}
\textbf{Models/Tasks} & \textbf{Fork on plate} & \textbf{Stack blocks} & \textbf{Tissues in basket} & \textbf{Pen on notebook} & \textbf{Block in box} & \textbf{Average} \\
\midrule
SmolVLA~\citep{shukor2025smolvlavisionlanguageactionmodelaffordable}        & 3.3                  & 5.0                  & 0.0                  & 3.3                  & 0.0                  & 2.3 \\
$\pi_{0}$-SO100/101~\citep{chen2025depi}            & \underline{30.0}     & \underline{6.7}      & \underline{20.0}     & \underline{80.0}     & \textbf{90.0}        & \underline{45.3} \\
{\color{ai2pink}\molmoacttso} & \textbf{70.0}        & \textbf{20.0}        & \textbf{73.3}        & \textbf{86.7}        & \underline{33.3}     & \textbf{56.7} \\
\end{tabular}%
}
\end{table}

\begin{figure}[t]
    \centering
    \includegraphics[width=\linewidth]{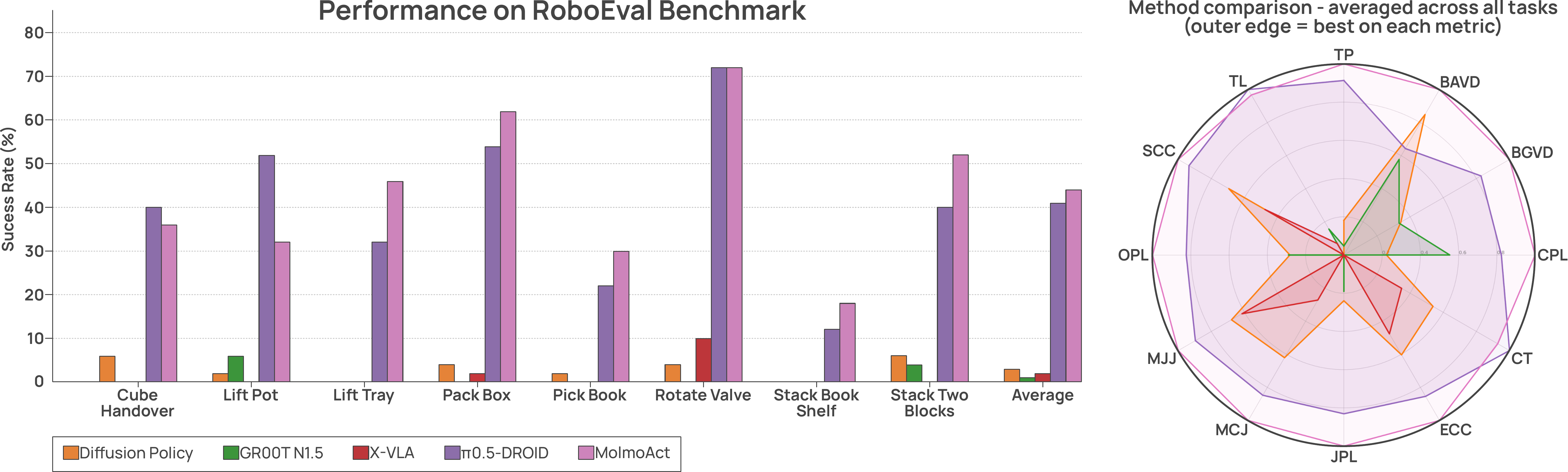}

   \caption{\textbf{Performance comparison on the RoboEval benchmark.}
(A) Task-wise success rates (\%) across eight manipulation tasks for Diffusion Policy, GR00T N1.5, X-VLA, $\pi_{0.5}$-DROID, and MolmoAct. MolmoAct achieves the highest or near-highest success rates on most tasks, with particularly strong performance on long-horizon tasks such as \textit{Pack Box} and \textit{Rotate Valve}. $\pi_{0.5}$-DROID is the strongest baseline overall, while other methods exhibit lower and more variable performance across tasks.
(B) Radar plot of normalized performance across behavioral and outcome metrics, averaged over all tasks (outer edge indicates best performance per metric). MolmoAct consistently dominates across nearly all metrics, indicating improvements not only in task success but also in efficiency, stability, and trajectory quality. \textit{Abbreviations:} CT = completion time, TL = trajectory length, JPL = joint path length, CPL = Cartesian path length, CJ = Cartesian jerk, JJ = joint jerk, SC = self-collisions, SL = slip count.}

\label{fig:roboeval-results}
\end{figure}

\definecolor{norapink}{RGB}{252, 224, 224}
\definecolor{noralavender}{RGB}{226, 226, 250}
\begin{table}[h]
\centering
\small
\caption{\textbf{LIBERO benchmark success rates} across four task categories (Spatial, Object, Goal, and Long-horizon) along
with the average performance. \molmoactt achieves the highest overall average success rate of 97.2\%, outperforming
all strong baselines, with strong performance across all categories, particularly scoring 100\% on the LIBERO-Object tasks. Furthermore improvement could be seem going from \molmoactt to \molmothink. \textbf{Bold} denotes the best and \underline{underline} the second-best per task.}
\begin{tabular}{lccccc}
\textbf{Baseline} & \textbf{Spatial} & \textbf{Object} & \textbf{Goal} & \textbf{Long} & \textbf{Average} \\
\midrule
TraceVLA~\citep{zheng2024tracevla}           & 84.6\% & 85.2\%          & 75.1\% & 54.1\% & 74.8\% \\
OpenVLA~\citep{kim2024openvla}            & 84.7\% & 88.4\%          & 79.2\% & 53.7\% & 76.5\% \\
SpatialVLA~\citep{qu2025spatialvla}         & 88.2\% & 89.9\%          & 78.6\% & 55.5\% & 78.1\% \\
CoT-VLA~\citep{zhao2025cot}            & 87.5\% & 91.6\%          & 87.6\% & 69.0\% & 83.9\% \\
$\pi_0$~\citep{black2410pi0}            & 96.8\% & 98.8\%          & 95.8\% & 85.2\% & 94.2\% \\
ThinkAct~\citep{huang2025thinkact}           & 88.3\% & 91.4\%          & 87.1\% & 70.9\% & 84.4\% \\
MolmoAct-7B-D~\citep{lee2025molmoactactionreasoningmodels}      & 87.0\% & 95.4\%          & 87.6\% & 77.2\% & 86.6\% \\
GR00T N1.7~\citep{gr00tn1_2025}         & 97.7\% & 97.5\%          & \underline{98.5\%} & \underline{94.4\%} & 97.0\% \\
$\pi_{0.5}$~\citep{intelligence2025pi05visionlanguageactionmodelopenworld}       & \underline{98.8\%} & 98.2\%          & 98.0\% & 92.4\% & 96.9\% \\
NORA-1.5 ~\citep{hung2025nora} & 97.3\% & 96.4\% & 94.5\% & 89.6\% & 94.5\% \\
{\color{ai2pink}\molmoactt}          & 97.8\% & \textbf{100.0\%} & 97.8\% & 93.2\% & \underline{97.2\%} \\
{\color{ai2pink}\molmothink}   & \textbf{98.8\%} & \underline{99.8\%} & \textbf{98.5\%} & \textbf{95.4\%} & \textbf{98.1\%} \\
\end{tabular}
\label{tab:baseline-comparison}
\end{table}

\begin{figure}[t]
    \centering
    \includegraphics[width=\linewidth]{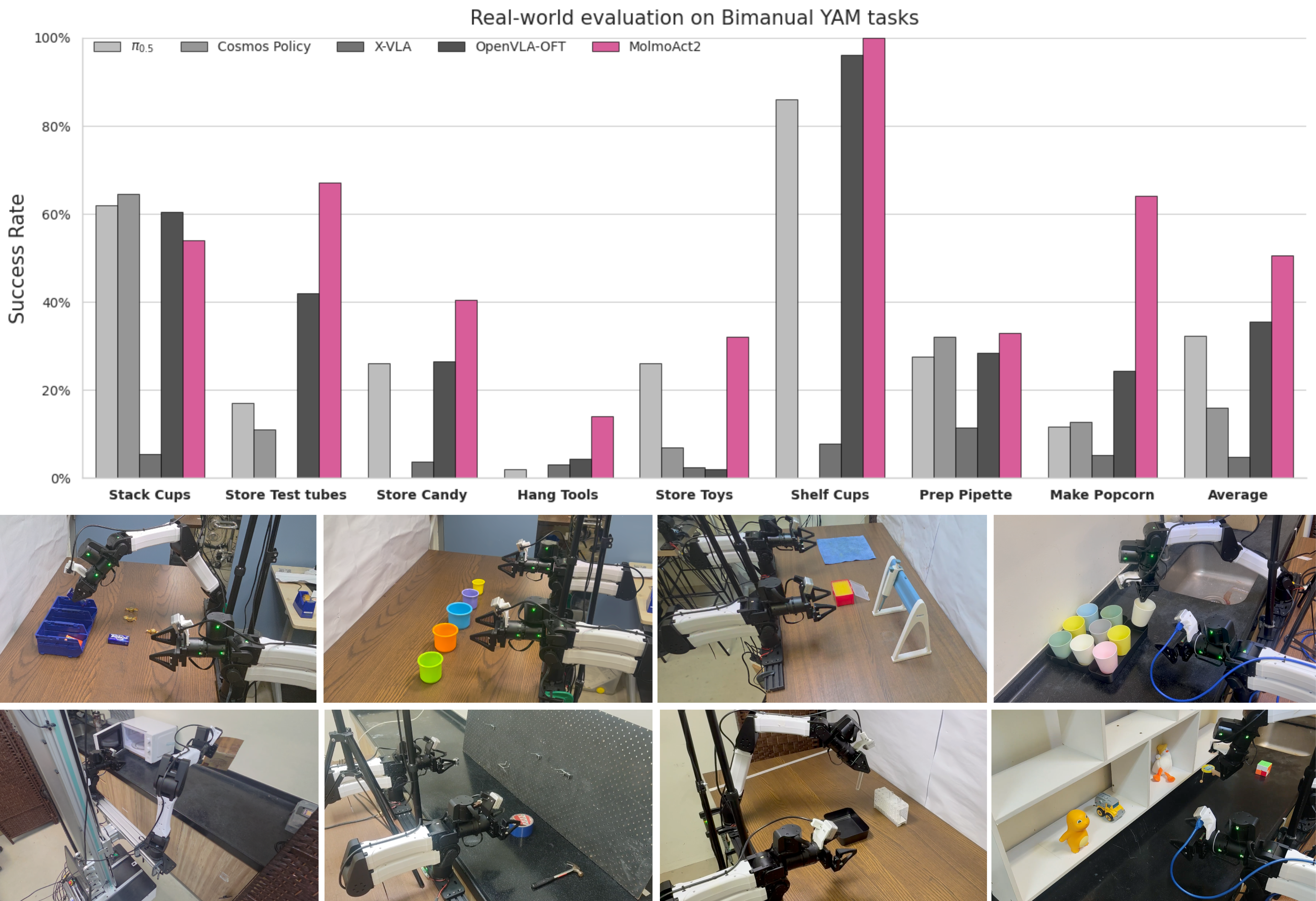}
    \caption{\textbf{Overview of efficient fine-tuning of \molmoactt.} 
We conduct a comprehensive evaluation across 8 real-world tasks outside of a 
laboratory setting, ranging from preparing a pipette for a chemist to placing 
toys back onto a shelf. \molmoactt, fine-tuned on the collected evaluation data, 
outperforms 4 strong baselines by a large margin of 15\% over the runner-up model.}
    \label{yamreal}
\end{figure}

Our experimental evaluation comprises one of the most extensive suites of studies of \molmoactt, benchmarked against a diverse set of strong generalist baseline models. Specifically, we assess \molmoactt across three main categories of evaluation: (i) we examine how \molmoer performs relative to generalist VLM models used as training backbones for VLAs, and quantify the performance gains attributable to a stronger backbone; (ii) we evaluate the out-of-the-box deployment capabilities of \molmoactt across a variety of embodiments; and (iii) we investigate the ease and efficiency with which \molmoactt can be adapted to novel tasks and unseen embodiments via fine-tuning. 

\noindent Hence, we evaluate \molmoactt across a comprehensive range of scenarios in both simulation and the real world, addressing the following research questions:

\begin{enumerate}
    \item \textbf{How well does \molmoer perform on established industry benchmarks for embodied reasoning in VLMs?} We address this question by benchmarking \molmoer against strong open-source and proprietary VLMs across 13 established embodied reasoning benchmarks.



    \item \textbf{How well does \molmoactt perform out-of-the-box?} We investigate this by evaluating \molmoactt on three benchmarks: two simulation benchmarks, MolmoBot~\citep{deshpande2026molmob0t} and MolmoSpaces~\citep{kim2026molmospaces}, and a series of real-world zero-shot evaluations across two embodiments (DROID and SO-100/101 setups) against strong baselines pretrained or fine-tuned on the DROID~\citep{khazatsky2024droid} and large-scale SO-100/101 datasets.


    \item \textbf{How effectively can \molmoactt be fine-tuned for new tasks and embodiments?} We study this by fine-tuning \molmoactt on custom datasets collected across three benchmarks: two simulation benchmarks, RoboEval \citep{wang2025roboeval} and LIBERO \citep{liu2023libero}, and one large-scale real-world evaluation suite consisting of in-the-wild bimanual YAM tasks.

    \item \textbf{Do adaptive depth-perception tokens improve the performance of \molmothink?}
    We investigate this question by comparing \molmoactt against \molmothink on LIBERO~\citep{liu2023libero}, MolmoSpace~\citep{kim2026molmospaces}, and the MolmoBot Benchmark~\citep{deshpande2026molmob0t}.

   \item \textbf{How robust is \molmoactt under out-of-distribution perturbations?} To investigate this, we fine-tune \molmoacttpos checkpoint on 4 tasks from the real-world Bimanual YAM benchmark and evaluate each task under 4 distinct out-of-distribution variants.

    \item \textbf{What is the quality of \molmoactt's trajectory rollouts beyond raw success rate?}
    For real-world deployment, trajectory quality matters beyond task success (e.g., stability, smoothness). We therefore evaluate \molmoactt against strong generalist baselines on RoboEval~\citep{wang2025roboeval} to assess the quality of the trajectories it generates while solving the tasks.

    \item \textbf{Which components contributed the largest performance gains to \molmoactt?}
    We investigate this question through systematic component-level ablations on the \molmoactt architecture and training configurations to understand how each of our novel improvements contributes to the model's performance, and we verify our findings on LIBERO~\citep{liu2023libero} for reproducibility.

    \item \textbf{How fast is \molmoactt at inference?}
    We measure end-to-end inference latency and control rate for \molmoactt and \molmothink on LIBERO~\citep{liu2023libero}, and evaluate the impact of our caching and CUDA Graph optimizations.

\end{enumerate}

\subsection{\molmoer evaluation}
\label{sec:exp-er}

\paragraph{Evaluation setup.} We evaluate \molmoer on 13 established vision-language benchmarks that are widely used to measure the embodied reasoning capabilities of VLMs: Point-Bench~\citep{cheng2025pointarena}, RefSpatial~\citep{zhou2025roborefer}, RoboSpatial-Point~\citep{song2025robospatial}, Where2Place~\citep{yuan2024robopoint}, BLINK~\citep{fu2024blink}, CV-Bench~\citep{tong2024cambrian}, ERQA~\citep{team2025gemini}, EmbSpatial~\citep{du2024embspatial}, MindCube~\citep{luopyspatial}, SAT~\citep{ray2025satdynamicspatialaptitude}, OpenEQA~\citep{majumdar2024openeqa}, and VSI-Bench~\citep{yang2025thinking}. Together, these benchmarks span visual question answering, 2D pointing, multi-image reasoning, ego–exo understanding, and video-based spatial reasoning. 

\paragraph{Baselines.}
We compare \molmoer against a diverse set of proprietary VLMs, reasoning-oriented VLMs, and open-weight VLMs, including the Gemini family~\citep{team2024gemini,comanici2025gemini}, the GPT-5 family~\citep{singh2025openai}, LLaVA-OneVision~\citep{li2024llava}, the Qwen3-VL family~\citep{bai2025qwen3}, and InternVLA~\citep{wang2025internvl3}. We additionally compare against \molmot~\citep{clark2026molmo2openweightsdata}, the base model from which \molmoer is trained.

\paragraph{Results.} \molmoer outperforms all baseline VLMs on 9 of the 13 benchmarks and achieves the highest overall average score of 63.8\%, exceeding the runner-up, Gemini-ER 1.5 Thinking, by 2.5 points. Notably, \molmoer improves over its base model \molmot by 17\%, demonstrating the substantial gains in embodied reasoning capability conferred by our approach.

\subsection{Out-of-the-box deployment}
\label{sec:exp-outbox}

\textbf{Evaluation setup and baselines.} To verify the out-of-the-box deployment capability of \molmoactt, we evaluate it in both simulation and real-world settings. In simulation, we evaluate the \molmoacttdroid checkpoint on two benchmarks: MolmoSpaces~\citep{kim2026molmospaces} and MolmoBot~\citep{deshpande2026molmob0t}. Both target single-cycle pick-and-place tasks across diverse objects and environments, with MolmoBot featuring more challenging objects and scenarios. Each benchmark replicates the original DROID~\citep{khazatsky2024droid} setup, enabling zero-shot evaluation of any policy pretrained or fine-tuned on the DROID dataset. On MolmoSpaces, we compare \molmoacttdroid against LAP-VLA~\citep{zha2026lap}, StereoVLA~\citep{deng2025stereovla}, $\pi_{0}$-DROID~\citep{black2410pi0}, and $\pi_{0.5}$-DROID~\citep{intelligence2025pi05visionlanguageactionmodelopenworld}; on MolmoBot, we compare against LAP-VLA, $\pi_{0.5}$-DROID~\citep{intelligence2025pi05visionlanguageactionmodelopenworld}, and X-VLA~\citep{zheng2025x}. All simulation evaluations strictly follow the protocols, trial counts, and procedures of the respective original papers. Further details are provided in the Appendix.

In the real world, we test \molmoacttdroid and \molmoacttso on their respective pretrained embodiment setups under challenging out-of-distribution conditions: camera poses are randomly initialized without conforming to any dataset visual distribution, using two cameras (wrist and a single allocentric view); all objects are unseen during training; and the environments themselves are out-of-distribution relative to the training data. For the DROID embodiment, we compare \molmoacttdroid against $\pi_{0.5}$-DROID~\citep{intelligence2025pi05visionlanguageactionmodelopenworld}, fine-tuned on the DROID dataset, and MolmoBot~\citep{deshpande2026molmob0t}, fine-tuned on the large-scale MolmoSpaces dataset. We evaluate on five tasks ranging from single- to multi-object pick-and-place: \texttt{apple\_on\_plate}, \texttt{pipette\_in\_tray}, \texttt{red\_cube\_in\_tape\_roll}, \texttt{knife\_in\_box}, and \texttt{objects\_in\_bowl}. For the SO-100/101 setup, we use the SO-100 robot, retaining the wrist camera and adding a third-person external camera with a randomly initialized position. We compare \molmoacttso against SmolVLA~\citep{shukor2025smolvlavisionlanguageactionmodelaffordable} and $\pi_{0}$-SO100/101, a variant of $\pi_{0}$ that we fine-tuned on \molmoacttsodata. Both baselines are pretrained or fine-tuned at scale on the SO-100/101 dataset. We evaluate five pick-and-place tasks with novel objects and randomly initialized camera poses. Every real-world policy is assessed over 15 trials, with partial credit awarded for near-successful executions. Further details are provided in the Appendix.

\textbf{Results.} \molmoacttdroid achieves strong zero-shot performance on all DROID-style setups, in both simulation and the real world. In simulation, \molmoacttdroid is the state-of-the-art VLA model across MolmoSpaces and MolmoBot, outperforming all baselines on both benchmarks. $\pi_{0.5}$-DROID is the runner-up on each, with \molmoactt achieving an absolute gain of $+10.6\%$ on MolmoBot and $+3.2\%$ on MolmoSpaces, averaged across tasks; trial counts in every case are sufficient for statistical significance. In real-world evaluation on the DROID setup, which is substantially more out-of-distribution given the random camera initialization and entirely novel scenes and objects relative to the DROID dataset~\citep{khazatsky2024droid}, \molmoactt again attains the highest performance, reaching $87.1\%$ and surpassing the runner-up MolmoBot by $38.7\%$. On the second zero-shot embodiment, SO-100/101, \molmoacttso reaches $56.7\%$, an $11.4\%$ gain over our open implementation of $\pi_{0}$ on SO-100/101, demonstrating affordable deployment on low-cost robots such as the SO-100/101.

\subsection{Effective fine-tuning}
\label{sec:exp-ft}

\textbf{Evaluation setups and baselines.} For rapid real-world deployment on novel tasks and embodiments, VLAs must adapt from only a handful of demonstrations. To assess \molmoact's capacity for such efficient adaptation, we benchmark it against several strong VLA baselines on downstream tasks, domains, and embodiments unseen during training. In all fine-tuning experiments, we initialize from the \molmoacttpos checkpoint and fine-tune on the benchmark-specific training data.

In simulation, we evaluate on two benchmarks. RoboEval~\citep{wang2025roboeval} comprises a bimanual Franka Emika Panda setup with human-expert demonstrations across 8 tasks requiring bimanual coordination. LIBERO~\citep{liu2023libero}, following prior work~\citep{kim2024openvla}, consists of four task suites—LIBERO-Spatial, LIBERO-Object, LIBERO-Goal, and LIBERO-Long—each containing 500 demonstrations across 10 tasks. We fully fine-tune \molmoactt and compare it against state-of-the-art VLAs trained under similar protocols, including TraceVLA~\citep{zheng2024tracevla}, OpenVLA~\citep{kim2024openvla}, SpatialVLA~\citep{qu2025spatialvla}, CoT-VLA~\citep{zhao2025cot}, $\pi_{0}$~\citep{black2410pi0}, ThinkAct~\citep{huang2025thinkact}, MolmoAct~\citep{lee2025molmoactactionreasoningmodels}, GR00T-N1~\citep{gr00tn1_2025}, $\pi_{0.5}$~\citep{intelligence2025pi05visionlanguageactionmodelopenworld}, and NORA-1.5~\citep{hung2025nora}.

For real-world evaluation, we conduct a large-scale, systematic study on both a bimanual YAM setup and a mobile bimanual YAM setup, covering 8 tasks that span static laboratory settings, in-the-wild environments (kitchen, study room, pantry, and wet labs), and mobile manipulation. The tasks—\texttt{stack\_cups}, \texttt{store\_test\_tubes}, \texttt{store\_candy}, \texttt{hang\_tools}, \texttt{store\_toys}, \texttt{shelf\_cups}, \texttt{prepare\_pipette}, and \texttt{make\_popcorn}—are each evaluated over 50 trials, with three spatial variants per task used for data collection. We fine-tune from the \molmoacttpos checkpoint for every task and compare against four strong VLA and world-model baselines: Cosmos Policy~\citep{kim2026cosmos}, X-VLA~\citep{zheng2025x}, OpenVLA-OFT~\citep{kim2025fine}, and $\pi_{0.5}$-DROID~\citep{intelligence2025pi05visionlanguageactionmodelopenworld}.

\textbf{Results.} On LIBERO, \molmoactt achieves an average success rate of 97.2\%, the highest among all compared methods—reaching 100\% on LIBERO-Object and improving over our previous MolmoAct-7B-D by 10.6\%. This trend is reinforced on RoboEval, where \molmoactt attains a 44.3\% success rate, surpassing the second-best model, $\pi_{0.5}$~\citep{intelligence2025pi05visionlanguageactionmodelopenworld}, by 3.8\%. In the real-world evaluation, \molmoactt outperforms all baselines on 7 of 8 deployment tasks, achieving an average success rate of 50.1\%—15\% above the runner-up, OpenVLA-OFT. Hence, \molmoactt demonstrate its effectiveness for rapid adaptation to new tasks, domains, and embodiments.

\subsection{Performance of \molmothink}
\label{depth}
To isolate the contribution of \molmothink, we fine-tuned it with the adaptive-depth training pipeline and compared it against \molmoactt, fine-tuned from \molmoacttpos under the standard recipe on LIBERO. \molmothink improves over \molmoactt on three of the four task suites and matches it on the fourth, where the baseline is already saturated at 100\%. Crucially, the largest gain (+2.2\%) appears on the most challenging suite, where the baseline leaves the most headroom (93.2\%), while gains on the easier suites are correspondingly smaller as performance approaches ceiling. Averaged across all suites and 2{,}000 rollouts, \molmothink achieves 98.1\% versus 97.2\%, a +0.9\% improvement that is consistent in sign and concentrated where the task is hardest which indicates that the adaptive-depth pipeline yields a real, non-incidental gain rather than noise around saturation.

\begin{table}[h]
\centering
\caption{\textbf{Robustness of \molmoactt{} under environmental perturbations.} 
We evaluate \molmoactt{} against four baselines on out-of-distribution scenarios 
spanning spatial variation, lighting, language rephrasing, and visual distractors. 
\molmoactt{} is the most robust, achieving an average success rate of 
\textbf{50.69\%} across all perturbation types—a \textbf{10.80\%} absolute 
improvement over the next-best model. Values report average success rates (\%); 
\textbf{bold} = best, \underline{underline} = second-best.}
\label{tab:perturbation_averages}
\begin{tabular}{lccccc}
\toprule
\textbf{Model} & \textbf{Spatial Var.} & \textbf{Lighting} & \textbf{Language} & \textbf{Distractor} & \textbf{Overall} \\
\midrule
$\pi_{0.5}$~\citep{intelligence2025pi05visionlanguageactionmodelopenworld}      & \underline{15.00}  & 33.70              & 26.15              & 33.20              & 27.01 \\
Cosmos Policy~\citep{kim2026cosmos}    & 8.75               & 17.50              & 5.00               & 13.75              & 11.25 \\
X-VLA~\citep{wen2025dexvla}     & 3.75               & 8.30               & 9.95               & 3.75               & 6.44 \\
OpenVLA-OFT~\citep{kim2025fine}   & 13.75              & \underline{46.25}  & \underline{51.25}  & \underline{48.30}  & \underline{39.89} \\
{\color{ai2pink}\molmothink}  & \textbf{26.25}     & \textbf{62.05}     & \textbf{60.35}     & \textbf{54.10}     & \textbf{50.69} \\
\bottomrule
\end{tabular}
\end{table}
\subsection{Robustness of \molmoactt to out-of-distribution shifts}
\textbf{Evaluation setup and baselines.} To investigate how robustly \molmoactt transfers after parameter-efficient fine-tuning on a new dataset, we select 4 of the 8 real-world bimanual tasks (\texttt{stack\_cups}, \texttt{store\_test\_tubes}, \texttt{store\_candy}, and \texttt{prepare\_pipette}) and fine-tune a separate checkpoint for each on top of \molmoacttpos. At test time, we evaluate each checkpoint under four out-of-distribution variants: \emph{spatial}, in which we place the target objects at locations outside the training distribution; \emph{lighting}, in which we alter the illumination of the scene; \emph{language variants}, in which we rephrase the language instruction in three different ways; and \emph{distractors}, in which we add unseen distractor objects while keeping the original in-distribution spatial layout. We compare \molmoactt to strong baselines such as $\pi_{0.5}$~\citep{intelligence2025pi05visionlanguageactionmodelopenworld}, Cosmos Policy~\citep{kim2026cosmos}, X-VLA~\citep{wen2025dexvla}, and OpenVLA-OFT~\citep{kim2025fine} .For each task each set, we will evaluate 20 trials, 5 trials for each perturbation. 

\textbf{Results.} Based on the evaluation, we find that \molmoactt is substantially more robust to perturbations than all other evaluated models, achieving a 50.7\% average success rate across all perturbation types, with a margin of $\sim$10.8\% over the second-best model, OpenVLA-OFT. While \molmoactt leads on every perturbation category, its advantage is narrowest on Distractor (a 5.8\% margin over OpenVLA-OFT), and it attains its lowest absolute score on Spatial Variance (26.25\%), indicating room for improvement on fine-grained spatial generalization.

\subsection{Quality of \molmoactt trajectories for real-world deployment}

While task success is a necessary measure of performance, it is insufficient for assessing readiness for real-world deployment. In practical settings, trajectory quality directly impacts efficiency, stability, and safety. We therefore evaluate \molmoactt fine-tuned on RoboEval for beyond success rates using a suite of behavioral and outcome metrics from RoboEval \citep{wang2025roboeval}, summarized in Fig.~\ref{fig:roboeval-results}B. Across tasks, \molmoactt demonstrates consistent improvements in efficiency-related metrics. For example, on \textit{Stack Two Blocks}, \molmoactt reduces completion time from $5.87$s ($\pi_{0.5}$) and $7.27$s (Diffusion) to $4.70$s, while also reducing joint path length from $2.16$ to $1.04$ (approximately $2\times$ shorter). Similar trends hold on longer-horizon tasks such as \textit{Rotate Valve}, where \molmoactt achieves the lowest completion time ($8.51$s vs.\ $9.69$s for $\pi_{0.5}$), along with reduced trajectory length.

In addition to efficiency, \molmoactt exhibits strong performance on stability-related metrics. As shown in Fig.~\ref{fig:roboeval-results}B, \molmoactt consistently operates near the best normalized value across both Cartesian and joint-space measures, indicating smoother and more stable trajectories compared to baselines. In contrast, other methods exhibit greater variability across these metrics, suggesting less consistent control behavior. Together, these results show that \molmoactt not only improves task success, but also produces shorter, more stable, and more efficient trajectories, which are critical properties for real-world deployment.

\subsection{Systematic analysis}
\label{sec:exp-systematic}

We ablate the main design choices in \molmoactt on \libero to isolate where the
performance gains come from. Unless otherwise noted, all variants use the same
data, image augmentation, action normalization, action horizon, and evaluation
protocol as the corresponding \libero fine-tuning run. For the full-suite
ablations, each row is evaluated on Spatial, Object, Goal, and Long, and the
average is computed across the four suites.

\paragraph{Embodied-reasoning backbone.}
We first test whether the \molmoer backbone improves action learning even before
the continuous action expert is introduced. We directly fine-tune \molmot and
\molmoer with the same discrete-action architecture on the single \libero Long
task suite, using action tokens produced by \molmoactfast and no continuous action expert. Both
models are trained for 60{,}000 steps with the same configuration. As shown in
\autoref{tab:ablation-er-backbone}, replacing \molmot with \molmoer improves
success from 77.6\% to 83.6\%, a 6.0-point gain. This confirms that the
embodied-reasoning specialization is not only useful for VLM benchmarks, but
also transfers directly into action-token prediction.

\begin{table}[h]
\centering
\caption{\textbf{Backbone ablation on \libero Long.} Both variants are trained
directly from the VLM checkpoint with only discrete action prediction using
\molmoactfast. The final row is marked in pink, which is what we initialize with
for \molmoacttpre.}
\begin{tabular}{lcc}
\textbf{Backbone} & \textbf{Action objective} & \textbf{Long} \\
\midrule
\molmot~\citep{clark2026molmo2openweightsdata} & Discrete & 77.6\% \\
{\color{ai2pink}\molmoer} & Discrete & \textbf{83.6\%} \\
\end{tabular}
\label{tab:ablation-er-backbone}
\end{table}

\paragraph{VLM-to-expert conditioning.}
We next ablate how the continuous action expert receives context from the VLM.
All variants start from \molmoacttpre, attach an action expert directly, and
use the same \libero training configuration. We compare hidden-state
conditioning, standard per-layer KV conditioning, and a per-head
per-layer KV conditioning variant. In the standard design, the VLM key and value heads at
each token are flattened before learned projections map them into the action
expert cross-attention space. In the per-head variant, the VLM keys and values remain
separated by head, and each head is projected into the corresponding
action-expert head. This preserves the head structure, but requires the number
of VLM KV heads to match the number of action-expert KV heads; we use 8 for
both in all settings. The standard per-layer KV conditioning is strongest on
average, reaching 95.9\%, compared with 94.8\% for the per-head variant and
94.0\% for hidden-state conditioning
(\autoref{tab:ablation-connection-source}).
Hidden-state conditioning is competitive on Spatial, but it drops more on
Object, Goal, and Long. This supports the architectural choice in
Sec.~\ref{sec:posttraining-action-expert}, where the action expert consumes the
same attention state used by the VLM rather than a single residual-stream
representation.

\begin{table}[h]
\centering
\caption{\textbf{Ablation of the VLM-to-action-expert conditioning source on
\libero.} All variants are trained from \molmoacttpre with the same
configuration. \textbf{Bold} denotes the best result per column. The final row is marked in pink, which is what we use in all \molmoactt and \molmothink post-training and fine-tuning runs.}
\begin{tabular}{lccccc}
\textbf{Conditioning source} & \textbf{Spatial} & \textbf{Object} & \textbf{Goal} & \textbf{Long} & \textbf{Average} \\
\midrule
Hidden-state conditioning & \textbf{97.0\%} & 95.4\% & 96.6\% & 87.0\% & 94.0\% \\
Per-head per-layer KV conditioning & 95.8\% & 97.2\% & 96.6\% & 89.6\% & 94.8\% \\
\color{ai2pink}{Per-layer KV conditioning} & 96.2\% & \textbf{99.0\%} & \textbf{98.6\%} & \textbf{89.8\%} & \textbf{95.9\%} \\
\end{tabular}
\label{tab:ablation-connection-source}
\end{table}

\paragraph{Multiple flow samples.}
We then vary the number of flow samples \(K\) per action chunk while holding
per-layer KV conditioning and all other training settings fixed.
These runs also start from \molmoacttpre without a prior post-training stage.
\autoref{tab:ablation-flow-steps} shows that increasing \(K\) generally
improves average performance. \(K=1\) is the weakest overall at 94.15\%,
while \(K=8\) gives the best average at 95.90\%.
The effect is not monotonic on every suite, because Long benefits strongly from
two flow samples, but the aggregate trend favors denser supervision along the flow
trajectory.

\begin{table}[h]
\centering
\caption{\textbf{Ablation of the number of flow samples on \libero.} All rows
use per-layer KV conditioning and are trained from \molmoacttpre with the same
configuration. \textbf{Bold} denotes the best result per column. The final row is marked in pink, which is what we stick to in all of \molmoactt and \molmothink fine-tuning runs.}
\begin{tabular}{cccccc}
\textbf{\(K\)} & \textbf{Spatial} & \textbf{Object} & \textbf{Goal} & \textbf{Long} & \textbf{Average} \\
\midrule
1 & 95.0\% & 95.8\% & 98.2\% & 87.6\% & 94.15\% \\
2 & \textbf{96.6\%} & 96.0\% & 95.4\% & \textbf{92.2\%} & 95.05\% \\
4 & 96.2\% & 98.4\% & 97.0\% & 89.0\% & 95.15\% \\
\color{ai2pink}8 & 96.2\% & \textbf{99.0\%} & \textbf{98.6\%} & 89.8\% & \textbf{95.90\%} \\
\end{tabular}
\label{tab:ablation-flow-steps}
\end{table}

\paragraph{Fine-tuning design.}
\autoref{tab:ablation-training-design} compares the final \molmoactt fine-tuning recipe
against targeted alternatives along three axes: whether we co-train the
discrete autoregressive action loss, whether we apply knowledge insulation to
the flow-matching loss, and which parameters are updated. Removing discrete
action co-training yields a similar average but shifts performance across
suites, improving Long while reducing Spatial and Object. Adding knowledge
insulation is also close but slightly worse than the final recipe. LoRA remains
strong, especially on Spatial, but loses 2.8 points on Long relative to full
fine-tuning. Tuning only the action expert is the clearest failure mode,
dropping the average to 93.05\%. Overall, full-model adaptation with discrete
and continuous action co-training provides the best average outcome.

\begin{table}[h]
\centering
\caption{\textbf{Ablation of \molmoactt fine-tuning design choices on
\libero.} Component columns indicate whether each design choice is enabled:
\textcolor{forestgreen}{\cmark} denotes enabled and \textcolor{darkred}{\xmark}
denotes disabled. \textbf{Bold} denotes the best result per column. The final row is the \molmoacttlib recipe, where its training design is marked in pink, which is what we stick to in all of \molmoactt fine-tuning runs.}
\resizebox{0.88\linewidth}{!}{%
\begin{tabular}{@{}cclccccc@{}}
\textbf{\makecell{Discrete\\co-training}} & \textbf{\makecell{Knowledge\\insulation}} & \textbf{Training type} & \textbf{Spatial} & \textbf{Object} & \textbf{Goal} & \textbf{Long} & \textbf{Average} \\
\midrule
\textcolor{darkred}{\xmark} & \textcolor{darkred}{\xmark} & Full fine-tuning & 95.8\% & 98.6\% & \textbf{98.4\%} & \textbf{95.0\%} & 96.95\% \\
\textcolor{forestgreen}{\cmark} & \textcolor{forestgreen}{\cmark} & Full fine-tuning & 96.8\% & 99.4\% & 97.2\% & 94.8\% & 97.05\% \\
\textcolor{forestgreen}{\cmark} & \textcolor{darkred}{\xmark} & LoRA & \textbf{98.0\%} & 99.4\% & 97.2\% & 90.4\% & 96.25\% \\
\textcolor{darkred}{\xmark} & \textcolor{darkred}{\xmark} & Action expert only & 91.6\% & 97.6\% & 93.8\% & 89.2\% & 93.05\% \\
\textcolor{ai2pink}{\cmark} & \textcolor{ai2pink}{\xmark} & {\color{ai2pink}Full fine-tuning} & 97.8\% & \textbf{100.0\%} & 97.8\% & 93.2\% & \textbf{97.20\%} \\
\end{tabular}%
}
\label{tab:ablation-training-design}
\end{table}

\paragraph{Depth-aware fine-tuning.}
Finally, we ablate the \molmothink depth fine-tuning recipe. The baseline
uses both 10\% depth-token noise and the learned per-layer depth gate described
in Sec.~\ref{sec:think-training}, while uniformly mixing action-only and
depth-and-action examples. Removing the depth-token noise and per-layer depth
gate reduces the average from 98.10\% to 97.65\%, mostly through a 1.8-point
drop on Goal. Removing mixed training as well, so that training uses only
depth-and-action examples, reduces the average further to 97.50\%. These
results show that the depth pathway is most useful when it is regularized for
imperfect inference-time depth predictions and when the policy retains a strong
action-only path.

\begin{table}[h]
\centering
\caption{\textbf{Ablation of \molmothink depth fine-tuning choices on
\libero.} Mixed training denotes uniform sampling of action-only and
depth-and-action examples. \textcolor{forestgreen}{\cmark} denotes enabled and
\textcolor{darkred}{\xmark} denotes disabled. \textbf{Bold} denotes the best result per column. The final all-enabled row is the \molmothinklibero recipe, where its training design is marked in pink, which is what we stick to in all of \molmothink fine-tuning runs.}
\resizebox{0.7\linewidth}{!}{%
\begin{tabular}{@{}cccccccc@{}}
\textbf{\makecell{Mixed\\training}} & \textbf{\makecell{Noise\\injection}} & \textbf{\makecell{Depth\\gate}} & \textbf{Spatial} & \textbf{Object} & \textbf{Goal} & \textbf{Long} & \textbf{Average} \\
\midrule
\textcolor{darkred}{\xmark} & \textcolor{darkred}{\xmark} & \textcolor{darkred}{\xmark} & 97.6\% & 99.6\% & 96.6\% & \textbf{96.2\%} & 97.50\% \\
\textcolor{forestgreen}{\cmark} & \textcolor{darkred}{\xmark} & \textcolor{darkred}{\xmark} & \textbf{98.8\%} & \textbf{99.8\%} & 96.6\% & 95.4\% & 97.65\% \\
\textcolor{ai2pink}{\cmark} & \textcolor{ai2pink}{\cmark} & \textcolor{ai2pink}{\cmark} & \textbf{98.8\%} & \textbf{99.8\%} & \textbf{98.4\%} & 95.4\% & \textbf{98.10\%} \\
\end{tabular}%
}
\label{tab:ablation-depth-training}
\end{table}

\subsection{Inference speed}
\label{speed}
We measure end-to-end action-generation latency on LIBERO using a single H100 GPU and an action horizon of 10, and report the amortized control rate as action horizon divided by latency. We compare three inference paths: the original implementation, an optimized eager path using reusable-cache optimizations, and the same optimized path with CUDA Graph replay enabled.

\begin{figure}[t]
  \centering
  \includegraphics[width=0.78\linewidth]{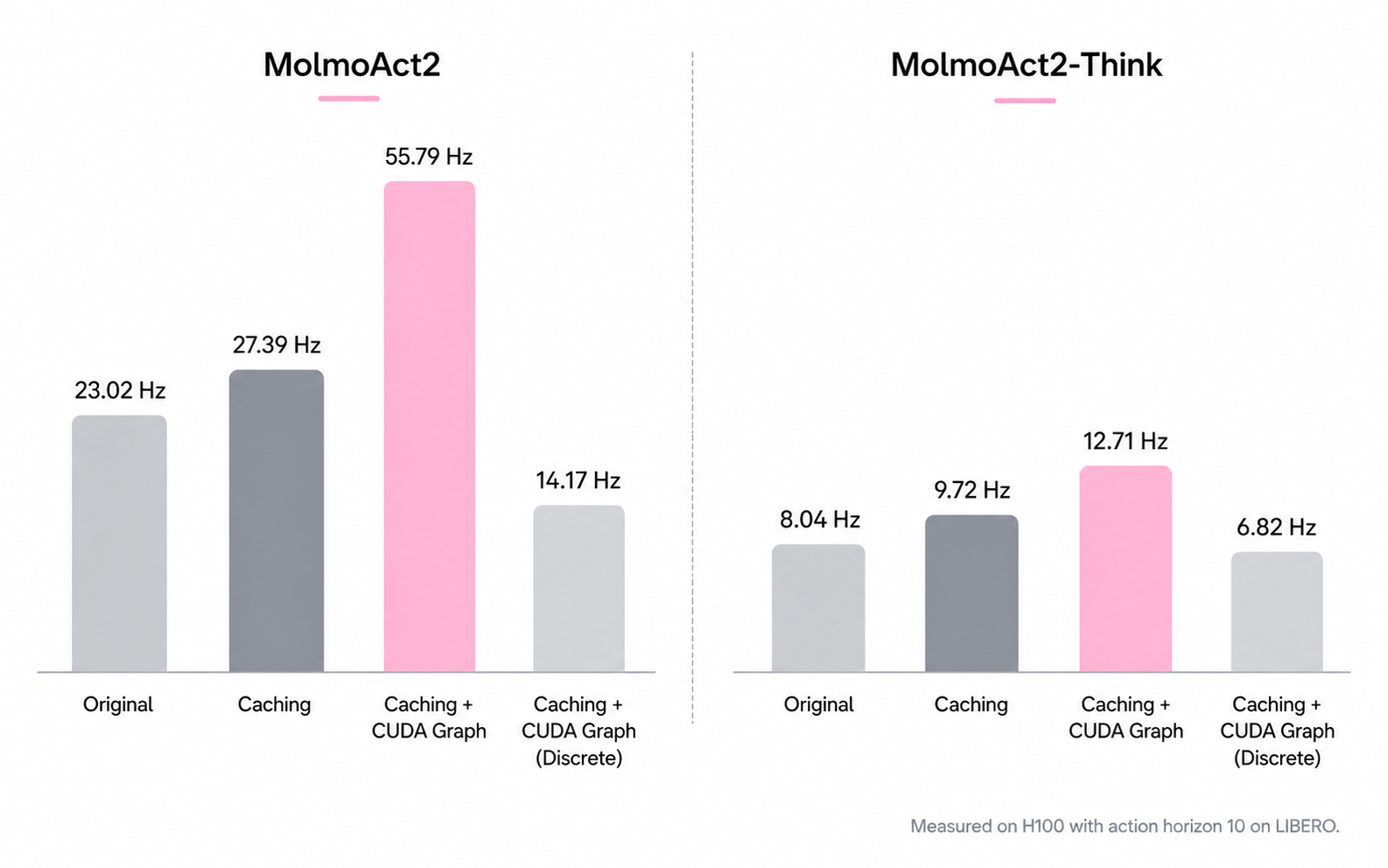}
  \caption{\textbf{Inference control rate after caching and CUDA Graph optimizations.} Control rate is action horizon divided by end-to-end action-generation latency; measurements use horizon 10 on LIBERO
with a single H100.}
  \label{fig:inference-speedup}
\end{figure}

As shown in \autoref{fig:inference-speedup}, caching alone improves \molmoactt from 23.02 Hz to 27.39 Hz and \molmothink from 8.04 Hz to 9.72 Hz. Enabling CUDA Graphs gives the larger gain: \molmoactt reaches 55.79 Hz, a 2.42\(\times\) speedup over the original path, while \molmothink reaches 12.71 Hz, a 1.58\(\times\) speedup. \molmoactt benefits substantially from CUDA Graph replay: its fixed-shape flow-matching steps form a regular, repeated computation pattern whose runtime is dominated by kernel launch overhead, which graph replay largely eliminates. \molmothink sees a smaller gain, since its adaptive-depth stage performs autoregressive decoding, whose sequential dependencies and variable-length execution are less amenable to graph capture.

\molmoactt also exposes a discrete action path, in which the VLM autoregressively decodes action tokens for the entire action chunk. We measured this path under the same optimized graph as above. \molmoactt runs at 14.17~Hz and \molmothink at 6.82~Hz i.e., \(3.94\times\) and \(1.86\times\) slower than the corresponding continuous path. This gap comes from the large VLM decoding overhead, while the action expert emits the entire chunk in a small fixed number of flow matching steps through a smaller cross-attention head, with the VLM cache reused across steps. We therefore use the continuous path as the default deployment option.

\section{Related work}
\label{sec:related}

\paragraph{Generalist robot manipulation policies.}
In recent years, robotic manipulation research has shifted focus from task-specific designs toward developing generalist policies capable of operating across diverse scenarios and embodiments~\citep{berscheid2019robot,brohan2022rt,dasari2019robonet,ebert2021bridge,fang2023rh20t,jang2022bc,khazatsky2024droid,mandlekar2018roboturk,walke2023bridgedata,shafiullah2023bringing}.
A representative approach is the development of vision-language-action (VLA) models.
In this paradigm, vision-language models (VLMs)---pretrained on vast, web-scale multimodal data (e.g., image, video, text)---are further trained on collection of diverse robotics datasets~\citep{black2410pi0, brohan2022rt,zheng2024tracevla,zhao2025cot,team2025gemini,li2024cogact,qu2025spatialvla,kim2024openvla}.
This knowledge transfer significantly reduces the reliance on domain-specific robotics data, which is typically expensive to collect via teleoperation.
However, with the notable exception of our previous work, MolmoAct~\citep{lee2025molmoactactionreasoningmodels}, reproducing frontier VLA models (e.g., the $\pi$ series~\citep{black2410pi0,intelligence2025pi05visionlanguageactionmodelopenworld})
remains highly challenging due to closed-source training data and code.
\molmoact{} represents our latest effort to democratize VLA research by releasing full training code, data, and model checkpoints that achieve state-of-the-art performance across a wide range of robotic manipulation benchmarks.

\paragraph{Embodied reasoning for robotic manipulation.}
While textual chain-of-thought (CoT) prompting~\citep{wei2022chain} has advanced reasoning in domains like program synthesis and mathematics, its benefits for robotic manipulation remain limited, likely because tacit physical knowledge is rarely articulated in the text corpora used to train backbone LLMs.
This has motivated a growing body of work on non-textual `embodied CoT' that leverages visual or spatial representations, including object bounding boxes and points~\citep{zawalski2024robotic,yuan2024robopoint,li2025hamster}, future image predictions~\citep{zhao2025cot}, latent reasoning~\citep{tur2026recurrent}, and 2D/3D point trajectories~\citep{sun2024emma,huang2025thinkact,lee2025molmoactactionreasoningmodels}.
Closest to our adaptive depth approach is PEEK~\citep{zhang2025peek}, which instead masks the RGB observation and conditions on visual tracing to learn a low-level policy.
A common drawback of these methods is heavy token consumption before action generation, which severely degrades inference latency.
\molmothink addresses this with adaptive depth reasoning: new depth tokens are computed only for novel visual signals, substantially improving latency, especially in third-person viewpoint setups.

\paragraph{Bridging VLM and action expert.}
While the architecture of VLA models has evolved significantly, the interface between VLM and action generators remains an open research problem.
Recent approaches either fully discretize actions via vector quantization~\citep{brohan2022rt,kim2024openvla,pertsch2025fast} or append a continuous generative action expert with diffusion~\citep{chi2025diffusion} or flow-matching objective~\citep{Lipman2022FlowModeling,black2410pi0}.
Standard continuous action experts typically condition only on the VLM's final hidden states.
\molmoactt{} departs from this by introducing a deep, layer-wise conditioning interface.
After pre-training the VLM on discrete action tokens produced by \molmoactfast for seamless multimodal data mixing, we post-train a continuous flow-matching expert using per-layer KV conditioning.
Unlike final-layer conditioning ~\citep{gr00tn1_2025,wang2026vla}, this per-layer KV conditioning gives the action expert dense access to the backbone's hierarchical visual-semantic features.
Finally, to explicitly ground this reasoning in 3D space without bottlenecking latency, \molmoactt-Think precedes this continuous action generation with an adaptive generation of discrete depth tokens only for dynamic scene regions.

\section{Conclusion}
\label{sec:conclude}

We introduced \molmoactt, a family of fully open action reasoning models built for real-world deployment across heterogeneous robot platforms. Building upon a spatially specialized Molmo2-ER backbone, \molmoactt produces performant and geometrically grounded behaviors across diverse manipulation tasks. We additionally release \molmoactt-Think, a thinking variant equipped with adaptive depth reasoning that delivers interpretable, depth-aware control with efficient inference. Our evaluations across simulation and real-world settings demonstrate that \molmoactt consistently outperforms strong VLA baselines out-of-the-box, fine-tunes efficiently from a handful of demonstrations, and transfers across three structurally different embodiments spanning the low-to-medium cost range. We release all model weights, training code, and data, including \molmoacttyamdata, the largest open bimanual manipulation dataset to date, alongside \molmoacttdroiddata and \molmoacttsodata, to enable reproducibility and foster community-driven research toward open foundation models that researchers and practitioners can both build on and deploy in the real world.

\clearpage

\section*{Author contributions}
\label{sec:contrib}
This project was made possible through the equal contributions of both co-first authors, listed in no particular order:
\begin{itemize}
    \item \textbf{Haoquan Fang}: Led the design and implementation of the \molmoactt model, training, and inference pipelines and infrastructure; contributed to data curation, evaluations, and writing.
    \item \textbf{Jiafei Duan}: Led the project and core method design; proposed and curated \molmoacttyamdata; led paper writing and the design of all simulation and real-world evaluations.
\end{itemize}

All other contributors are also deeply appreciated for their effort, which is critical to the success of the \molmoactt project. As not all of these can be captured, we indicate their primary contributing role in \molmoactt:
\begin{itemize}
    \item \textbf{Donovan Clay}: Led the training and curation of \molmoactfast
    \item \textbf{Sam Wang}: Led the curation and filtering of the pre-training data mixture for \molmoactt
    \item \textbf{Weikai Huang}: Led the curation and training of \molmoer
    \item \textbf{Xiang Fan}: Led the development of \molmothink
    \item \textbf{Shirui Chen}: Led the curation of \molmoacttsodata
    \item \textbf{Shanli Xing}: Led the inference optimization of \molmoactt and \molmothink
    \item For real-world robot infrastructure: Shuo Liu, Wei-Chuan Tsai, Ying-Chun Lee, Shanli Xing, Angad Wadhwa and Rose Hendrix
    \item For real-world data collection, curation, and evaluation: Jiafei Duan,  Donovan Clay, Angad Wadhwa, Suveen Ellawela, Lucas Ngoo, Cole Harrison and Sam Wang
    \item For simulation training and evaluation: Yi Ru Wang, Haoquan Fang, Jaemin Cho, Jae Sung Park, Ainaz Eftekhar, and Peter Sushko
    \item For paper writing and figures: Jiafei Duan, Haoquan Fang, Zhongzheng Ren,  Shirui Chen, Jaemin Cho, Cole Harrison, Donovan Clay, Sam Wang, Shuo Liu, Xiang Fan, Winson Han, and Eli VanderBilt
    \item For project management: Karen Farley.
    \item For research advisory: Ranjay Krishna, Dieter Fox, Joyce Chai, Zhongzheng Ren, and Ali Farhadi.
    \item Project PI: Ranjay Krishna
\end{itemize}

\section*{Acknowledgment}

This work would not be possible without the support of our colleagues at Ai2:

\begin{itemize}
    \item We thank Christopher Clark, Abhay Deshpande, Yejin Kim, Max Argus, Ishneet Sukhvinder Singh and Wilbert Pumacay for helpful research discussions and sharing of relevant findings across related projects.
    \item We thank for David Albright, Crystal Nam, Kristin Cha, Cailin Brashear, Jae Pak, Sophie Lebrecht, Kyle Wiggers, Kelsey MacMillan, Katie Morigi, Peter Clark and Megan Bartot for project management, support to robot room and publicity of \molmoactt
    \item We thank Yoganand Chandrasekhar, Johann Dahm, Michael Wilson, Fangzhou Hu, and Caroline Wu for their work on the Ai2 cluster.

\end{itemize}

\molmoactt would not have been possible without the support of many other institutions. In particular, we thank Cirrascale for their
on-going support of Ai2’s cluster. Jiafei Duan is supported by the Agency for Science, Technology and Research (A*STAR) National
Science Fellowship.

\clearpage

\bibliographystyle{abbrvnat}
\bibliography{reference}

\clearpage
\appendix

\section*{Appendix}
The appendix includes the following sections:
\begin{itemize}
\itemsep0em
    \item \S\ref{supp:model} - Model Details
    \item \S\ref{supp:training} - Training Details
    \item \S\ref{supp:eval} - Evaluation Details
    \item \S\ref{supp:data} - Datasets Details
    \item \S\ref{supp:limit} - Limitations and Potential Solutions

\end{itemize}

\section{Model Details}
\label{supp:model}

This appendix expands the model description in Sec.~\ref{sec:pretraining} and
Sec.~\ref{sec:posttraining}. \molmoactt is built in two architectural stages.
First, \molmoacttpre adapts the \molmoer vision-language backbone into a
discrete autoregressive robot policy. It keeps the Molmo2 visual and text token
interface, but adds robot-specific state, setup/control, and action-output
tokens so that robot trajectories can be trained with the same next-token
objective as vision-language data. Second, post-training attaches a continuous
DiT-style action expert to this already robot-aware backbone. The expert uses
per-layer KV conditioning on the VLM and learns a flow-matching velocity field
over continuous robot action chunks. \molmothink keeps the same architecture and
adds an autoregressive depth-token prefix before action prediction.

\subsection{\molmoactt Backbone}
\label{supp:model-backbone}

\paragraph{Backbone initialization and visual inputs.}
\molmoactt initializes from \molmoer, a Molmo2-based VLM specialized for
embodied and spatial reasoning. The backbone therefore follows the Molmo2
architecture~\citep{clark2026molmo2openweightsdata}: visual inputs are encoded
by a SigLIP2 vision transformer, pooled and projected by a vision-language
connector, and consumed by an autoregressive language model together with text.
The important MolmoAct2-specific choice is how this interface is used for robot
control. Robot examples can include multiple camera views, proprioceptive state
tokens, setup/control descriptors, and action targets, so sequence length is a
central constraint. For the robot-policy stages in Sec.~\ref{sec:pretraining}
and Sec.~\ref{sec:posttraining}, each camera observation is represented as a
single resized crop rather than a tiled high-resolution image. Video examples
are sampled at up to 2 FPS and capped to at most 8 frames. Multi-camera robot
episodes are serialized as multiple image inputs, and the camera order is
randomized at the episode level during pre-training.

\paragraph{Vision encoder and connector.}
Each crop is encoded by a SigLIP2 vision transformer~\citep{tschannen2025siglip}.
The connector follows the Molmo2 design. It reads patch features from the
third-to-last and ninth-from-last ViT layers, then pools local windows with a
multi-head attention layer. For images, \(2\times2\) patch windows are pooled
into one visual token using the mean patch feature as the query. For videos, a
\(3\times3\) pooling window is used to further reduce the number of frame
tokens. The pooled features are projected into the language-model embedding
space with a shared MLP. This produces a sequence of visual tokens that can be
interleaved with text tokens.

\paragraph{Language model input.}
The language model receives visual tokens together with the robot instruction
and robot-specific text. Multi-camera observations are identified by image-index
text, and video frames use the same frame markers as Molmo2. Since MolmoAct2
robot observations use single resized crops, the model does not rely on the
multi-crop column-token machinery for robot control. Visual tokens are allowed
to forward-attend to one another, including across different frames or camera
views, so the backbone can reason over the full visual context before producing
state-conditioned robot outputs.

\paragraph{Added tokens.}
The robot interface keeps the model in the same autoregressive format as the
base VLM by adding robot-specific tokens to the backbone vocabulary. A robot
example contains visual observations, a language instruction, optional setup
and control descriptors, discrete state tokens, and an output marker that tells
the model which response type to produce. The main markers are
\texttt{\detokenize{<setup_start>}}, \texttt{\detokenize{<setup_end>}},
\texttt{\detokenize{<control_start>}}, \texttt{\detokenize{<control_end>}},
\texttt{\detokenize{<state_start>}}, \texttt{\detokenize{<state_end>}}, and
\texttt{\detokenize{<action_output>}}. We also add indexed token families for
discrete robot quantities: action tokens
\texttt{\detokenize{<action_0>}}--\texttt{\detokenize{<action_2047>}} and
state tokens
\texttt{\detokenize{<state_0>}}--\texttt{\detokenize{<state_255>}}. The
depth tokens are added only for \molmothink post-training: the model uses
\texttt{\detokenize{<depth_output>}}, \texttt{\detokenize{<depth_start>}}, and
\texttt{\detokenize{<depth_end>}}, together with depth-code tokens
\texttt{\detokenize{<depth_0>}}--\texttt{\detokenize{<depth_127>}}.

Setup and control descriptors make embodiment and action semantics explicit in
the prompt. For example, a bimanual YAM prompt can indicate that the robot is a
pair of YAM arms and that the control target is an absolute joint pose, while a
DROID prompt can indicate a Franka arm with delta end-effector control. This
lets the same tokenizer and backbone support different robots without forcing
all datasets into a single physical coordinate convention.

Continuous robot states are normalized using dataset statistics and discretized
into one of 256 state tokens per scalar dimension. The resulting state-token
sequence is appended to the prompt before the action output marker. Continuous
actions are normalized, padded to a maximum width of 32 dimensions, and encoded
by \molmoactfast into discrete action tokens with a 2048-token action
vocabulary. During pre-training, these action tokens are the only robot target.
During post-training and fine-tuning, the same example also provides a
continuous action chunk for the action expert. The discrete action-token span is
masked from the expert conditioning path, so the continuous policy cannot
condition on the ground-truth action tokens it is trained to predict.

\begin{table}[t]
\newcommand\modelvalue[1]{\cellcolor{olive!5} #1}
\newcommand\navalue{--}
\centering
\footnotesize
\setlength\tabcolsep{6pt}
\renewcommand\arraystretch{1.1}
\begin{tabular}{@{}lcccc@{}}
\toprule
\textbf{Hyperparameter} &
\textbf{\makecell{Image\\Encoder}} &
\textbf{\makecell{V/L\\Connector}} &
\textbf{LLM} &
\textbf{\makecell{Action\\Expert}} \\
\midrule
Params & \modelvalue{380M} & \modelvalue{57M} & \modelvalue{4.0B} & \modelvalue{621M} \\
Dim & \modelvalue{1152} & \modelvalue{1152} & \modelvalue{2560} & \modelvalue{768} \\
MLP Dim & \modelvalue{4304} & \modelvalue{9728} & \modelvalue{9728} & \modelvalue{3072} \\
Act. & \modelvalue{GELU} & \modelvalue{SwiGLU} & \modelvalue{SwiGLU} & \modelvalue{SwiGLU} \\
Heads & \modelvalue{16} & \modelvalue{16} & \modelvalue{32} & \modelvalue{8} \\
KV Heads & \modelvalue{16} & \navalue & \modelvalue{8} & \navalue \\
Layers & \modelvalue{27} & \navalue & \modelvalue{36} & \modelvalue{36} \\
Dropout & \modelvalue{0.0} & \modelvalue{0.0} & \modelvalue{0.1} & \modelvalue{0.0} \\
\midrule
Image Size & \modelvalue{\(384{\times}384\)} & \navalue & \navalue & \navalue \\
Patch Size & \modelvalue{14} & \navalue & \navalue & \navalue \\
Image Pool Size & \navalue & \modelvalue{\(2{\times}2\)} & \navalue & \navalue \\
Video Pool Size & \navalue & \modelvalue{\(3{\times}3\)} & \navalue & \navalue \\
Pool Dim & \navalue & \modelvalue{1152} & \navalue & \navalue \\
Pool Heads & \navalue & \modelvalue{16} & \navalue & \navalue \\
Embed & \navalue & \navalue & \modelvalue{151936} & \navalue \\
Theta & \navalue & \navalue & \modelvalue{1M} & \navalue \\
\midrule
Max Horizon & \navalue & \navalue & \navalue & \modelvalue{30} \\
Max Action Dim & \navalue & \navalue & \navalue & \modelvalue{32} \\
Time Embed Dim & \navalue & \navalue & \navalue & \modelvalue{256} \\
Conditioning & \navalue & \navalue & \navalue & \modelvalue{Per-layer KV} \\
\bottomrule
\end{tabular}
\caption{\textbf{\molmoactt architecture hyperparameters.} The image encoder,
connector, and LLM columns follow the 4B \molmoer backbone used by
\molmoactt. The action-expert column describes the MolmoAct2 continuous
flow-matching module used for continuous action prediction.}
\label{tab:hyperparams_model}
\end{table}

\subsection{Continuous Action Expert}
\label{supp:model-action-expert}

\paragraph{Input and output.}
The action expert is a noncausal transformer over a short action chunk. It maps
a noisy normalized trajectory
\(x_t \in \mathbb{R}^{H \times D_{\max}}\) to a velocity prediction of the same
shape, where \(H\) is the action horizon and \(D_{\max}=32\) is the shared
maximum action width. The released \molmoactt checkpoint uses \(H=30\) for
post-training, while the \libero fine-tuned checkpoints use \(H=10\). The
corresponding padding masks remove padded action steps and padded action
dimensions from both the noisy input and the loss.

\paragraph{Flow-matching objective.}
Let \(a\) be a normalized target action chunk and let
\(\epsilon \sim \mathcal{N}(0,I)\). For a sampled time \(t \in [0,1]\), we
construct
\begin{equation}
    x_t = (1-t)\epsilon + t a,
    \qquad
    u^\star = a - \epsilon .
\end{equation}
The action expert \(f_\theta\) predicts \(u^\star\) from the noisy trajectory,
the time embedding, and the VLM context \(c\):
\begin{equation}
    \mathcal{L}_{\mathrm{flow}}
    =
    \mathbb{E}_{a,\epsilon,t}
    \left[
    \left\|
    m \odot \left(f_\theta(x_t,t,c) - u^\star\right)
    \right\|_2^2
    \right],
\end{equation}
where \(m\) masks padded action steps and padded action dimensions. In
post-training, we evaluate multiple sampled flow-matching times for each robot
action chunk while reusing the same VLM context. At inference, we initialize
the trajectory from Gaussian noise and integrate the learned velocity field for
a fixed number of Euler steps:
\begin{equation}
    z_{i+1} = z_i + \Delta t \, f_\theta(z_i,t_i,c),
    \qquad
    t_i = \frac{i}{N},
    \qquad
    \Delta t = \frac{1}{N}.
\end{equation}
The released checkpoints use \(N=10\) inference steps. The final normalized
trajectory is sliced to the embodiment-specific action width and unnormalized
with the corresponding dataset statistics.

\paragraph{Expert block.}
The action expert has one transformer block for each VLM layer, giving
\(L=36\) expert blocks. A noisy action chunk is first projected from 32
continuous dimensions to the expert hidden width. Each block contains
bidirectional self-attention over the action chunk, cross-attention to the VLM
context, and an MLP. A sinusoidal time embedding is passed through a small MLP,
then used to produce DiT-style shift, scale, and gate parameters for all three
residual branches. Schematically, block \(\ell\) computes
\begin{align}
    h'_\ell
        &= h_\ell
        + g^{\mathrm{sa}}_\ell
        \operatorname{SA}\!\left(
        \operatorname{AdaRMS}^{\mathrm{sa}}_\ell(h_\ell,t)
        \right), \\
    \bar{h}_\ell
        &= h'_\ell
        + g^{\mathrm{ca}}_\ell
        \operatorname{CA}\!\left(
        \operatorname{AdaRMS}^{\mathrm{ca}}_\ell(h'_\ell,t),
        \tilde{K}_\ell,
        \tilde{V}_\ell
        \right), \\
    h_{\ell+1}
        &= \bar{h}_\ell
        + g^{\mathrm{ff}}_\ell
        \operatorname{MLP}\!\left(
        \operatorname{AdaRMS}^{\mathrm{ff}}_\ell(\bar{h}_\ell,t)
        \right).
\end{align}
Here, \(\operatorname{AdaRMS}\) denotes RMS normalization followed by the
time-dependent affine modulation. The self-attention and cross-attention layers
use query-key normalization. The expert uses rotary position embeddings for the
action sequence, which gives the denoising transformer an explicit ordering
over the predicted trajectory steps. After the final expert block, a modulated
RMS normalization layer and a linear projection map the hidden states back to
32 action dimensions. The final projection is initialized to produce near-zero
velocity predictions at the beginning of post-training, which makes the newly
attached expert easier to optimize.

\paragraph{Per-layer KV conditioning.}
The action expert receives VLM context through per-layer KV conditioning. For
VLM layer \(\ell\), let \((K^{\mathrm{vlm}}_\ell,V^{\mathrm{vlm}}_\ell)\) be
the key and value tensors produced by that layer's self-attention over the
prompt, images, state tokens, and non-target robot text. The expert uses learned
adapter projections \(P_K\) and \(P_V\) to map these tensors into the expert
cross-attention width:
\begin{equation}
    \tilde{K}_\ell =
    \operatorname{reshape}\!\left(P_K K^{\mathrm{vlm}}_\ell\right),
    \qquad
    \tilde{V}_\ell =
    \operatorname{reshape}\!\left(P_V V^{\mathrm{vlm}}_\ell\right).
\end{equation}
Expert block \(\ell\) then cross-attends to
\((\tilde{K}_\ell,\tilde{V}_\ell)\):
\begin{equation}
    \operatorname{CA}(Q_\ell,\tilde{K}_\ell,\tilde{V}_\ell)
    =
    \operatorname{softmax}\!\left(
        \frac{Q_\ell \tilde{K}_\ell^\top}{\sqrt{d_h}}
    \right)\tilde{V}_\ell .
\end{equation}
This gives each expert block direct access to the attention state at the same
depth in the VLM. In the released architecture, the VLM has 8 KV heads with
head dimension 128, so each token contributes a 1024-dimensional key and value
state. The adapter projections map this state to the action expert width of
768, which is reshaped into 8 expert attention heads of width 96. In
post-training, this conditioning path is detached from the VLM for the flow
loss. The language-model loss still updates the VLM, while the flow loss trains
the action expert and the VLM-to-expert adapter projections.

\subsection{Adaptive-Depth Extension}
\label{supp:model-depth}

\molmothink keeps the same backbone and action expert, but adds an
autoregressive depth-token interface before continuous action prediction. A
depth response is requested with \texttt{\detokenize{<depth_output>}} and
serialized between \texttt{\detokenize{<depth_start>}} and
\texttt{\detokenize{<depth_end>}}. The released depth checkpoints predict a
\(10\times10\) depth-token grid, giving 100 depth positions, and each position
takes one of 128 learned depth-code values. For depth-only examples, the
assistant response begins with
\texttt{\detokenize{<depth_output>}} and the model predicts depth tokens. For
depth-and-action examples, the response begins with
\texttt{\detokenize{<depth_output><action_output>}}. The depth tokens are
generated autoregressively, then included in the context used by the action
expert.

The \libero depth fine-tuned model also includes a learned depth gate on the
expert conditioning path. For action-expert layer \(\ell\), let \(M_t=1\)
denote positions belonging to the depth-output trigger, depth delimiters, or
depth-code tokens, and let \(A_t\) denote valid context positions. The gate is
computed from the non-depth context of the corresponding VLM layer,
\begin{equation}
    c_\ell
    =
    \frac{
        \sum_t A_t(1-M_t)V^{\mathrm{vlm}}_{\ell,t}
    }{
        \sum_t A_t(1-M_t)
    },
    \qquad
    g_\ell
    =
    \sigma(w_\ell^\top c_\ell + b_\ell),
\end{equation}
and then applied only to depth-token keys and values:
\begin{equation}
    \bar{K}^{\mathrm{vlm}}_{\ell,t}
    =
    \left(1-M_t+M_t g_\ell\right)K^{\mathrm{vlm}}_{\ell,t},
    \qquad
    \bar{V}^{\mathrm{vlm}}_{\ell,t}
    =
    \left(1-M_t+M_t g_\ell\right)V^{\mathrm{vlm}}_{\ell,t}.
\end{equation}
The gated \((\bar{K}^{\mathrm{vlm}}_{\ell},\bar{V}^{\mathrm{vlm}}_{\ell})\)
are then projected into the action expert as in \autoref{sec:posttraining}.
The gate is initialized with bias \(-4\), so fine-tuning begins close to the
standard action-conditioning path and learns how strongly each expert layer
should use the depth prefix.

\clearpage

\section{Training Details}
\label{supp:training}

\subsection{Implementation}
\paragraph{Training infrastructure.}
Our training implementation mainly follows Molmo2. We train in PyTorch with
Fully Sharded Data Parallel v2 (FSDP2), and use PyTorch's scaled dot-product
attention implementation rather than FlashAttention because our packed
multimodal sequences and robot-conditioning masks require custom attention
masks. We use \texttt{torch.compile} for throughput and keep the ViT and LLM
shapes static through padding and fixed sequence budgets, which allows the
compiled graph to be reused efficiently across training steps.

We use automatic mixed precision with \texttt{bfloat16} for most operations,
while keeping numerically sensitive operations such as layer normalization and
RoPE in full precision. Gradients are computed on local minibatches and then
averaged across devices. For token-level losses, each device divides by the
average number of loss tokens across all devices rather than by its local
number of loss tokens. This avoids over-weighting examples with short target
spans, which is important because our mixtures contain both short robot-action
targets and longer language or video examples. During fine-tuning, dataset
mixing is performed within each batch so that each update contains examples
from multiple sources. Examples longer than the stage-specific maximum
sequence length are truncated; this affects fewer than 0.1\% of examples and
usually occurs for long video examples with subtitles or dense annotations. We
find training stable under this setup, without loss spikes or NaNs.

\paragraph{Packing.}
Our packing implementation also follows Molmo2. Each data-loader worker keeps
a pool of \(M=48\) preprocessed and tokenized examples. When the pool is not
full, new examples are drawn from the training mixture and added to the pool.
Once the pool is full, a dynamic-programming solver selects the subset that
maximizes
\begin{equation}
    T + w_i I
    \quad\text{subject to}\quad
    T \leq T_{\max},\ I \leq I_{\max},
\end{equation}
where \(T\) is the total number of text tokens, \(I\) is the total number of
image crops, and \(w_i=30\) balances token and image utilization. The selected
examples are yielded as one packed sequence and removed from the pool. In the
Molmo2 setting, \(T_{\max}=16384\) and \(I_{\max}=128\), while long-context
training uses \(T_{\max}=36864\) and \(I_{\max}=384\). For \molmoactt, the
same solver is used with the stage-specific sequence budgets described in
\autoref{sec:pretraining-recipe} and \autoref{sec:posttraining-recipe}. In
practice, we solve a quantized version of the packing problem by rounding token
counts to the nearest multiple of 32.

Increasing the pool beyond 48 examples gives diminishing returns in packing
efficiency. The method is also generally robust to \(w_i\), although values
that are too small can leave the pool dominated by examples with many image
crops, which are difficult to pack with other examples. Implementing packing
inside PyTorch's \texttt{DataLoader} makes it easy to use, since each worker
runs the algorithm independently. This adds some overhead when many workers are
used, but in practice data loading is still dominated by video decoding and
frame extraction.

\paragraph{Image augmentation.}
We use the same image augmentation scheme in all of our robot pre-training, post-training, and fine-tuning runs. The image augmentation is applied to all images and videos during training, not limited to robot data. Augmentation is
disabled for evaluation and inference.

The image augmentation applies the stochastic transforms before image
normalization and final resizing. In simplified form, the recipe is:

\begin{center}
\begin{minipage}{0.78\linewidth}
\begin{lstlisting}[language=Python,basicstyle=\ttfamily\footnotesize,breaklines=true,frame=single,framerule=0.4pt]
import torchvision.transforms as T

height, width = image.height, image.width

transform = T.Compose([
    T.RandomCrop(size=(int(height * 0.95), int(width * 0.95))),
    T.Resize((height, width)),
    T.RandomRotation(degrees=5),
    T.ColorJitter(
        brightness=0.2,
        contrast=(0.8, 1.2),
        saturation=(0.8, 1.2),
        hue=0.05,
    ),
    T.RandomApply(
        [T.GaussianBlur(kernel_size=5, sigma=(0.1, 1.0))],
        p=0.2,
    ),
])

image = transform(image)
\end{lstlisting}
\end{minipage}
\end{center}

\paragraph{Input prompts.}
Robot examples are formatted with one of three output styles. Standard
\molmoactt uses the action style, which asks the model to predict the robot
action for the given instruction, setup, state, and control mode.
\molmothink uses all three styles: action prediction, depth prediction, and
depth-then-action prediction. The depth-only style asks for the depth map of
the main image. The depth-then-action style asks the model to first produce the
depth representation and then produce the action, so that the continuous action
expert can condition on the input context together with the predicted depth
tokens.

After constructing the natural-language prompt, we wrap it with the same
user/assistant chat template used by the Molmo2 formatter. For robot runs, the
user message is placed between \texttt{\detokenize{<|im_start|>user}} and
\texttt{\detokenize{<|im_end|>}}, followed by the assistant prefix
\texttt{\detokenize{<|im_start|>assistant}}. We then append an output trigger
on the assistant side: action examples use
\texttt{\detokenize{<action_output>}}, depth examples use
\texttt{\detokenize{<depth_output>}}, and depth-then-action examples use
\texttt{\detokenize{<depth_output><action_output>}}. During training, the
target response follows this trigger. During inference, we provide the same
formatted user message, assistant prefix, and trigger as the generation prefix,
which selects the desired output style before autoregressive decoding or
continuous action generation begins.

Before insertion into the prompt, raw task strings are normalized into concise
natural-language task descriptions. Dataset-specific bookkeeping is removed,
punctuation and whitespace are standardized, and embodiment or control
information is placed into the separate setup and control fields rather than
being duplicated in the task text. The setup and control fields can also be
wrapped with dedicated special tokens, while the robot state is included as an
optional state clause when discrete state tokens are available. Representative
formatted prompt prefixes are:

\begin{center}
\begin{minipage}{0.88\linewidth}
\begin{lstlisting}[basicstyle=\ttfamily\footnotesize,breaklines=true,frame=single,framerule=0.4pt]
Action:
<|im_start|>user
The task is to put the red mug on the tray. The setup is <setup_start>bimanual yam robotic arms in molmoact2<setup_end>. The current state of the robot is <state_start><state_14><state_87>...<state_203><state_end>. The expected control mode is <control_start>absolute joint pose<control_end>. Given these, what action should the robot take to complete the task?
<|im_end|>
<|im_start|>assistant
<action_output>

Depth:
<|im_start|>user
The task is to open the middle drawer. The setup is <setup_start>single franka robotic arm in libero<setup_end>. The expected control mode is <control_start>delta end-effector pose<control_end>. Given these, what is the depth map of the main image?
<|im_end|>
<|im_start|>assistant
<depth_output>

Depth then action:
<|im_start|>user
The task is to place the bowl next to the plate. The setup is <setup_start>single franka robotic arm in libero<setup_end>. The current state of the robot is <state_start><state_31><state_52>...<state_118><state_end>. The expected control mode is <control_start>delta end-effector pose<control_end>. Given these, first predict the depth map of the main image and then predict the action the robot should take to complete the task?
<|im_end|>
<|im_start|>assistant
<depth_output><action_output>
\end{lstlisting}
\end{minipage}
\end{center}

\paragraph{Training hyperparameters.}
We summarize the stage-level training hyperparameters in
\autoref{tab:hyperparams_train}. The columns correspond to the three training
stages used throughout the paper: pre-training, post-training, and
embodiment-specific fine-tuning.

\begin{table}[!htbp]
\centering
\footnotesize
\setlength\tabcolsep{6pt}
\renewcommand\arraystretch{1.1}

\begin{tabular}{lccc}
\toprule
\textbf{Hyperparameter} &
\cellcolor{olive!5}\textbf{Pre-train} &
\cellcolor{olive!5}\textbf{Post-train} &
\cellcolor{olive!5}\textbf{Fine-tune} \\
\midrule
Warm-up ViT        & 200 & 200 & 200 \\
Warm-up Conn.      & 200 & 200 & 200 \\
Warm-up LLM        & 200 & 200 & 200 \\
Warm-up AE        & 200 & 200 & 200 \\
LR ViT             & $5{\times}10^{-6}$ & $5{\times}10^{-6}$ & $5{\times}10^{-6}$ \\
LR Conn.           & $5{\times}10^{-6}$ & $5{\times}10^{-6}$ & $5{\times}10^{-6}$ \\
LR LLM             & $1{\times}10^{-5}$ & $1{\times}10^{-5}$ & $1{\times}10^{-5}$ \\
LR AE             & $5{\times}10^{-5}$ & $5{\times}10^{-5}$ & $5{\times}10^{-5}$ \\
Cosine Decay       & 10\% & 10\% & 10\% \\
Eps.               & $10^{-6}$ & $10^{-6}$ & $10^{-6}$ \\
Betas              & 0.9/0.95 & 0.9/0.95 & 0.9/0.95 \\
Steps              & 200k & 100k & 100k \\
Global Batch Size  & 128 & 128 & 64 or 128 (BimanualYAM) \\
Sequence Length  & 4200 & 2100 (Robot) and 4200 (Multimodal) & 2100 \\
GPUs (H100s)       & 64 & 64 & 32 or 64 (BimanualYAM) \\
Time (Hours)       & 90 & 36 & 36 \\
GPU Hours          & 5760 & 2304 & 1152 or 2304 (BimanualYAM) \\
\addlinespace[2pt]  
\bottomrule
\end{tabular}
\caption{\textbf{\molmoactt training hyperparameters.} We summarize the
stage-level hyperparameters used for pre-training, post-training, and
embodiment-specific fine-tuning. We only show the fine-tuning hyperparameters for the training on major datasets.}
\label{tab:hyperparams_train}
\end{table}

\subsection{\molmoactfast}
The \molmoactfast architecture is an open-data implementation based on the FAST framework released by Physical Intelligence \citep{pertsch2025fast}. While we utilize the core logic of frequency-domain compression (DCT-BPE), \molmoactfast is distinguished by its use of a fully transparent training mix addressing the data opacity of prior open-weight releases.

\paragraph{Data Format}
We adopt a standardized 32-dimensional vector format to accommodate various robot morphologies. The vector is structured as follows: Single-Arm: $[A_1, \dots, A_n, G_1, 0, \dots, 0]$ where $A$ represents $n$ arm joints and $G_1$ is the gripper state. Bimanual: $[A_{L1}, \dots, A_{Ln}, G_L, A_{R1}, \dots, A_{Rn}, G_R, 0, \dots, 0]$ representing Left and Right arms respectively.

\paragraph{Multimodal Representation }: \molmoactfast does not require a unified coordinate frame (e.g., converting all data to Delta End-Effector). Instead, the tokenizer is trained on a heterogeneous mix of "dialects," including Absolute Joint positions and Delta End-Effector velocities. During training, the VLM backbone learns to associate these disparate action representations with the specific embodiment mentioned in the task prompt or visual context.

\subsection{GPU Cluster}

\molmoactt was trained on Jupiter, an Ai2 GPU cluster in Austin, Texas. \molmoactt workloads were scheduled using Beaker \citep{guerquin2022beaker}, a custom workload management system. Jupiter comprises 128 GPU nodes and is operated by Cirrascale Cloud Services\footnote{\href{https://www.cirrascale.com/}{\path{cirrascale.com}}}.

\paragraph{Compute} Jupiter provides 1{,}024 NVIDIA H100 GPUs (80GB HBM3, 700W) across 128 servers. Each server has 2,$\times$,Intel Xeon Platinum8468 CPUs, 2TB DDR5 system memory, and 18~TB local NVMe storage.

\paragraph{Storage} The servers are connected over an 800Gbps local network to a WEKA high-performance storage cluster\footnote{\href{https://www.weka.io/}{\path{weka.io}}}. The storage system provides 1PB of NVMe SSD across 11 storage servers and 5PB of HDD across 12 hosts. Each Jupiter server has two bonded 25Gbps Mellanox Ethernet NICs (50Gbps per host). In benchmarks, we achieved 761Gbps aggregate read/write throughput using 64 client machines.

\paragraph{Interconnect} Cross-node GPU communication uses RDMA over InfiniBand on a two-tier Rail-Optimized, balanced, full-bisection network \citep{wang2023rail}. Each server is equipped with eight 400Gbps InfiniBand adapters (3.2Tbps peak per host), supporting concurrent distributed jobs without topological scheduling.

\paragraph{Cooling} The servers are racked in \textit{Dynamic Density Cabinets}\footnote{\href{https://www.cirrascale.com/products-and-services/cabinet-technologies}{\path{cirrascale.com/products-and-services/cabinet-technologies}}}. Each cabinet houses five servers with dedicated cooling and power. Air circulates in a closed loop through an overhead plenum where it is cooled via heat transfer to water, enabling a datacenter PUE of~1.2. Under heavy utilization, H100 temperatures peak around $75^\circ\mathrm{C}$, with typical averages between $60^\circ\mathrm{C}$ and $65^\circ\mathrm{C}$.

\clearpage

\section{Evaluation Details}
\label{supp:eval}

\subsection{RoboEval}
\textbf{Training.} We fine-tune the \molmoacttpos checkpoint on the RoboEval dataset using 8 GPUs with a global batch size of 64 for 10{,}000 steps. All eight tasks are trained jointly in a multi-task setup, with demonstrations from every task and variation pooled into a single training distribution.

\textbf{Evaluation.} We evaluate \molmoacttpos on the eight bimanual manipulation tasks introduced in RoboEval---Cube Handover, Lift Pot, Lift Tray, Pack Box, Pick Single Book From Table, Rotate Valve, Stack Single Book Shelf, and Stack Two Blocks---each instantiated with 3--5 structured variations that systematically perturb the initial scene. Following the benchmark's protocol, every task is run under a Static configuration as well as Position (Pos), Orientation (Rot), and combined Position+Rotation (PR) perturbations of the manipulated objects, with additional task-specific variants where applicable (e.g., a Vertical configuration for Cube Handover, a Drag setup for Lift Tray that initializes the tray outside the bimanual workspace, and an obstacle-occluded valve in Rotate Valve). For each (task, variation) pair, we roll out the policy from initial states sampled from the task's distribution $\rho_0$ and score each rollout against the task's geometric success set $\mathcal{S}_{\text{success}}$, reporting the mean success rate together with its standard error across rollouts. Beyond binary success, we log the full set of RoboEval diagnostic metrics during inference: trajectory-based metrics (joint and Cartesian path length, and joint and Cartesian jerk) to characterize motion efficiency and smoothness; spatial metrics (self-collisions, environment collisions, and object slip counts) to capture contact stability and execution safety; coordination metrics (vertical end-effector height discrepancy $\Delta z$ and inter-arm velocity divergence $\Delta v$) to assess bimanual alignment and synchronization; and stagewise task-progression flags that mark completion of each semantically grounded subphase (e.g., grasp, lift, transfer, place). This combination lets us report not only whether \molmoacttpos completes each task, but also where in the task pipeline it tends to fail and how cleanly it executes when it succeeds.

\begin{table*}[htbp]
\centering
\footnotesize
\renewcommand{\arraystretch}{1.2}
\caption{\textbf{Summary of performance metrics across all methods on the combined variation.} Bold values indicate the best result for each metric within the task. Values are reported as mean $\pm$ 95\% confidence interval. The final row reports human demonstration statistics (computed from successful teleoperated demos only), which provide a reference band for efficient execution.}
\begin{adjustbox}{width=\textwidth,center}
\begin{tabular}{lr|lr|lrlrlrlrlrlrlrlrlrlrlrlrlr}
\toprule
\textbf{Method} & \multicolumn{2}{|c|}{\textbf{Success \(\uparrow\)}} & \multicolumn{2}{c}{\textbf{TP \(\uparrow\)}} & \multicolumn{2}{c}{\textbf{BAVD \(\downarrow\)}} & \multicolumn{2}{c}{\textbf{BGVD \(\downarrow\)}} & \multicolumn{2}{c}{\textbf{CPL \(\downarrow\)}} & \multicolumn{2}{c}{\textbf{CT \(\downarrow\)}} & \multicolumn{2}{c}{\textbf{ECC \(\downarrow\)}} & \multicolumn{2}{c}{\textbf{JPL \(\downarrow\)}} & \multicolumn{2}{c}{\textbf{MCJ \(\downarrow\)}} & \multicolumn{2}{c}{\textbf{MJJ \(\downarrow\)}} & \multicolumn{2}{c}{\textbf{OPL \(\downarrow\)}} & \multicolumn{2}{c}{\textbf{SCC \(\downarrow\)}} & \multicolumn{2}{c}{\textbf{SC \(\downarrow\)}} & \multicolumn{2}{c}{\textbf{TL \(\downarrow\)}} \\
\midrule
\multicolumn{29}{c}{\textbf{Cube Handover}} \\
\midrule
\texttt{Diffusion} & \multicolumn{2}{|c|}{$0.06\,\pm\,0.10$} & \multicolumn{2}{c}{$0.27\,\pm\,0.07$} & \multicolumn{2}{c}{$0.32\,\pm\,0.03$} & \multicolumn{2}{c}{$0.13\,\pm\,0.01$} & \multicolumn{2}{c}{$5.07\,\pm\,0.35$} & \multicolumn{2}{c}{$4.80\,\pm\,0.28$} & \multicolumn{2}{c}{$2.94\,\pm\,0.49$} & \multicolumn{2}{c}{$26.14\,\pm\,2.02$} & \multicolumn{2}{c}{$24.73\,\pm\,0.90$} & \multicolumn{2}{c}{$133.00\,\pm\,4.88$} & \multicolumn{2}{c}{$19.61\,\pm\,1.59$} & \multicolumn{2}{c}{$2.46\,\pm\,0.57$} & \multicolumn{2}{c}{$0.22\,\pm\,0.14$} & \multicolumn{2}{c}{$939.76\,\pm\,52.04$} \\
\texttt{GR00T} & \multicolumn{2}{|c|}{$0.00\,\pm\,0.07$} & \multicolumn{2}{c}{$0.01\,\pm\,0.01$} & \multicolumn{2}{c}{$0.50\,\pm\,0.01$} & \multicolumn{2}{c}{$0.10\,\pm\,0.01$} & \multicolumn{2}{c}{$4.69\,\pm\,0.26$} & \multicolumn{2}{c}{$9.52\,\pm\,0.26$} & \multicolumn{2}{c}{$2.90\,\pm\,0.49$} & \multicolumn{2}{c}{$33.15\,\pm\,1.11$} & \multicolumn{2}{c}{$42.62\,\pm\,1.49$} & \multicolumn{2}{c}{$235.44\,\pm\,4.45$} & \multicolumn{2}{c}{$23.34\,\pm\,0.97$} & \multicolumn{2}{c}{$25.88\,\pm\,1.91$} & \multicolumn{2}{c}{\cellcolor{ForestGreen!20}$\mathbf{0.00}\,\pm\,0.07$} & \multicolumn{2}{c}{$976.10\,\pm\,23.48$} \\
\texttt{xVLA} & \multicolumn{2}{|c|}{$0.00\,\pm\,0.07$} & \multicolumn{2}{c}{$0.04\,\pm\,0.03$} & \multicolumn{2}{c}{$0.56\,\pm\,0.04$} & \multicolumn{2}{c}{$0.15\,\pm\,0.02$} & \multicolumn{2}{c}{$5.51\,\pm\,0.68$} & \multicolumn{2}{c}{$8.50\,\pm\,3.26$} & \multicolumn{2}{c}{$3.40\,\pm\,0.59$} & \multicolumn{2}{c}{$31.99\,\pm\,3.01$} & \multicolumn{2}{c}{$31.00\,\pm\,3.20$} & \multicolumn{2}{c}{$146.84\,\pm\,13.51$} & \multicolumn{2}{c}{$23.25\,\pm\,2.47$} & \multicolumn{2}{c}{$13.68\,\pm\,2.53$} & \multicolumn{2}{c}{$0.04\,\pm\,0.09$} & \multicolumn{2}{c}{$962.24\,\pm\,42.87$} \\
\texttt{Pi\_0.5} & \multicolumn{2}{|c|}{\cellcolor{ForestGreen!20}$\mathbf{0.40}\,\pm\,0.14$} & \multicolumn{2}{c}{\cellcolor{ForestGreen!20}$\mathbf{0.65}\,\pm\,0.10$} & \multicolumn{2}{c}{$0.34\,\pm\,0.03$} & \multicolumn{2}{c}{$0.05\,\pm\,0.01$} & \multicolumn{2}{c}{$2.22\,\pm\,0.44$} & \multicolumn{2}{c}{\cellcolor{ForestGreen!20}$\mathbf{3.66}\,\pm\,0.68$} & \multicolumn{2}{c}{\cellcolor{ForestGreen!20}$\mathbf{1.32}\,\pm\,0.47$} & \multicolumn{2}{c}{$11.22\,\pm\,2.31$} & \multicolumn{2}{c}{$18.14\,\pm\,1.78$} & \multicolumn{2}{c}{$88.53\,\pm\,7.98$} & \multicolumn{2}{c}{$8.00\,\pm\,1.68$} & \multicolumn{2}{c}{$2.30\,\pm\,0.99$} & \multicolumn{2}{c}{$0.26\,\pm\,0.13$} & \multicolumn{2}{c}{\cellcolor{ForestGreen!20}$\mathbf{620.58}\,\pm\,120.85$} \\
\texttt{MolmoAct} & \multicolumn{2}{|c|}{$0.36\,\pm\,0.14$} & \multicolumn{2}{c}{$0.58\,\pm\,0.10$} & \multicolumn{2}{c}{$0.29\,\pm\,0.03$} & \multicolumn{2}{c}{\cellcolor{ForestGreen!20}$\mathbf{0.04}\,\pm\,0.00$} & \multicolumn{2}{c}{$2.04\,\pm\,0.31$} & \multicolumn{2}{c}{$4.21\,\pm\,0.66$} & \multicolumn{2}{c}{$1.40\,\pm\,0.44$} & \multicolumn{2}{c}{$10.78\,\pm\,1.73$} & \multicolumn{2}{c}{$18.43\,\pm\,1.91$} & \multicolumn{2}{c}{$90.54\,\pm\,7.73$} & \multicolumn{2}{c}{$7.99\,\pm\,1.37$} & \multicolumn{2}{c}{\cellcolor{ForestGreen!20}$\mathbf{1.10}\,\pm\,0.43$} & \multicolumn{2}{c}{$0.26\,\pm\,0.17$} & \multicolumn{2}{c}{$695.86\,\pm\,111.27$} \\
\midrule
\texttt{Human} & \multicolumn{2}{|c|}{$1.00\,\pm\,0.00$} & \multicolumn{2}{c}{$1.00\,\pm\,0.00$} & \multicolumn{2}{c}{$0.33\,\pm\,0.03$} & \multicolumn{2}{c}{$0.03\,\pm\,0.00$} & \multicolumn{2}{c}{$0.53\,\pm\,0.03$} & \multicolumn{2}{c}{$0.51\,\pm\,0.07$} & \multicolumn{2}{c}{$0.10\,\pm\,0.16$} & \multicolumn{2}{c}{$2.26\,\pm\,0.18$} & \multicolumn{2}{c}{$28.94\,\pm\,3.07$} & \multicolumn{2}{c}{$120.52\,\pm\,11.54$} & \multicolumn{2}{c}{$1.38\,\pm\,0.18$} & \multicolumn{2}{c}{$0.17\,\pm\,0.17$} & \multicolumn{2}{c}{$0.03\,\pm\,0.13$} & \multicolumn{2}{c}{$103.53\,\pm\,14.28$} \\
\midrule
\multicolumn{29}{c}{\textbf{Lift Pot}} \\
\midrule
\texttt{Diffusion} & \multicolumn{2}{|c|}{$0.02\,\pm\,0.08$} & \multicolumn{2}{c}{$0.07\,\pm\,0.03$} & \multicolumn{2}{c}{$0.36\,\pm\,0.05$} & \multicolumn{2}{c}{$0.18\,\pm\,0.03$} & \multicolumn{2}{c}{$4.58\,\pm\,0.56$} & \multicolumn{2}{c}{$8.87\,\pm\,0.40$} & \multicolumn{2}{c}{$0.20\,\pm\,0.18$} & \multicolumn{2}{c}{$24.46\,\pm\,2.63$} & \multicolumn{2}{c}{$27.14\,\pm\,3.77$} & \multicolumn{2}{c}{$142.55\,\pm\,18.69$} & \multicolumn{2}{c}{$16.55\,\pm\,1.88$} & \multicolumn{2}{c}{$10.64\,\pm\,2.02$} & \multicolumn{2}{c}{\cellcolor{ForestGreen!20}$\mathbf{0.12}\,\pm\,0.12$} & \multicolumn{2}{c}{$978.06\,\pm\,32.64$} \\
\texttt{GR00T} & \multicolumn{2}{|c|}{$0.06\,\pm\,0.10$} & \multicolumn{2}{c}{$0.30\,\pm\,0.05$} & \multicolumn{2}{c}{$0.51\,\pm\,0.02$} & \multicolumn{2}{c}{$0.09\,\pm\,0.02$} & \multicolumn{2}{c}{$2.13\,\pm\,0.30$} & \multicolumn{2}{c}{$7.43\,\pm\,1.22$} & \multicolumn{2}{c}{$4.54\,\pm\,0.76$} & \multicolumn{2}{c}{$21.44\,\pm\,2.66$} & \multicolumn{2}{c}{$28.17\,\pm\,1.91$} & \multicolumn{2}{c}{$231.37\,\pm\,6.96$} & \multicolumn{2}{c}{$13.14\,\pm\,1.64$} & \multicolumn{2}{c}{$5.88\,\pm\,2.26$} & \multicolumn{2}{c}{$1.34\,\pm\,0.53$} & \multicolumn{2}{c}{$628.14\,\pm\,92.83$} \\
\texttt{xVLA} & \multicolumn{2}{|c|}{$0.00\,\pm\,0.07$} & \multicolumn{2}{c}{$0.21\,\pm\,0.05$} & \multicolumn{2}{c}{$0.67\,\pm\,0.06$} & \multicolumn{2}{c}{$0.22\,\pm\,0.04$} & \multicolumn{2}{c}{$5.31\,\pm\,0.84$} & \multicolumn{2}{c}{$8.29\,\pm\,3.48$} & \multicolumn{2}{c}{$2.26\,\pm\,0.69$} & \multicolumn{2}{c}{$30.12\,\pm\,3.74$} & \multicolumn{2}{c}{$36.35\,\pm\,3.80$} & \multicolumn{2}{c}{$173.17\,\pm\,16.93$} & \multicolumn{2}{c}{$21.79\,\pm\,2.89$} & \multicolumn{2}{c}{$4.96\,\pm\,1.66$} & \multicolumn{2}{c}{$0.32\,\pm\,0.20$} & \multicolumn{2}{c}{$747.02\,\pm\,90.35$} \\
\texttt{Pi\_0.5} & \multicolumn{2}{|c|}{\cellcolor{ForestGreen!20}$\mathbf{0.52}\,\pm\,0.13$} & \multicolumn{2}{c}{$0.41\,\pm\,0.08$} & \multicolumn{2}{c}{$0.34\,\pm\,0.03$} & \multicolumn{2}{c}{$0.07\,\pm\,0.02$} & \multicolumn{2}{c}{$2.73\,\pm\,0.62$} & \multicolumn{2}{c}{$3.86\,\pm\,0.78$} & \multicolumn{2}{c}{$0.06\,\pm\,0.10$} & \multicolumn{2}{c}{$14.23\,\pm\,3.13$} & \multicolumn{2}{c}{$27.71\,\pm\,2.54$} & \multicolumn{2}{c}{$139.02\,\pm\,12.27$} & \multicolumn{2}{c}{$9.53\,\pm\,2.06$} & \multicolumn{2}{c}{$0.80\,\pm\,0.67$} & \multicolumn{2}{c}{$0.28\,\pm\,0.15$} & \multicolumn{2}{c}{$550.78\,\pm\,117.31$} \\
\texttt{MolmoAct} & \multicolumn{2}{|c|}{$0.32\,\pm\,0.14$} & \multicolumn{2}{c}{$0.47\,\pm\,0.07$} & \multicolumn{2}{c}{$0.28\,\pm\,0.04$} & \multicolumn{2}{c}{\cellcolor{ForestGreen!20}$\mathbf{0.05}\,\pm\,0.01$} & \multicolumn{2}{c}{$2.21\,\pm\,0.42$} & \multicolumn{2}{c}{$4.67\,\pm\,0.83$} & \multicolumn{2}{c}{$0.44\,\pm\,0.28$} & \multicolumn{2}{c}{$11.36\,\pm\,1.87$} & \multicolumn{2}{c}{$26.56\,\pm\,4.39$} & \multicolumn{2}{c}{$129.13\,\pm\,20.69$} & \multicolumn{2}{c}{$7.86\,\pm\,1.25$} & \multicolumn{2}{c}{\cellcolor{ForestGreen!20}$\mathbf{0.48}\,\pm\,0.38$} & \multicolumn{2}{c}{$0.52\,\pm\,0.28$} & \multicolumn{2}{c}{$646.72\,\pm\,116.82$} \\
\midrule
\texttt{Human} & \multicolumn{2}{|c|}{$1.00\,\pm\,0.00$} & \multicolumn{2}{c}{$1.00\,\pm\,0.00$} & \multicolumn{2}{c}{$0.49\,\pm\,0.05$} & \multicolumn{2}{c}{$0.03\,\pm\,0.01$} & \multicolumn{2}{c}{$0.61\,\pm\,0.07$} & \multicolumn{2}{c}{$0.45\,\pm\,0.06$} & \multicolumn{2}{c}{$0.00\,\pm\,0.14$} & \multicolumn{2}{c}{$3.33\,\pm\,0.34$} & \multicolumn{2}{c}{$62.59\,\pm\,4.70$} & \multicolumn{2}{c}{$287.53\,\pm\,20.08$} & \multicolumn{2}{c}{$2.40\,\pm\,0.21$} & \multicolumn{2}{c}{$0.00\,\pm\,0.14$} & \multicolumn{2}{c}{$0.00\,\pm\,0.14$} & \multicolumn{2}{c}{$71.04\,\pm\,9.68$} \\
\midrule
\multicolumn{29}{c}{\textbf{Lift Tray}} \\
\midrule
\texttt{Diffusion} & \multicolumn{2}{|c|}{$0.00\,\pm\,0.07$} & \multicolumn{2}{c}{$0.47\,\pm\,0.06$} & \multicolumn{2}{c}{$0.47\,\pm\,0.04$} & \multicolumn{2}{c}{$0.21\,\pm\,0.04$} & \multicolumn{2}{c}{$5.01\,\pm\,0.52$} & \multicolumn{2}{c}{$9.85\,\pm\,0.28$} & \multicolumn{2}{c}{$5.28\,\pm\,1.29$} & \multicolumn{2}{c}{$29.97\,\pm\,2.22$} & \multicolumn{2}{c}{$27.50\,\pm\,3.48$} & \multicolumn{2}{c}{$147.21\,\pm\,13.54$} & \multicolumn{2}{c}{$21.38\,\pm\,1.78$} & \multicolumn{2}{c}{$6.70\,\pm\,2.20$} & \multicolumn{2}{c}{$0.48\,\pm\,0.20$} & \multicolumn{2}{c}{$1000.00\,\pm\,0.00$} \\
\texttt{GR00T} & \multicolumn{2}{|c|}{$0.00\,\pm\,0.07$} & \multicolumn{2}{c}{$0.46\,\pm\,0.05$} & \multicolumn{2}{c}{$0.51\,\pm\,0.02$} & \multicolumn{2}{c}{$0.12\,\pm\,0.02$} & \multicolumn{2}{c}{$3.93\,\pm\,0.30$} & \multicolumn{2}{c}{$13.28\,\pm\,0.25$} & \multicolumn{2}{c}{$6.62\,\pm\,1.25$} & \multicolumn{2}{c}{$32.10\,\pm\,1.15$} & \multicolumn{2}{c}{$34.08\,\pm\,1.98$} & \multicolumn{2}{c}{$228.16\,\pm\,6.87$} & \multicolumn{2}{c}{$20.66\,\pm\,0.93$} & \multicolumn{2}{c}{$17.06\,\pm\,2.25$} & \multicolumn{2}{c}{$0.50\,\pm\,0.23$} & \multicolumn{2}{c}{$1000.00\,\pm\,0.00$} \\
\texttt{xVLA} & \multicolumn{2}{|c|}{$0.00\,\pm\,0.07$} & \multicolumn{2}{c}{$0.35\,\pm\,0.04$} & \multicolumn{2}{c}{$0.63\,\pm\,0.06$} & \multicolumn{2}{c}{$0.24\,\pm\,0.04$} & \multicolumn{2}{c}{$7.01\,\pm\,0.92$} & \multicolumn{2}{c}{$10.49\,\pm\,0.43$} & \multicolumn{2}{c}{$4.88\,\pm\,0.94$} & \multicolumn{2}{c}{$37.46\,\pm\,3.72$} & \multicolumn{2}{c}{$32.82\,\pm\,4.39$} & \multicolumn{2}{c}{$151.68\,\pm\,16.81$} & \multicolumn{2}{c}{$28.46\,\pm\,3.09$} & \multicolumn{2}{c}{$10.46\,\pm\,2.68$} & \multicolumn{2}{c}{\cellcolor{ForestGreen!20}$\mathbf{0.16}\,\pm\,0.12$} & \multicolumn{2}{c}{$1000.00\,\pm\,0.00$} \\
\texttt{Pi\_0.5} & \multicolumn{2}{|c|}{$0.32\,\pm\,0.14$} & \multicolumn{2}{c}{$0.75\,\pm\,0.07$} & \multicolumn{2}{c}{$0.52\,\pm\,0.05$} & \multicolumn{2}{c}{$0.14\,\pm\,0.03$} & \multicolumn{2}{c}{$4.83\,\pm\,0.84$} & \multicolumn{2}{c}{$5.87\,\pm\,0.94$} & \multicolumn{2}{c}{$2.62\,\pm\,0.99$} & \multicolumn{2}{c}{$29.67\,\pm\,5.06$} & \multicolumn{2}{c}{$31.27\,\pm\,2.31$} & \multicolumn{2}{c}{$181.95\,\pm\,11.42$} & \multicolumn{2}{c}{$21.82\,\pm\,3.78$} & \multicolumn{2}{c}{$5.22\,\pm\,2.07$} & \multicolumn{2}{c}{$0.92\,\pm\,0.22$} & \multicolumn{2}{c}{$719.16\,\pm\,115.13$} \\
\texttt{MolmoAct} & \multicolumn{2}{|c|}{\cellcolor{ForestGreen!20}$\mathbf{0.46}\,\pm\,0.14$} & \multicolumn{2}{c}{\cellcolor{ForestGreen!20}$\mathbf{0.91}\,\pm\,0.05$} & \multicolumn{2}{c}{$0.38\,\pm\,0.03$} & \multicolumn{2}{c}{$0.07\,\pm\,0.02$} & \multicolumn{2}{c}{$2.18\,\pm\,0.46$} & \multicolumn{2}{c}{\cellcolor{ForestGreen!20}$\mathbf{5.04}\,\pm\,1.06$} & \multicolumn{2}{c}{\cellcolor{ForestGreen!20}$\mathbf{0.98}\,\pm\,0.57$} & \multicolumn{2}{c}{$14.06\,\pm\,2.95$} & \multicolumn{2}{c}{$21.12\,\pm\,2.41$} & \multicolumn{2}{c}{$127.85\,\pm\,13.57$} & \multicolumn{2}{c}{$9.58\,\pm\,2.04$} & \multicolumn{2}{c}{\cellcolor{ForestGreen!20}$\mathbf{1.98}\,\pm\,1.17$} & \multicolumn{2}{c}{$0.90\,\pm\,0.35$} & \multicolumn{2}{c}{\cellcolor{ForestGreen!20}$\mathbf{589.02}\,\pm\,125.85$} \\
\midrule
\texttt{Human} & \multicolumn{2}{|c|}{$1.00\,\pm\,0.00$} & \multicolumn{2}{c}{$1.00\,\pm\,0.00$} & \multicolumn{2}{c}{$0.39\,\pm\,0.04$} & \multicolumn{2}{c}{$0.03\,\pm\,0.00$} & \multicolumn{2}{c}{$0.52\,\pm\,0.03$} & \multicolumn{2}{c}{$0.56\,\pm\,0.03$} & \multicolumn{2}{c}{$0.00\,\pm\,0.06$} & \multicolumn{2}{c}{$2.87\,\pm\,0.13$} & \multicolumn{2}{c}{$26.77\,\pm\,2.93$} & \multicolumn{2}{c}{$153.00\,\pm\,16.82$} & \multicolumn{2}{c}{$2.04\,\pm\,0.08$} & \multicolumn{2}{c}{$0.00\,\pm\,0.06$} & \multicolumn{2}{c}{$0.00\,\pm\,0.06$} & \multicolumn{2}{c}{$83.62\,\pm\,3.96$} \\
\midrule
\multicolumn{29}{c}{\textbf{Pack Box}} \\
\midrule
\texttt{Diffusion} & \multicolumn{2}{|c|}{$0.04\,\pm\,0.09$} & \multicolumn{2}{c}{$0.27\,\pm\,0.06$} & \multicolumn{2}{c}{$0.43\,\pm\,0.04$} & \multicolumn{2}{c}{$0.11\,\pm\,0.02$} & \multicolumn{2}{c}{$7.19\,\pm\,0.41$} & \multicolumn{2}{c}{$5.42\,\pm\,0.31$} & \multicolumn{2}{c}{$0.58\,\pm\,0.34$} & \multicolumn{2}{c}{$36.50\,\pm\,2.51$} & \multicolumn{2}{c}{$29.98\,\pm\,2.07$} & \multicolumn{2}{c}{$154.68\,\pm\,11.16$} & \multicolumn{2}{c}{$25.85\,\pm\,1.81$} & \multicolumn{2}{c}{$9.82\,\pm\,3.05$} & \multicolumn{2}{c}{$0.44\,\pm\,0.16$} & \multicolumn{2}{c}{$962.98\,\pm\,44.27$} \\
\texttt{GR00T} & \multicolumn{2}{|c|}{$0.00\,\pm\,0.07$} & \multicolumn{2}{c}{$0.07\,\pm\,0.03$} & \multicolumn{2}{c}{$0.50\,\pm\,0.02$} & \multicolumn{2}{c}{$0.24\,\pm\,0.04$} & \multicolumn{2}{c}{$4.77\,\pm\,0.36$} & \multicolumn{2}{c}{$10.75\,\pm\,0.20$} & \multicolumn{2}{c}{$1.04\,\pm\,0.47$} & \multicolumn{2}{c}{$29.29\,\pm\,1.37$} & \multicolumn{2}{c}{$40.49\,\pm\,3.08$} & \multicolumn{2}{c}{$207.72\,\pm\,9.51$} & \multicolumn{2}{c}{$19.50\,\pm\,1.10$} & \multicolumn{2}{c}{$13.30\,\pm\,2.92$} & \multicolumn{2}{c}{$0.30\,\pm\,0.19$} & \multicolumn{2}{c}{$1000.00\,\pm\,0.00$} \\
\texttt{xVLA} & \multicolumn{2}{|c|}{$0.02\,\pm\,0.08$} & \multicolumn{2}{c}{$0.12\,\pm\,0.05$} & \multicolumn{2}{c}{$0.79\,\pm\,0.05$} & \multicolumn{2}{c}{$0.25\,\pm\,0.03$} & \multicolumn{2}{c}{$9.76\,\pm\,0.88$} & \multicolumn{2}{c}{$7.29\,\pm\,0.35$} & \multicolumn{2}{c}{$1.30\,\pm\,0.53$} & \multicolumn{2}{c}{$47.57\,\pm\,3.10$} & \multicolumn{2}{c}{$46.60\,\pm\,3.63$} & \multicolumn{2}{c}{$200.34\,\pm\,12.90$} & \multicolumn{2}{c}{$34.53\,\pm\,2.36$} & \multicolumn{2}{c}{$9.24\,\pm\,3.08$} & \multicolumn{2}{c}{\cellcolor{ForestGreen!20}$\mathbf{0.14}\,\pm\,0.11$} & \multicolumn{2}{c}{$988.12\,\pm\,23.28$} \\
\texttt{Pi\_0.5} & \multicolumn{2}{|c|}{$0.54\,\pm\,0.14$} & \multicolumn{2}{c}{$0.55\,\pm\,0.06$} & \multicolumn{2}{c}{$0.73\,\pm\,0.07$} & \multicolumn{2}{c}{$0.10\,\pm\,0.01$} & \multicolumn{2}{c}{$4.60\,\pm\,0.89$} & \multicolumn{2}{c}{$3.70\,\pm\,0.72$} & \multicolumn{2}{c}{$0.90\,\pm\,0.41$} & \multicolumn{2}{c}{$22.61\,\pm\,4.56$} & \multicolumn{2}{c}{$27.59\,\pm\,1.39$} & \multicolumn{2}{c}{$130.35\,\pm\,7.35$} & \multicolumn{2}{c}{$15.84\,\pm\,3.25$} & \multicolumn{2}{c}{$1.24\,\pm\,1.19$} & \multicolumn{2}{c}{$0.44\,\pm\,0.20$} & \multicolumn{2}{c}{$566.54\,\pm\,113.98$} \\
\texttt{MolmoAct} & \multicolumn{2}{|c|}{\cellcolor{ForestGreen!20}$\mathbf{0.62}\,\pm\,0.14$} & \multicolumn{2}{c}{\cellcolor{ForestGreen!20}$\mathbf{0.57}\,\pm\,0.05$} & \multicolumn{2}{c}{$0.58\,\pm\,0.06$} & \multicolumn{2}{c}{$0.08\,\pm\,0.02$} & \multicolumn{2}{c}{$3.22\,\pm\,0.71$} & \multicolumn{2}{c}{\cellcolor{ForestGreen!20}$\mathbf{3.14}\,\pm\,0.78$} & \multicolumn{2}{c}{\cellcolor{ForestGreen!20}$\mathbf{0.44}\,\pm\,0.31$} & \multicolumn{2}{c}{$15.26\,\pm\,3.72$} & \multicolumn{2}{c}{$24.28\,\pm\,1.58$} & \multicolumn{2}{c}{$112.64\,\pm\,7.83$} & \multicolumn{2}{c}{$10.10\,\pm\,2.59$} & \multicolumn{2}{c}{$1.12\,\pm\,0.72$} & \multicolumn{2}{c}{$0.48\,\pm\,0.25$} & \multicolumn{2}{c}{\cellcolor{ForestGreen!20}$\mathbf{470.42}\,\pm\,115.06$} \\
\midrule
\texttt{Human} & \multicolumn{2}{|c|}{$1.00\,\pm\,0.00$} & \multicolumn{2}{c}{$1.00\,\pm\,0.00$} & \multicolumn{2}{c}{$0.51\,\pm\,0.04$} & \multicolumn{2}{c}{$0.05\,\pm\,0.00$} & \multicolumn{2}{c}{$1.30\,\pm\,0.06$} & \multicolumn{2}{c}{$0.68\,\pm\,0.04$} & \multicolumn{2}{c}{$0.84\,\pm\,0.37$} & \multicolumn{2}{c}{$5.38\,\pm\,0.34$} & \multicolumn{2}{c}{$28.24\,\pm\,1.50$} & \multicolumn{2}{c}{$134.77\,\pm\,6.73$} & \multicolumn{2}{c}{$3.35\,\pm\,0.21$} & \multicolumn{2}{c}{$0.18\,\pm\,0.16$} & \multicolumn{2}{c}{$0.03\,\pm\,0.07$} & \multicolumn{2}{c}{$129.66\,\pm\,8.00$} \\
\midrule
\multicolumn{29}{c}{\textbf{Pick Single Book From Table}} \\
\midrule
\texttt{Diffusion} & \multicolumn{2}{|c|}{$0.02\,\pm\,0.08$} & \multicolumn{2}{c}{$0.15\,\pm\,0.09$} & \multicolumn{2}{c}{$0.59\,\pm\,0.04$} & \multicolumn{2}{c}{$0.24\,\pm\,0.03$} & \multicolumn{2}{c}{$4.42\,\pm\,0.49$} & \multicolumn{2}{c}{$11.20\,\pm\,0.83$} & \multicolumn{2}{c}{$5.52\,\pm\,1.27$} & \multicolumn{2}{c}{$28.17\,\pm\,2.59$} & \multicolumn{2}{c}{$21.24\,\pm\,1.98$} & \multicolumn{2}{c}{$136.43\,\pm\,10.11$} & \multicolumn{2}{c}{$17.38\,\pm\,1.69$} & \multicolumn{2}{c}{$4.26\,\pm\,1.50$} & \multicolumn{2}{c}{$0.04\,\pm\,0.09$} & \multicolumn{2}{c}{$926.04\,\pm\,61.77$} \\
\texttt{GR00T} & \multicolumn{2}{|c|}{$0.00\,\pm\,0.07$} & \multicolumn{2}{c}{$0.05\,\pm\,0.05$} & \multicolumn{2}{c}{$0.56\,\pm\,0.03$} & \multicolumn{2}{c}{$0.26\,\pm\,0.03$} & \multicolumn{2}{c}{$2.96\,\pm\,0.36$} & \multicolumn{2}{c}{$15.47\,\pm\,1.63$} & \multicolumn{2}{c}{$9.12\,\pm\,1.30$} & \multicolumn{2}{c}{$28.12\,\pm\,2.69$} & \multicolumn{2}{c}{$28.97\,\pm\,1.90$} & \multicolumn{2}{c}{$237.89\,\pm\,7.47$} & \multicolumn{2}{c}{$18.37\,\pm\,1.90$} & \multicolumn{2}{c}{$10.90\,\pm\,2.44$} & \multicolumn{2}{c}{$0.04\,\pm\,0.09$} & \multicolumn{2}{c}{$821.42\,\pm\,83.40$} \\
\texttt{xVLA} & \multicolumn{2}{|c|}{$0.00\,\pm\,0.07$} & \multicolumn{2}{c}{$0.02\,\pm\,0.08$} & \multicolumn{2}{c}{$0.72\,\pm\,0.06$} & \multicolumn{2}{c}{$0.21\,\pm\,0.03$} & \multicolumn{2}{c}{$5.10\,\pm\,0.72$} & \multicolumn{2}{c}{$13.53\,\pm\,0.81$} & \multicolumn{2}{c}{$5.58\,\pm\,0.96$} & \multicolumn{2}{c}{$31.56\,\pm\,2.94$} & \multicolumn{2}{c}{$22.35\,\pm\,2.75$} & \multicolumn{2}{c}{$114.50\,\pm\,11.31$} & \multicolumn{2}{c}{$22.67\,\pm\,2.35$} & \multicolumn{2}{c}{$7.90\,\pm\,1.75$} & \multicolumn{2}{c}{\cellcolor{ForestGreen!20}$\mathbf{0.02}\,\pm\,0.08$} & \multicolumn{2}{c}{$949.08\,\pm\,47.67$} \\
\texttt{Pi\_0.5} & \multicolumn{2}{|c|}{$0.22\,\pm\,0.13$} & \multicolumn{2}{c}{$0.44\,\pm\,0.13$} & \multicolumn{2}{c}{$0.72\,\pm\,0.07$} & \multicolumn{2}{c}{$0.20\,\pm\,0.02$} & \multicolumn{2}{c}{$1.89\,\pm\,0.47$} & \multicolumn{2}{c}{$6.03\,\pm\,1.54$} & \multicolumn{2}{c}{$3.02\,\pm\,0.88$} & \multicolumn{2}{c}{$11.37\,\pm\,2.88$} & \multicolumn{2}{c}{$17.94\,\pm\,1.51$} & \multicolumn{2}{c}{$91.87\,\pm\,6.23$} & \multicolumn{2}{c}{$6.67\,\pm\,1.76$} & \multicolumn{2}{c}{$0.28\,\pm\,0.24$} & \multicolumn{2}{c}{$0.12\,\pm\,0.11$} & \multicolumn{2}{c}{\cellcolor{ForestGreen!20}$\mathbf{458.76}\,\pm\,120.93$} \\
\texttt{MolmoAct} & \multicolumn{2}{|c|}{\cellcolor{ForestGreen!20}$\mathbf{0.30}\,\pm\,0.14$} & \multicolumn{2}{c}{\cellcolor{ForestGreen!20}$\mathbf{0.48}\,\pm\,0.13$} & \multicolumn{2}{c}{$0.62\,\pm\,0.08$} & \multicolumn{2}{c}{\cellcolor{ForestGreen!20}$\mathbf{0.18}\,\pm\,0.01$} & \multicolumn{2}{c}{$1.96\,\pm\,0.35$} & \multicolumn{2}{c}{$8.79\,\pm\,1.68$} & \multicolumn{2}{c}{$2.70\,\pm\,0.77$} & \multicolumn{2}{c}{$13.06\,\pm\,2.38$} & \multicolumn{2}{c}{$15.55\,\pm\,2.22$} & \multicolumn{2}{c}{$82.43\,\pm\,8.31$} & \multicolumn{2}{c}{$7.10\,\pm\,1.30$} & \multicolumn{2}{c}{$0.78\,\pm\,0.59$} & \multicolumn{2}{c}{\cellcolor{ForestGreen!20}$\mathbf{0.02}\,\pm\,0.08$} & \multicolumn{2}{c}{$641.82\,\pm\,123.19$} \\
\midrule
\texttt{Human} & \multicolumn{2}{|c|}{$1.00\,\pm\,0.00$} & \multicolumn{2}{c}{$1.00\,\pm\,0.00$} & \multicolumn{2}{c}{$0.87\,\pm\,0.00$} & \multicolumn{2}{c}{$0.18\,\pm\,0.00$} & \multicolumn{2}{c}{$0.76\,\pm\,0.00$} & \multicolumn{2}{c}{$1.20\,\pm\,0.00$} & \multicolumn{2}{c}{$0.00\,\pm\,0.00$} & \multicolumn{2}{c}{$4.09\,\pm\,0.00$} & \multicolumn{2}{c}{$30.32\,\pm\,0.00$} & \multicolumn{2}{c}{$138.21\,\pm\,0.00$} & \multicolumn{2}{c}{$1.90\,\pm\,0.00$} & \multicolumn{2}{c}{$0.00\,\pm\,0.00$} & \multicolumn{2}{c}{$0.00\,\pm\,0.00$} & \multicolumn{2}{c}{$118.00\,\pm\,0.00$} \\
\midrule
\multicolumn{29}{c}{\textbf{Rotate Valve}} \\
\midrule
\texttt{Diffusion} & \multicolumn{2}{|c|}{$0.04\,\pm\,0.09$} & \multicolumn{2}{c}{$0.23\,\pm\,0.07$} & \multicolumn{2}{c}{$0.44\,\pm\,0.04$} & \multicolumn{2}{c}{$0.10\,\pm\,0.02$} & \multicolumn{2}{c}{$2.10\,\pm\,0.23$} & \multicolumn{2}{c}{$11.62\,\pm\,0.52$} & \multicolumn{2}{c}{$10.18\,\pm\,1.47$} & \multicolumn{2}{c}{$13.61\,\pm\,1.18$} & \multicolumn{2}{c}{$5.88\,\pm\,0.54$} & \multicolumn{2}{c}{$41.95\,\pm\,3.08$} & \multicolumn{2}{c}{$9.76\,\pm\,0.94$} & \multicolumn{2}{c}{$2.86\,\pm\,1.06$} & \multicolumn{2}{c}{\cellcolor{ForestGreen!20}$\mathbf{0.00}\,\pm\,0.07$} & \multicolumn{2}{c}{$977.16\,\pm\,35.82$} \\
\texttt{GR00T} & \multicolumn{2}{|c|}{$0.00\,\pm\,0.07$} & \multicolumn{2}{c}{$0.11\,\pm\,0.04$} & \multicolumn{2}{c}{$0.63\,\pm\,0.03$} & \multicolumn{2}{c}{$0.08\,\pm\,0.01$} & \multicolumn{2}{c}{$2.69\,\pm\,0.17$} & \multicolumn{2}{c}{$18.11\,\pm\,0.63$} & \multicolumn{2}{c}{$23.70\,\pm\,2.64$} & \multicolumn{2}{c}{$21.46\,\pm\,0.77$} & \multicolumn{2}{c}{$14.40\,\pm\,0.81$} & \multicolumn{2}{c}{$88.73\,\pm\,2.50$} & \multicolumn{2}{c}{$15.09\,\pm\,0.75$} & \multicolumn{2}{c}{$16.40\,\pm\,2.50$} & \multicolumn{2}{c}{\cellcolor{ForestGreen!20}$\mathbf{0.00}\,\pm\,0.07$} & \multicolumn{2}{c}{$1000.00\,\pm\,0.00$} \\
\texttt{xVLA} & \multicolumn{2}{|c|}{$0.10\,\pm\,0.11$} & \multicolumn{2}{c}{$0.15\,\pm\,0.06$} & \multicolumn{2}{c}{$0.73\,\pm\,0.05$} & \multicolumn{2}{c}{$0.13\,\pm\,0.02$} & \multicolumn{2}{c}{$3.07\,\pm\,0.29$} & \multicolumn{2}{c}{$11.74\,\pm\,0.62$} & \multicolumn{2}{c}{$12.34\,\pm\,2.19$} & \multicolumn{2}{c}{$19.40\,\pm\,1.35$} & \multicolumn{2}{c}{$7.30\,\pm\,0.55$} & \multicolumn{2}{c}{$38.85\,\pm\,2.53$} & \multicolumn{2}{c}{$14.54\,\pm\,1.06$} & \multicolumn{2}{c}{$2.40\,\pm\,0.91$} & \multicolumn{2}{c}{\cellcolor{ForestGreen!20}$\mathbf{0.00}\,\pm\,0.07$} & \multicolumn{2}{c}{$953.00\,\pm\,44.60$} \\
\texttt{Pi\_0.5} & \multicolumn{2}{|c|}{\cellcolor{ForestGreen!20}$\mathbf{0.72}\,\pm\,0.14$} & \multicolumn{2}{c}{$0.88\,\pm\,0.05$} & \multicolumn{2}{c}{$0.36\,\pm\,0.02$} & \multicolumn{2}{c}{$0.03\,\pm\,0.00$} & \multicolumn{2}{c}{$0.62\,\pm\,0.14$} & \multicolumn{2}{c}{$8.65\,\pm\,1.59$} & \multicolumn{2}{c}{$12.42\,\pm\,2.36$} & \multicolumn{2}{c}{$4.67\,\pm\,0.95$} & \multicolumn{2}{c}{$2.99\,\pm\,0.22$} & \multicolumn{2}{c}{$19.81\,\pm\,1.29$} & \multicolumn{2}{c}{$3.43\,\pm\,0.69$} & \multicolumn{2}{c}{\cellcolor{ForestGreen!20}$\mathbf{0.02}\,\pm\,0.08$} & \multicolumn{2}{c}{\cellcolor{ForestGreen!20}$\mathbf{0.00}\,\pm\,0.07$} & \multicolumn{2}{c}{$527.64\,\pm\,95.25$} \\
\texttt{MolmoAct} & \multicolumn{2}{|c|}{\cellcolor{ForestGreen!20}$\mathbf{0.72}\,\pm\,0.14$} & \multicolumn{2}{c}{\cellcolor{ForestGreen!20}$\mathbf{0.90}\,\pm\,0.05$} & \multicolumn{2}{c}{\cellcolor{ForestGreen!20}$\mathbf{0.30}\,\pm\,0.02$} & \multicolumn{2}{c}{\cellcolor{ForestGreen!20}$\mathbf{0.03}\,\pm\,0.00$} & \multicolumn{2}{c}{\cellcolor{ForestGreen!20}$\mathbf{0.45}\,\pm\,0.10$} & \multicolumn{2}{c}{\cellcolor{ForestGreen!20}$\mathbf{8.51}\,\pm\,1.96$} & \multicolumn{2}{c}{\cellcolor{ForestGreen!20}$\mathbf{8.84}\,\pm\,1.28$} & \multicolumn{2}{c}{\cellcolor{ForestGreen!20}$\mathbf{3.42}\,\pm\,0.73$} & \multicolumn{2}{c}{\cellcolor{ForestGreen!20}$\mathbf{2.56}\,\pm\,0.20$} & \multicolumn{2}{c}{\cellcolor{ForestGreen!20}$\mathbf{16.77}\,\pm\,1.05$} & \multicolumn{2}{c}{\cellcolor{ForestGreen!20}$\mathbf{2.48}\,\pm\,0.51$} & \multicolumn{2}{c}{$0.14\,\pm\,0.24$} & \multicolumn{2}{c}{\cellcolor{ForestGreen!20}$\mathbf{0.00}\,\pm\,0.07$} & \multicolumn{2}{c}{\cellcolor{ForestGreen!20}$\mathbf{481.94}\,\pm\,107.26$} \\
\midrule
\texttt{Human} & \multicolumn{2}{|c|}{$1.00\,\pm\,0.00$} & \multicolumn{2}{c}{$1.00\,\pm\,0.00$} & \multicolumn{2}{c}{$0.37\,\pm\,0.04$} & \multicolumn{2}{c}{$0.01\,\pm\,0.00$} & \multicolumn{2}{c}{$0.20\,\pm\,0.02$} & \multicolumn{2}{c}{$1.98\,\pm\,0.22$} & \multicolumn{2}{c}{$6.38\,\pm\,0.73$} & \multicolumn{2}{c}{$1.24\,\pm\,0.11$} & \multicolumn{2}{c}{$3.60\,\pm\,0.31$} & \multicolumn{2}{c}{$22.26\,\pm\,2.06$} & \multicolumn{2}{c}{$1.04\,\pm\,0.10$} & \multicolumn{2}{c}{$0.00\,\pm\,0.19$} & \multicolumn{2}{c}{$0.00\,\pm\,0.19$} & \multicolumn{2}{c}{$150.12\,\pm\,16.61$} \\
\midrule
\multicolumn{29}{c}{\textbf{Stack Single Book Shelf}} \\
\midrule
\texttt{Diffusion} & \multicolumn{2}{|c|}{$0.00\,\pm\,0.07$} & \multicolumn{2}{c}{$0.01\,\pm\,0.01$} & \multicolumn{2}{c}{$0.78\,\pm\,0.06$} & \multicolumn{2}{c}{$0.24\,\pm\,0.02$} & \multicolumn{2}{c}{$6.01\,\pm\,0.63$} & \multicolumn{2}{c}{$11.95\,\pm\,0.62$} & \multicolumn{2}{c}{$5.74\,\pm\,0.83$} & \multicolumn{2}{c}{$35.11\,\pm\,2.58$} & \multicolumn{2}{c}{$29.84\,\pm\,3.97$} & \multicolumn{2}{c}{$166.93\,\pm\,15.36$} & \multicolumn{2}{c}{$22.73\,\pm\,1.84$} & \multicolumn{2}{c}{$3.54\,\pm\,1.22$} & \multicolumn{2}{c}{\cellcolor{ForestGreen!20}$\mathbf{0.00}\,\pm\,0.07$} & \multicolumn{2}{c}{$966.84\,\pm\,45.49$} \\
\texttt{GR00T} & \multicolumn{2}{|c|}{$0.00\,\pm\,0.07$} & \multicolumn{2}{c}{$0.04\,\pm\,0.03$} & \multicolumn{2}{c}{$0.56\,\pm\,0.02$} & \multicolumn{2}{c}{$0.29\,\pm\,0.03$} & \multicolumn{2}{c}{$3.17\,\pm\,0.41$} & \multicolumn{2}{c}{$16.17\,\pm\,1.54$} & \multicolumn{2}{c}{$9.46\,\pm\,1.34$} & \multicolumn{2}{c}{$28.67\,\pm\,2.49$} & \multicolumn{2}{c}{$31.38\,\pm\,2.57$} & \multicolumn{2}{c}{$236.75\,\pm\,5.66$} & \multicolumn{2}{c}{$18.50\,\pm\,1.71$} & \multicolumn{2}{c}{$11.38\,\pm\,2.70$} & \multicolumn{2}{c}{$0.04\,\pm\,0.09$} & \multicolumn{2}{c}{$842.26\,\pm\,76.69$} \\
\texttt{xVLA} & \multicolumn{2}{|c|}{$0.00\,\pm\,0.07$} & \multicolumn{2}{c}{$0.01\,\pm\,0.01$} & \multicolumn{2}{c}{$0.78\,\pm\,0.06$} & \multicolumn{2}{c}{$0.24\,\pm\,0.04$} & \multicolumn{2}{c}{$5.53\,\pm\,0.77$} & \multicolumn{2}{c}{$13.64\,\pm\,0.82$} & \multicolumn{2}{c}{$7.28\,\pm\,1.31$} & \multicolumn{2}{c}{$33.77\,\pm\,3.27$} & \multicolumn{2}{c}{$29.08\,\pm\,3.57$} & \multicolumn{2}{c}{$150.93\,\pm\,15.57$} & \multicolumn{2}{c}{$24.37\,\pm\,2.47$} & \multicolumn{2}{c}{$10.16\,\pm\,2.45$} & \multicolumn{2}{c}{$0.02\,\pm\,0.08$} & \multicolumn{2}{c}{$937.06\,\pm\,49.54$} \\
\texttt{Pi\_0.5} & \multicolumn{2}{|c|}{$0.12\,\pm\,0.12$} & \multicolumn{2}{c}{$0.27\,\pm\,0.10$} & \multicolumn{2}{c}{$0.79\,\pm\,0.06$} & \multicolumn{2}{c}{$0.20\,\pm\,0.01$} & \multicolumn{2}{c}{$2.04\,\pm\,0.56$} & \multicolumn{2}{c}{\cellcolor{ForestGreen!20}$\mathbf{5.96}\,\pm\,1.60$} & \multicolumn{2}{c}{\cellcolor{ForestGreen!20}$\mathbf{2.16}\,\pm\,0.58$} & \multicolumn{2}{c}{$12.82\,\pm\,3.59$} & \multicolumn{2}{c}{$19.90\,\pm\,1.25$} & \multicolumn{2}{c}{$104.69\,\pm\,5.90$} & \multicolumn{2}{c}{$7.39\,\pm\,2.09$} & \multicolumn{2}{c}{$2.16\,\pm\,1.45$} & \multicolumn{2}{c}{$0.16\,\pm\,0.17$} & \multicolumn{2}{c}{\cellcolor{ForestGreen!20}$\mathbf{414.04}\,\pm\,114.28$} \\
\texttt{MolmoAct} & \multicolumn{2}{|c|}{\cellcolor{ForestGreen!20}$\mathbf{0.18}\,\pm\,0.13$} & \multicolumn{2}{c}{\cellcolor{ForestGreen!20}$\mathbf{0.33}\,\pm\,0.11$} & \multicolumn{2}{c}{\cellcolor{ForestGreen!20}$\mathbf{0.67}\,\pm\,0.07$} & \multicolumn{2}{c}{\cellcolor{ForestGreen!20}$\mathbf{0.19}\,\pm\,0.02$} & \multicolumn{2}{c}{\cellcolor{ForestGreen!20}$\mathbf{2.27}\,\pm\,0.40$} & \multicolumn{2}{c}{$8.76\,\pm\,1.69$} & \multicolumn{2}{c}{$2.80\,\pm\,0.86$} & \multicolumn{2}{c}{\cellcolor{ForestGreen!20}$\mathbf{14.01}\,\pm\,2.59$} & \multicolumn{2}{c}{\cellcolor{ForestGreen!20}$\mathbf{16.60}\,\pm\,1.33$} & \multicolumn{2}{c}{\cellcolor{ForestGreen!20}$\mathbf{93.43}\,\pm\,6.99$} & \multicolumn{2}{c}{\cellcolor{ForestGreen!20}$\mathbf{7.45}\,\pm\,1.41$} & \multicolumn{2}{c}{\cellcolor{ForestGreen!20}$\mathbf{0.30}\,\pm\,0.31$} & \multicolumn{2}{c}{$0.10\,\pm\,0.11$} & \multicolumn{2}{c}{$600.90\,\pm\,117.81$} \\
\midrule
\texttt{Human} & \multicolumn{2}{|c|}{$1.00\,\pm\,0.00$} & \multicolumn{2}{c}{$1.00\,\pm\,0.00$} & \multicolumn{2}{c}{$0.85\,\pm\,0.00$} & \multicolumn{2}{c}{$0.20\,\pm\,0.00$} & \multicolumn{2}{c}{$1.10\,\pm\,0.00$} & \multicolumn{2}{c}{$2.12\,\pm\,0.00$} & \multicolumn{2}{c}{$0.00\,\pm\,0.00$} & \multicolumn{2}{c}{$6.29\,\pm\,0.00$} & \multicolumn{2}{c}{$19.79\,\pm\,0.00$} & \multicolumn{2}{c}{$100.20\,\pm\,0.00$} & \multicolumn{2}{c}{$2.90\,\pm\,0.00$} & \multicolumn{2}{c}{$0.00\,\pm\,0.00$} & \multicolumn{2}{c}{$0.00\,\pm\,0.00$} & \multicolumn{2}{c}{$194.00\,\pm\,0.00$} \\
\midrule
\multicolumn{29}{c}{\textbf{Stack Two Blocks}} \\
\midrule
\texttt{Diffusion} & \multicolumn{2}{|c|}{$0.06\,\pm\,0.10$} & \multicolumn{2}{c}{$0.13\,\pm\,0.05$} & \multicolumn{2}{c}{$0.34\,\pm\,0.03$} & \multicolumn{2}{c}{$0.08\,\pm\,0.01$} & \multicolumn{2}{c}{$4.57\,\pm\,0.23$} & \multicolumn{2}{c}{$7.27\,\pm\,0.24$} & \multicolumn{2}{c}{$4.58\,\pm\,0.47$} & \multicolumn{2}{c}{$24.32\,\pm\,1.55$} & \multicolumn{2}{c}{$18.74\,\pm\,0.83$} & \multicolumn{2}{c}{$106.54\,\pm\,4.95$} & \multicolumn{2}{c}{$18.54\,\pm\,1.23$} & \multicolumn{2}{c}{$2.04\,\pm\,0.66$} & \multicolumn{2}{c}{$0.10\,\pm\,0.11$} & \multicolumn{2}{c}{$978.98\,\pm\,29.57$} \\
\texttt{GR00T} & \multicolumn{2}{|c|}{$0.04\,\pm\,0.09$} & \multicolumn{2}{c}{$0.05\,\pm\,0.03$} & \multicolumn{2}{c}{$0.50\,\pm\,0.02$} & \multicolumn{2}{c}{$0.12\,\pm\,0.02$} & \multicolumn{2}{c}{$4.70\,\pm\,0.35$} & \multicolumn{2}{c}{$12.16\,\pm\,0.63$} & \multicolumn{2}{c}{$2.56\,\pm\,0.50$} & \multicolumn{2}{c}{$32.41\,\pm\,1.85$} & \multicolumn{2}{c}{$41.87\,\pm\,2.31$} & \multicolumn{2}{c}{$233.11\,\pm\,6.53$} & \multicolumn{2}{c}{$22.17\,\pm\,1.39$} & \multicolumn{2}{c}{$23.60\,\pm\,2.49$} & \multicolumn{2}{c}{\cellcolor{ForestGreen!20}$\mathbf{0.00}\,\pm\,0.07$} & \multicolumn{2}{c}{$965.66\,\pm\,47.12$} \\
\texttt{xVLA} & \multicolumn{2}{|c|}{$0.00\,\pm\,0.07$} & \multicolumn{2}{c}{$0.01\,\pm\,0.02$} & \multicolumn{2}{c}{$0.54\,\pm\,0.05$} & \multicolumn{2}{c}{$0.17\,\pm\,0.03$} & \multicolumn{2}{c}{$4.34\,\pm\,0.46$} & \multicolumn{2}{c}{$8.74\,\pm\,0.36$} & \multicolumn{2}{c}{$3.22\,\pm\,0.59$} & \multicolumn{2}{c}{$27.38\,\pm\,2.32$} & \multicolumn{2}{c}{$22.73\,\pm\,2.43$} & \multicolumn{2}{c}{$115.21\,\pm\,11.41$} & \multicolumn{2}{c}{$19.44\,\pm\,1.86$} & \multicolumn{2}{c}{$9.34\,\pm\,2.24$} & \multicolumn{2}{c}{\cellcolor{ForestGreen!20}$\mathbf{0.00}\,\pm\,0.07$} & \multicolumn{2}{c}{$974.08\,\pm\,29.35$} \\
\texttt{Pi\_0.5} & \multicolumn{2}{|c|}{$0.40\,\pm\,0.14$} & \multicolumn{2}{c}{$0.44\,\pm\,0.07$} & \multicolumn{2}{c}{$0.34\,\pm\,0.02$} & \multicolumn{2}{c}{$0.05\,\pm\,0.01$} & \multicolumn{2}{c}{$2.03\,\pm\,0.31$} & \multicolumn{2}{c}{$5.87\,\pm\,0.88$} & \multicolumn{2}{c}{$2.16\,\pm\,0.59$} & \multicolumn{2}{c}{$10.44\,\pm\,1.76$} & \multicolumn{2}{c}{$11.37\,\pm\,0.75$} & \multicolumn{2}{c}{$57.57\,\pm\,4.17$} & \multicolumn{2}{c}{$7.60\,\pm\,1.29$} & \multicolumn{2}{c}{$1.94\,\pm\,1.38$} & \multicolumn{2}{c}{$0.12\,\pm\,0.12$} & \multicolumn{2}{c}{$710.78\,\pm\,108.67$} \\
\texttt{MolmoAct} & \multicolumn{2}{|c|}{\cellcolor{ForestGreen!20}$\mathbf{0.52}\,\pm\,0.13$} & \multicolumn{2}{c}{\cellcolor{ForestGreen!20}$\mathbf{0.48}\,\pm\,0.07$} & \multicolumn{2}{c}{\cellcolor{ForestGreen!20}$\mathbf{0.31}\,\pm\,0.04$} & \multicolumn{2}{c}{\cellcolor{ForestGreen!20}$\mathbf{0.04}\,\pm\,0.00$} & \multicolumn{2}{c}{\cellcolor{ForestGreen!20}$\mathbf{1.34}\,\pm\,0.29$} & \multicolumn{2}{c}{\cellcolor{ForestGreen!20}$\mathbf{4.70}\,\pm\,1.05$} & \multicolumn{2}{c}{\cellcolor{ForestGreen!20}$\mathbf{1.04}\,\pm\,0.36$} & \multicolumn{2}{c}{\cellcolor{ForestGreen!20}$\mathbf{6.16}\,\pm\,1.35$} & \multicolumn{2}{c}{\cellcolor{ForestGreen!20}$\mathbf{12.91}\,\pm\,1.46$} & \multicolumn{2}{c}{\cellcolor{ForestGreen!20}$\mathbf{58.79}\,\pm\,5.30$} & \multicolumn{2}{c}{\cellcolor{ForestGreen!20}$\mathbf{4.49}\,\pm\,1.02$} & \multicolumn{2}{c}{\cellcolor{ForestGreen!20}$\mathbf{0.38}\,\pm\,0.63$} & \multicolumn{2}{c}{$0.04\,\pm\,0.09$} & \multicolumn{2}{c}{\cellcolor{ForestGreen!20}$\mathbf{545.88}\,\pm\,123.97$} \\
\midrule
\texttt{Human} & \multicolumn{2}{|c|}{$1.00\,\pm\,0.00$} & \multicolumn{2}{c}{$1.00\,\pm\,0.00$} & \multicolumn{2}{c}{$0.40\,\pm\,0.03$} & \multicolumn{2}{c}{$0.05\,\pm\,0.00$} & \multicolumn{2}{c}{$0.42\,\pm\,0.04$} & \multicolumn{2}{c}{$0.58\,\pm\,0.06$} & \multicolumn{2}{c}{$0.24\,\pm\,0.11$} & \multicolumn{2}{c}{$1.56\,\pm\,0.17$} & \multicolumn{2}{c}{$23.24\,\pm\,1.28$} & \multicolumn{2}{c}{$95.97\,\pm\,5.04$} & \multicolumn{2}{c}{$1.11\,\pm\,0.14$} & \multicolumn{2}{c}{$0.00\,\pm\,0.05$} & \multicolumn{2}{c}{$0.00\,\pm\,0.05$} & \multicolumn{2}{c}{$83.78\,\pm\,9.01$} \\
\bottomrule
\end{tabular}
\end{adjustbox}
\label{tab:metrics_summary_combined}
\vspace{2pt}
\noindent\begin{minipage}{\linewidth}
\footnotesize
\textbf{Abbreviations:} Success = Success Rate [\%]; TP = Task Progression [\%]; BAVD = Bimanual Arm Velocity Difference [m/s]; BGVD = Bimanual Gripper Vertical Difference [m]; CPL = Cartesian Path Length [m]; CT = Completion Time [s]; ECC = Environment Collision Count; JPL = Joint Path Length [rad]; MCJ = Mean Cartesian Jerk [m/s\textsuperscript{3}]; MJJ = Mean Joint Jerk [rad/s\textsuperscript{3}]; OPL = Orientation Path Length [rad]; SCC = Self-Collision Count; SC = Slip Count; TL = Trajectory Length [steps].
\end{minipage}
\vspace{-1.5em}
\end{table*}

\subsection{Real-world Zero-shot DROID}
\textbf{Implementation.} We adopt the DROID setup~\citep{khazatsky2024droid} in a kitchen environment, and design five tasks, each with three spatial variants, for evaluation. Camera positions are held fixed across all three models.

\begin{figure}[htb!]
    \centering
    \includegraphics[width=\linewidth]{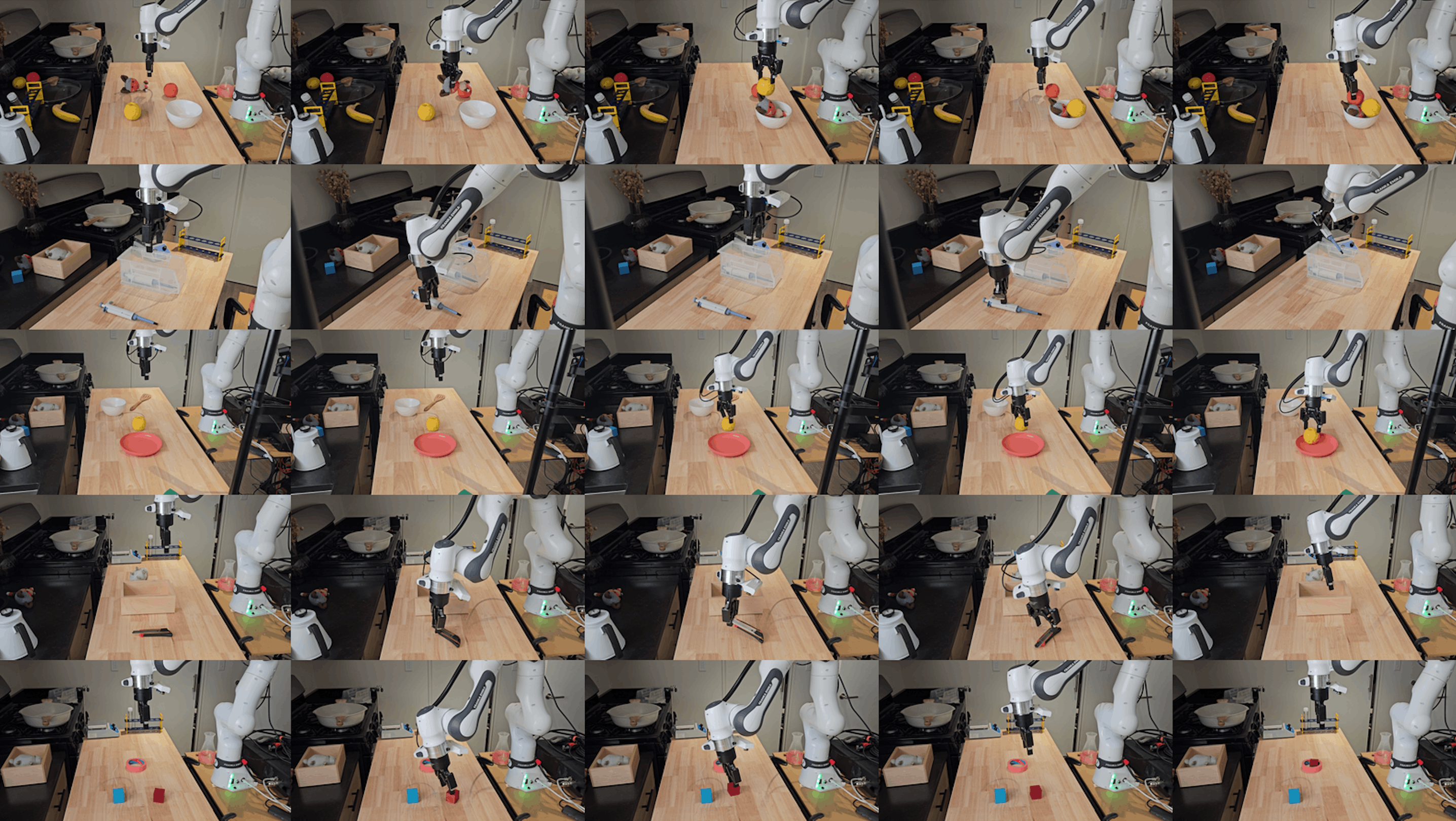}
    \caption{\textbf{Sample trajectories from the DROID evaluation of MolmoAct2.} }
    \label{fig:droid_sample}
\end{figure}

Figure \ref{fig:droid_sample} shows sample trajectories of MolmoAct2 from 5 different tasks. Table \ref{tab:policy-results} presents the evaluation results for MolmoAct2, $\pi_0$, and MolmoBot in the DROID setup.

\begin{table}[h]
\centering
\caption{Per-task and per-position success rates across policies. Each task uses 15 trajectories (5 per position).}
\label{tab:policy-results}
\begin{threeparttable}
\begin{tabular}{llcccc}
\toprule
\textbf{Task} & \textbf{Policy} & \textbf{Pos.\ 1} & \textbf{Pos.\ 2} & \textbf{Pos.\ 3} & \textbf{Overall} \\
\midrule
\multirow{3}{*}{Put the apple on the plate}
  & MolmoAct2 & 5/5 (100\%) & 5/5 (100\%) & 5/5 (100\%) & \textbf{15/15 (100.0\%)} \\
  & $\pi_05$-DROID   & 5/5 (100\%) & 3/5 (60\%)  & 2/5 (40\%)  & 10/15 (66.7\%) \\
  & MolmoBot  & 5/5 (100\%) & 4/5 (80\%)  & 4/5 (80\%)  & 13/15 (86.7\%) \\
\midrule
\multirow{3}{*}{Put the pipette in the tray}
  & MolmoAct2 & 4/5 (80\%)  & 4/5 (80\%)  & 5/5 (100\%) & \textbf{13/15 (86.7\%)} \\
  & $\pi_05$-DROID   & 2/5 (40\%)  & 3/5 (60\%)  & 0/5 (0\%)   & 5/15 (33.3\%) \\
  & MolmoBot  & 3/5 (60\%)  & 2/5 (40\%)  & 3/5 (60\%)  & 8/15 (53.3\%) \\
\midrule
\multirow{3}{*}{Put the red cube inside the tape roll}
  & MolmoAct2 & 5/5 (100\%) & 4/5 (80\%)  & 5/5 (100\%) & \textbf{14/15 (93.3\%)} \\
  & $\pi_05$-DROID   & 3/5 (60\%)  & 5/5 (100\%) & 0/5 (0\%)   & 8/15 (53.3\%) \\
  & MolmoBot  & 1/5 (20\%)  & 0/5 (0\%)   & 4/5 (80\%)  & 5/15 (33.3\%) \\
\midrule
\multirow{3}{*}{Put the knife in the box}
  & MolmoAct2 & 5/5 (100\%) & 5/5 (100\%) & 4/5 (80\%)  & \textbf{14/15 (93.3\%)} \\
  & $\pi_05$-DROID   & 3/5 (60\%)  & 1/5 (20\%)  & 0/5 (0\%)   & 4/15 (26.7\%) \\
  & MolmoBot  & 1/5 (20\%)  & 0/5 (0\%)   & 5/5 (100\%) & 6/15 (40.0\%) \\
\midrule
\multirow{3}{*}{Put the objects in the bowl\tnote{a}}
  & MolmoAct2 & 73.2\% & 53.0\% & 59.8\% & \textbf{62.0\%} \\
  & $\pi_05$-DROID   & 39.6\% & 39.6\% & 59.4\% & 46.2\% \\
  & MolmoBot  & 33.0\% & 26.4\% & 26.4\% & 28.6\% \\
\bottomrule
\end{tabular}
\begin{tablenotes}\footnotesize
\item[a] Partial-credit task scored on $\{0,\,0.33,\,0.66,\,1.0\}$ per trajectory; reported as mean success rate.
\end{tablenotes}
\end{threeparttable}
\end{table}

\subsection{Real-world Bimanual YAM}
\textbf{Implementation.} We trained \molmoacttpos as eight separate single-task policies; to ensure a fair comparison, we applied the identical training protocol to each of the four baseline policies. All policies were trained to convergence, and the resulting checkpoints were used for evaluation. Evaluation and setup were carried out by Cortex AI: each task was collected under three spatial variants and evaluated under the same three variants.

\begin{figure}[htb!]
    \centering
    \includegraphics[width=\linewidth]{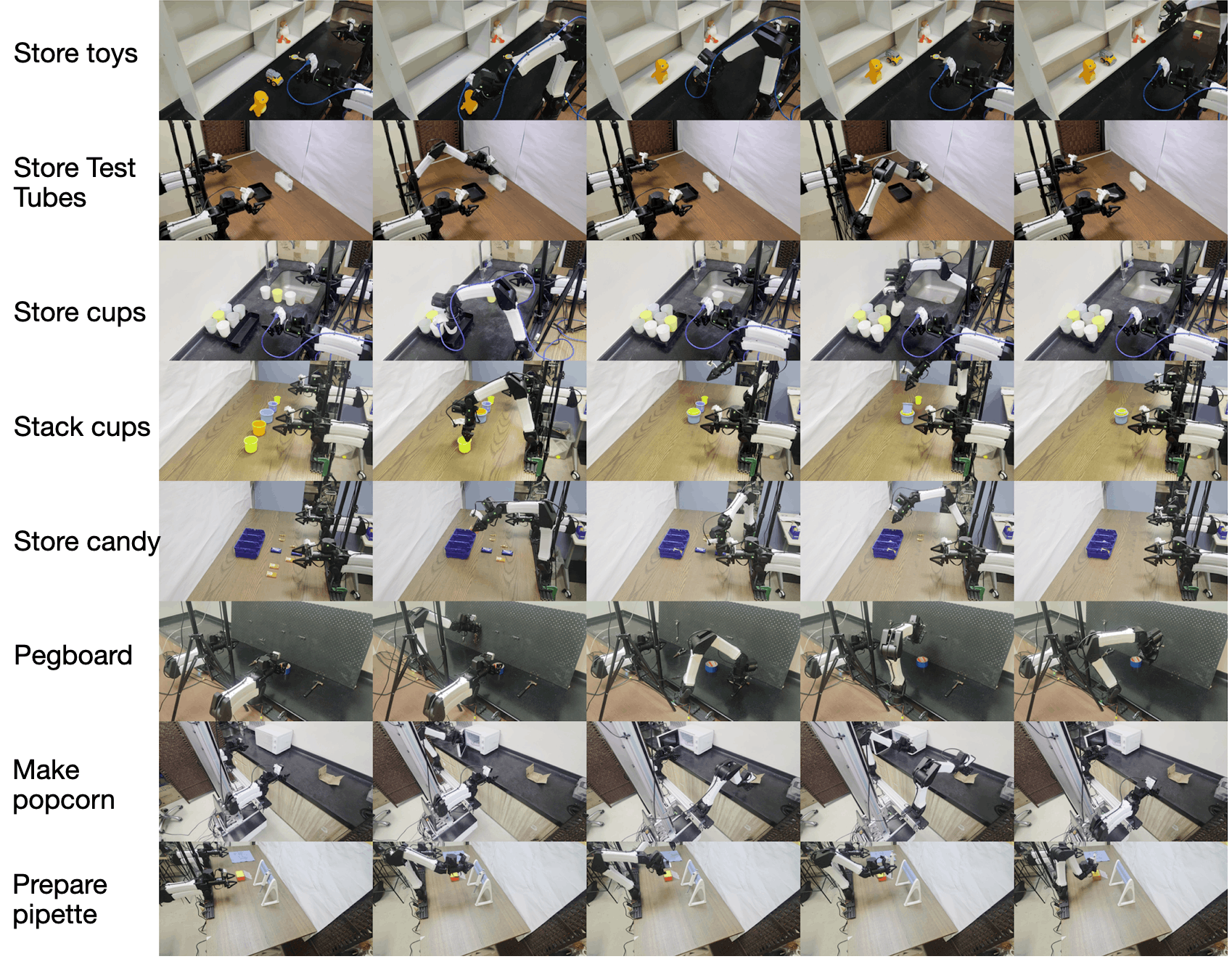}
    \caption{\textbf{Sample trajectories from the Bimanual YAM evaluation of MolmoAct2.} }
    \label{fig:so100_sample}
\end{figure}

\begin{figure}[htb!]
    \centering
    \includegraphics[width=\linewidth]{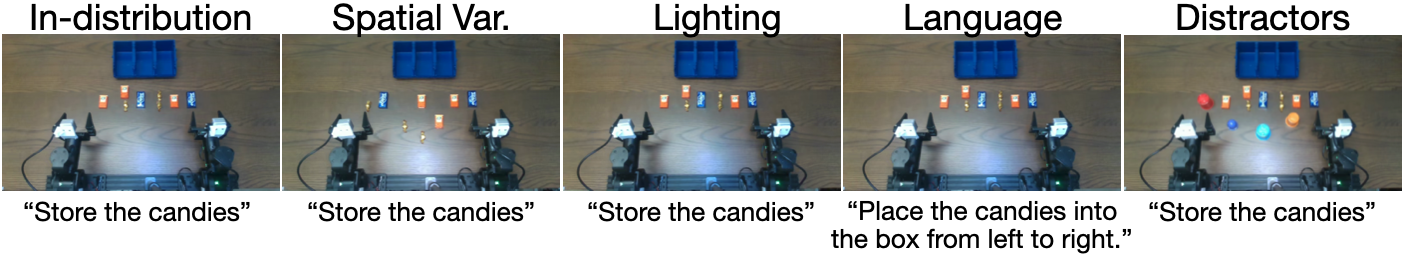}
    \caption{\textbf{Sample trajectories from the out-of-distribution Bimanual YAM evaluation of MolmoAct2.} }
    \label{fig:so100_sample}
\end{figure}

\begin{table}[t]
\centering
\caption{Overall success rate (\%) across 8 manipulation tasks (50 episodes each). Best model per task is \textbf{bold}.}
\label{tab:overall_results}
\setlength{\tabcolsep}{5pt}
\renewcommand{\arraystretch}{1.15}
\begin{tabular}{lccccc}
\toprule
\textbf{Task} & \textbf{$\pi_{0.5}$} & \textbf{Cosmos} & \textbf{X-VLA} & \textbf{OpenVLA} & \textbf{MolmoAct} \\
\midrule
Cup Stacking         & 62.0 & \textbf{64.5} & 5.5  & 60.5 & 54.0 \\
Linearbot            & 11.6 & 9.2  & 5.2  & 24.4 & \textbf{64.0} \\
Pegboard             & 2.0  & 0.0  & 3.0  & 4.5  & \textbf{14.0} \\
Cup Storing          & 85.9 & 0.0  & 7.9  & 96.0 & \textbf{100.0} \\
Candy Storing        & 26.0 & 0.0  & 3.8  & 26.5 & \textbf{40.5} \\
Store Test Tube      & 17.0 & 11.0 & 0.0  & 42.0 & \textbf{67.0} \\
Store Toys           & 26.0 & 7.0  & 2.5  & 2.0  & \textbf{32.0} \\
Prepare Pipette      & 27.5 & 32.0 & 11.5 & 28.5 & \textbf{33.0} \\
\midrule
\textbf{Average}     & 32.2 & 15.5 & 4.9  & 35.5 & \textbf{50.6} \\
\bottomrule
\end{tabular}
\end{table}

\subsection{Real-world SO-100/101 zero-shot}

\textbf{Implementation.} We conduct zero-shot evaluation on the SO-100 platform using a fixed, pre-initialized camera viewpoint, and compare against DePi and SmolVLA. Episodes are graded under a partial-credit scoring scheme, and no additional training of the policies is performed.

\begin{table*}[t]
\centering
\caption{Zero-shot success rate (\%) on SO-100 tasks with fixed initial camera position. Each task is graded with partial credit at three milestones; partial scores of 0.25, 0.5, and 1.0 are awarded for reaching the corresponding stage. 15 episodes per task. Best model per task is \textbf{bold}.}
\label{tab:so100_results}
\setlength{\tabcolsep}{5pt}
\renewcommand{\arraystretch}{1.2}
\begin{tabular}{lcccccc}
\toprule
& \multicolumn{3}{c}{\textbf{Partial-Credit Milestones}} & \multicolumn{3}{c}{\textbf{Success Rate (\%)}} \\
\cmidrule(lr){2-4}\cmidrule(lr){5-7}
\textbf{Task} & \textbf{0.25} & \textbf{0.5} & \textbf{1.0} & \textbf{SmolVLA} & \textbf{MolmoAct v2} & \textbf{DePi$_0$} \\
\midrule
Place fork on plate     & reach & pickup fork  & place on plate & 3.3 & \textbf{70.0} & 30.0          \\
Stack blocks            & reach & pickup block & stack blocks   & 5.0 & \textbf{20.0} & 6.7           \\
Place tissues in basket & reach & pickup       & place          & 0.0 & \textbf{73.3} & 20.0          \\
Place pen on notebook   & reach & pickup       & place          & 3.3 & \textbf{86.7} & 80.0          \\
Place block in box      & reach & pickup       & place          & 0.0 & 33.3          & \textbf{90.0} \\
\midrule
\multicolumn{4}{l}{\textbf{Average}} & 2.3 & \textbf{56.7} & 45.3 \\
\bottomrule
\end{tabular}
\end{table*}

\begin{figure}[htb!]
    \centering
    \includegraphics[width=\linewidth]{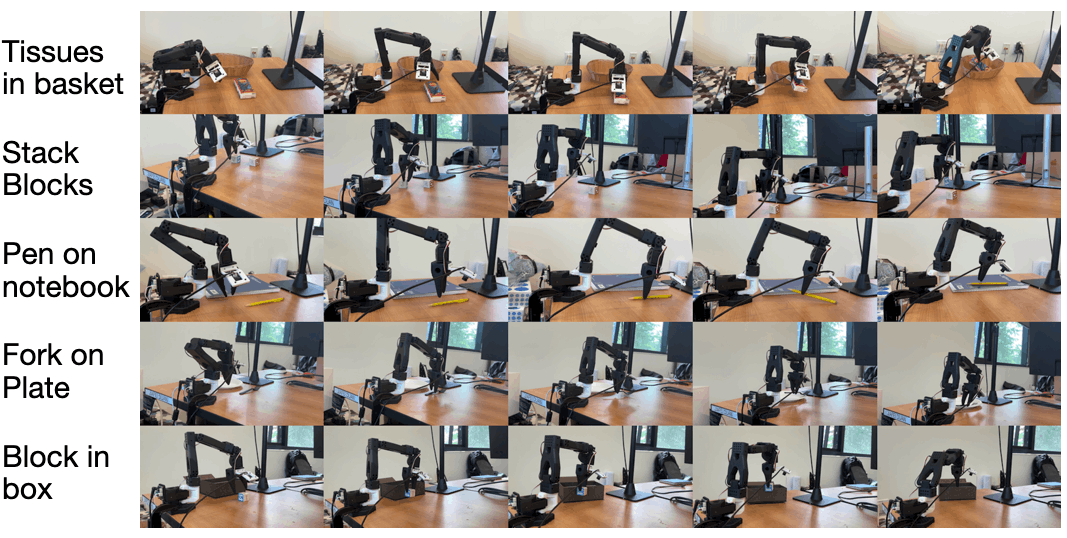}
    \caption{\textbf{Sample trajectories from the SO-100 evaluation of MolmoAct2.} }
    \label{fig:so100_sample}
\end{figure}

\clearpage

\section{Data Details}
\label{supp:data}
\subsection{\molmoacttsodata Statistics}
\label{supp:so100-so101-stats}

\begin{figure}[htb!]
    \centering
    \includegraphics[width=0.9\linewidth]{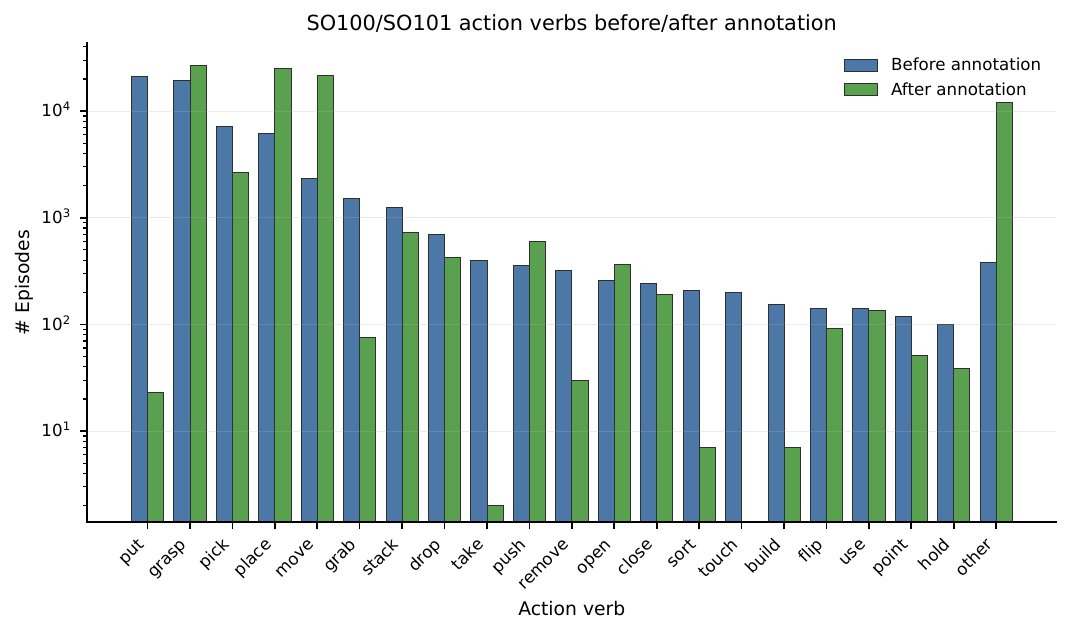}
    \caption{\textbf{Action verb distribution before and after language annotation.} }
    \label{fig:action_verbs}
\end{figure}

We use \molmoacttsodata in the \molmoactt training mixture. We first collect 1{,}660 dataset entries; then we use TOPReward \citep{chen2026topreward} to filter out 438
low-quality dataset entries, yielding 1{,}222 final
datasets. All statistics below are computed only over this final filtered set. \Cref{tab:so100-so101-summary} presents the summary statistics for the SO-100/101 training partition. \Cref{tab:so100-so101-robot-type} presents the number of datasets by the type of SO-10x variants. \Cref{tab:so100-so101-metadata} presents the distribution of different metadata such as FPS, camera configuration, and action/state dimensions. \Cref{fig:action_verbs} presents the distribution of action verbs before and after the language annotation pipeline described in \Cref{sec:lang_annot}. \Cref{fig:so100_sample} shows several sample trajectories from the dataset partition. 

\begin{figure}[htb!]
    \centering
    \includegraphics[width=0.9\linewidth]{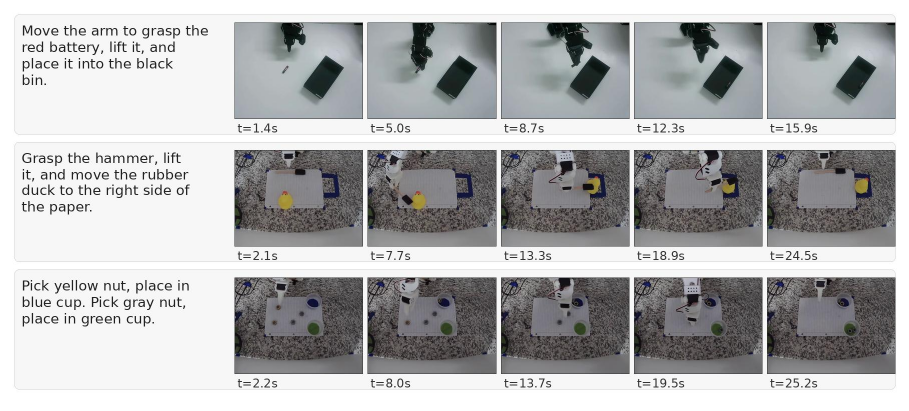}
    \caption{\textbf{Sample trajectories from the SO10x dataset partition.} }
    \label{fig:so100_sample}
\end{figure}

\begin{table}[ht]
\centering
\footnotesize
\caption{\textbf{Summary statistics for the SO-100/101 training subset.} We
report statistics for the final filtered \molmoacttsodata partition.}
\setlength{\tabcolsep}{6pt}
\renewcommand{\arraystretch}{1.1}
\begin{tabular}{lr}
\textbf{Statistic} & \textbf{Value} \\
\midrule
\# datasets & 1{,}222 \\
Contributors & 377 \\
Episodes & 38{,}059 \\
Frames & 19.8M \\
Estimated duration & 183.6 hours \\
Metadata task count & 1{,}322 \\
Instruction rows & 38{,}059 \\
Globally unique instructions & 12{,}818 \\
Unique-instruction ratio & 33.7\% \\
Datasets with annotated task metadata & 1{,}222 \\
\end{tabular}
\label{tab:so100-so101-summary}
\end{table}

\begin{table}[t]
\centering
\footnotesize
\caption{\textbf{SO-10x data partition breakdown by robot type.}}
\setlength{\tabcolsep}{5pt}
\renewcommand{\arraystretch}{1.1}
\begin{tabular}{lrrrr}
\textbf{Robot type} & \textbf{Datasets} & \textbf{Episodes} & \textbf{Frames} & \textbf{Hours} \\
\midrule
SO-100 & 921 & 31{,}101 & 15.18M & 141.7 \\
SO-101 & 299 & 6{,}898 & 4.57M & 41.6 \\
MOSS & 2 & 60 & 15.6K & 0.2 \\
\end{tabular}
\label{tab:so100-so101-robot-type}
\end{table}

\begin{table}[t]
\centering
\footnotesize
\caption{\textbf{Recording and feature metadata.} Most demonstrations are
recorded at 30 Hz with two camera streams; all final filtered datasets use
six-dimensional action and state vectors.}
\setlength{\tabcolsep}{6pt}
\renewcommand{\arraystretch}{1.1}
\begin{tabular}{lrrrr}
\textbf{Metadata field} & \textbf{Value} & \textbf{Datasets} & \textbf{Episodes} & \textbf{Frames} \\
\midrule
FPS & 30 & 1{,}114 & 36{,}828 & 18.96M \\
FPS & 20 & 78 & 527 & 261.4K \\
FPS & 25 & 18 & 423 & 256.7K \\
FPS & 60 & 8 & 208 & 231.6K \\
Camera count & 1 & 230 & 4{,}159 & 2.26M \\
Camera count & 2 & 835 & 24{,}151 & 13.05M \\
Camera count & 3 & 151 & 9{,}235 & 4.11M \\
Camera count & 4 & 6 & 514 & 342.1K \\
Action dimension & 6 & 1{,}222 & 38{,}059 & 19.76M \\
State dimension & 6 & 1{,}222 & 38{,}059 & 19.76M \\
\end{tabular}
\label{tab:so100-so101-metadata}
\end{table}

\subsection{Language Annotation Pipeline Implementation Details}
\label{supp:language-annotation-details}
Running inference with temperature 0.1 and max\_length set to 1024 tokens, we feed the following prompt into Qwen3.5-27B:
\begin{quote} 
You are an expert robot arm task instruction generator. You are provided with multiple demonstration videos, each showing the exact same task execution captured simultaneously from different camera angles. The task is to \{original\_instruction\}. Based on these videos, generate a clear imperative instruction describing all of the actions performed by the robot arm. Carefully study the evolution of the state of the scene, to understand how the robot arm achieves the goal. Pay attention to the color and position of the objects the robot interacts with. Start directly with an action verb and make sure you do not miss any actions. Note that the robot arm is not always grasping an object and maybe pushing objects around instead. Verify any grasps by checking the gripper state, and only say the arm is grasping an object if this is shown in the wrist camera when the gripper is closed. Your prompts should be almost exactly \{word\_limit\} words long.
\end{quote}

 word\_limit is a number randomly sampled from a right tailed power distribution, with a minimum value of 5 and a maximum value of 25. In addition to the text prompt, for each of the cameras in the trajectory that are not frozen, we sample 12 frames that are each equally temporally spaced and pass the frames to the model. 
 
 Running this pipeline results in the changes to the number of unique instructions in the datasets shown in Table \ref{tab:language-annot_increased}.

\begin{table}[t]
\centering
\footnotesize
\caption{\textbf{Dataset Language Annotation Diversity.} Running the language annotation pipeline increases the number of unique instructions for all of the robotics datasets in \molmoactt's training mix.}
\setlength{\tabcolsep}{5pt}
\renewcommand{\arraystretch}{1.1}
\begin{tabular}{lcc}
\textbf{Dataset} & \textbf{Original Instructions} & \textbf{Relabeled Instructions}\\
\midrule
\molmoacttyamdata & 34 (0.1\%)     & 11448 (35\%)   \\
\midrule
\molmoacttdroiddata            & 49623 (67\%)   & 55905 (75\%)   \\
\midrule
\molmoacttsodata            & 707 (1.5\%)    & 16205 (34\%)   \\
\midrule
\molmoactdata       & 134 (1\%)      & 2959 (28\%)    \\
\midrule
BC-Z             & 104 (0.3\%)    & 9981 (25\%)    \\
\midrule
Fractal          & 598 (0.7\%)    & 16729 (19\%)   \\
\midrule
Bridge           & 19973 (37\%)   & 35096 (66\%)   \\
\midrule
\textbf{Total}   & \textbf{71121 (22\%)} & \textbf{146485 (46\%)} \\
\end{tabular}
\label{tab:language-annot_increased}
\end{table}

\begin{table}[t]
\centering
\footnotesize
\caption{\textbf{Past Tabletop Bimanual Manipulation Datasets.} \molmoacttyamdata is the largest open-source tabletop bimanual robotics dataset. In this table, we compare against other open and closed-source tabletop bimanual datasets from the literature.}
\setlength{\tabcolsep}{5pt}
\renewcommand{\arraystretch}{1.1}
\begin{tabular}{lccc}
\toprule
\textbf{Dataset Name} & \textbf{Embodiment} & \textbf{Number of Trajectories} & \textbf{Open / Closed?}\\
\midrule
{\color{ai2pink}\molmoacttyamdata (Ours)} & {\color{ai2pink}Bimanual YAM} & {\color{ai2pink}34,500} & {\color{ai2pink}Open}  \\ 
\midrule
AIST Bimanual Manipulation Dataset~\citep{motoda2025aistdata} & ALOHA & 10,000 & Open \\ 
\midrule
BiPlay Dataset~\citep{dasari2025ingredients} & ALOHA & 7,000 & Open \\ 
\midrule
RDT-1B Finetuning Dataset~\citep{liu2024rdt} & ALOHA & 6,000 & Open \\ 
\midrule
ALOHA Unleashed Training Data~\citep{zhao2024aloha} & ALOHA & 26,000 & Closed \\
\midrule
SARM RA-BC Dataset~\citep{chen2025sarm} & Bimanual YAM & 200 Hours & Closed \\ 
\bottomrule
\end{tabular}
\label{tab:past_robot_data}
\end{table}

\clearpage



\section{Limitations and Potential Solutions}
\label{supp:limit}

While \molmoactt is all quite capable as a general-purpose action reasoning model, it is not without limitations. In the following sections, we discuss some of these limitations and potential solutions.

\paragraph{Action chunks and the absence of real-time re-chunking.}
\molmoactt predicts fixed-horizon action chunks (30 steps at 30~Hz for YAM/SO-100/101, 15 at 15~Hz for DROID, 10 at 10~Hz for LIBERO) and executes them open-loop before re-querying the policy. This has two consequences worth acknowledging. First, motion smoothness across chunk boundaries is not enforced anywhere in the pipeline: the flow-matching expert is trained to denoise each chunk independently, with no continuity loss tying the end of chunk $t$ to the start of chunk $t+1$. In practice this can produce visible velocity or acceleration discontinuities at the seams, particularly when the VLM context shifts between queries (new observation, slightly different attention state). Second, because the policy commits to a full chunk before observing its consequences, it cannot react within-chunk to perturbations, contact events, or its own tracking error --- the 55.79~Hz number in Sec \ref{speed} is amortized chunk throughput, not closed-loop reactivity.

\paragraph{Embodiment-specific zero-shot deployment.}
The zero-shot deployment story for \textsc{MolmoAct2} is currently tied to the three embodiments for which we have large-scale training data: \molmoacttyam, \molmoacttso, and \molmoacttdroid can each be deployed out-of-the-box on their respective robot setups (bimanual YAM, SO-100/101, and DROID Franka) Sec. \ref{sec:exp-outbox}. This is a direct consequence of our data strategy --- concentrating collection and curation effort on these three platforms to span the low-to-medium cost range --- but it also bounds what zero-shot means here. \textsc{MolmoAct2} is not a universal controller that transfers without adaptation to arbitrary new embodiments: deployment on robots outside this set (e.g., other bimanual platforms, dexterous hands, mobile manipulators with different kinematics, or humanoids) requires effective fine-tuning on target-embodiment demonstrations, as shown in our fine-tuning studies in Sec \ref{sec:exp-ft}. Extending the zero-shot envelope to a broader range of embodiments is a matter of scaling data collection along the same axes we have begun here, and we view the released datasets and training recipe as a foundation for that effort rather than a finished product.

\end{document}